\documentclass[journal]{IEEEtran}

\usepackage{cite}
\ifCLASSINFOpdf
   \usepackage[pdftex]{graphicx}
\else
\fi
\usepackage{amsmath}
\usepackage{xcolor}
\usepackage{amssymb}
\usepackage[switch]{lineno}

\begin{document}
%
\title{Uncertainty-Aware Unsupervised Domain Adaptation in Object Detection}
%
%
%

\author{Michael~Shell,~\IEEEmembership{Member,~IEEE,}
        John~Doe,~\IEEEmembership{Fellow,~OSA,}
        and~Jane~Doe,~\IEEEmembership{Life~Fellow,~IEEE}
\thanks{M. Shell was with the Department
of Electrical and Computer Engineering, Georgia Institute of Technology, Atlanta,
GA, 30332 USA e-mail: (see http://www.michaelshell.org/contact.html).}
\thanks{J. Doe and J. Doe are with Anonymous University.}
\thanks{Manuscript received April 19, 2005; revised August 26, 2015.}}

\author{Dayan Guan, Jiaxing Huang, Aoran Xiao, Shijian Lu, Yanpeng Cao
\thanks{Corresponding author: Shijian Lu (e-mail: Shijian.Lu@ntu.edu.sg)}
\thanks{D. Guan, J. Huang, A. Xiao and S. Lu are with Singtel Cognitive and Artificial Intelligence Lab for Enterprises, Nanyang Technological University, Singapore.}
\thanks{Y. Cao is with the School of Mechanical Engineering, Zhejiang University, Hangzhou, China.}}

\maketitle

\begin{abstract}
Unsupervised domain adaptive object detection aims to adapt detectors from a labelled source domain to an unlabelled target domain. Most existing works take a two-stage strategy that first generates region proposals and then detects objects of interest, where adversarial learning is widely adopted to mitigate the inter-domain discrepancy in both stages. However, adversarial learning may impair the alignment of well-aligned samples as it merely aligns the global distributions across domains. To address this issue, we design an uncertainty-aware domain adaptation network (UaDAN) that introduces conditional adversarial learning to align well-aligned and poorly-aligned samples separately in different manners. Specifically, we design an uncertainty metric that assesses the alignment of each sample and adjusts the strength of adversarial learning for well-aligned and poorly-aligned samples adaptively. In addition, we exploit the uncertainty metric to achieve curriculum learning that first performs easier image-level alignment and then more difficult instance-level alignment progressively. Extensive experiments over four challenging domain adaptive object detection datasets show that UaDAN achieves superior performance as compared with state-of-the-art methods. 
\end{abstract}

\begin{IEEEkeywords}
Unsupervised domain adaptation, object detection, adversarial learning, curriculum learning.
\end{IEEEkeywords}

%
\IEEEpeerreviewmaketitle

\section{Introduction}

\IEEEPARstart{O}{bject} detection aims to locate and recognize objects of interest in images, which has been a longstanding challenge in computer vision research~\cite{papageorgiou1998general,viola2001robust,nascimento2006performance,felzenszwalb2009object,li2013co,dollar2014fast}. With the development of deep convolutional neural networks in recent years, object detection has achieved great progress~\cite{girshick2014rich,girshick2015fast,ren2015faster,redmon2016you,liu2016ssd,redmon2017yolo9000,lin2017focal,law2018cornernet,tian2019fcos,dai2016r,lin2017feature,hu2018relation,cai2019cascade,pang2019libra,chen2017s,qiu2020hierarchical,tan2020efficientdet,liu2020structured} over multiple large-scale datasets~\cite{everingham2010pascal,lin2014microsoft,cordts2016cityscapes,zhou2017scene,neuhold2017mapillary}. However, existing object detection methods usually experience drastic performance drops when applied to new datasets due to various domain biases in camera settings, environmental illumination, object appearance, etc. Annotating a fair number of samples for each new data collection can alleviate this problem effectively, but it is often prohibitively time-consuming and unscalable while facing various new data.

\begin{figure}[ht]
\centering
\includegraphics[width=.98\linewidth]{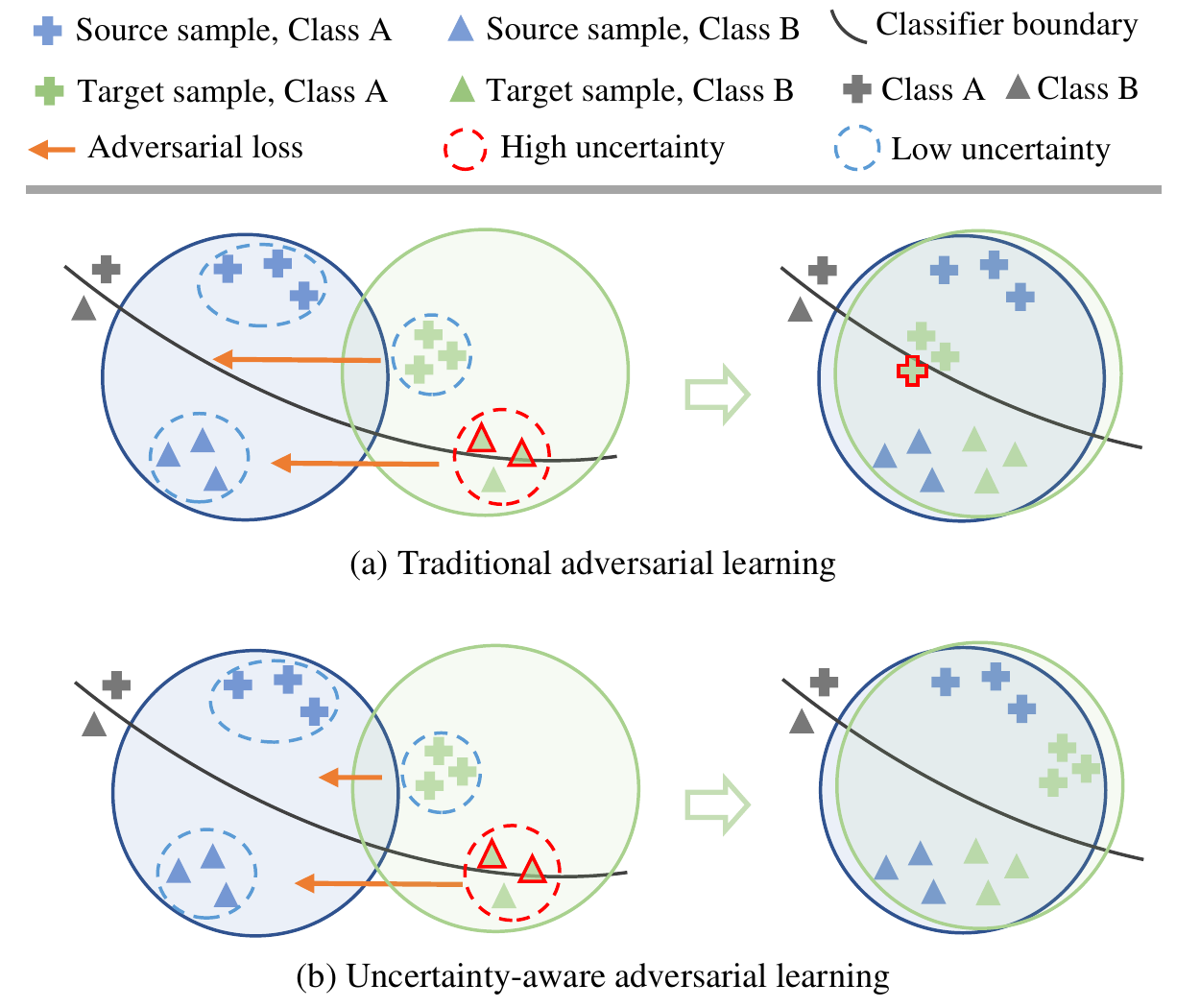}
\caption{
The proposed uncertainty-aware adversarial learning (UaAL) performs adversarial learning adaptively: Traditional adversarial learning assigns adversarial loss of the same weight to all samples (global alignment), which may misalign well-aligned samples (far from classifier boundary with low uncertainty) to incorrect classes as in (a). UaAL weights adversarial loss adaptively based on the uncertainty of each sample. It can thus protect well-aligned samples from further alignment while focusing on aligning under-aligned samples (near to classifier boundary with high uncertainty) as in (b). Here falsely classified samples are highlighted in red color and the length of orange-color arrows denotes adversarial-loss weights - UaAL assigns smaller weights (shorter arrow in (b)) to well-aligned samples. Best viewed in color.}
\label{fig:intro}
\end{figure}

Unsupervised domain adaptation has been explored extensively~\cite{roy2013towards,yang2014cross,huang2021fsdr,zhai2020ad,huang2021cross,yan2019weighted,huang2020contextual,guan2021scale} to address domain biases by learning a well-performed model on an unlabeled target domain with supervision of a labeled source domain.
Motivated by the domain adaptation theory~\cite{ben2010theory} that the upper bound of target-domain errors could be reduced by minimizing the divergence between source and target domains, most existing domain adaptive object detection methods adopt adversarial learning to minimize the cross-domain discrepancy~\cite{chen2018domain,saito2019strong,zhang2021self,zhuang2020ifan,he2020domain,zhao2020collaborative,zheng2020cross,xu2020exploring,zhang2021detr,zhan2019ga}. Though these works have achieved impressive performance, they usually suffer from a common constraint of the adversarial learning - it merely considers the alignment of global distributions and may pull well-aligned samples to incorrect classes as illustrated in Fig.~\ref{fig:intro}~(a).

In this paper, we propose an uncertainty-aware domain adaptation network (UaDAN) that aligns well-aligned and poorly-aligned samples adaptively by considering their uncertainty. With supervised models trained using source samples, well-aligned target samples are often predicted with low uncertainty while poorly-aligned target samples predicted with high uncertainty. We exploit this uncertainty information and achieve uncertainty-aware cross-domain alignment with two innovative designs. The first design is uncertainty-aware adversarial learning (UaAL) as illustrated in Fig.~\ref{fig:intro}~(b). Instead of assigning the same adversarial-loss weight to all target samples equally, UaAL introduces uncertainty to measure an alignment score for each sample and adjusts the adversarial-loss weights for well-aligned and poorly-aligned target samples adaptively. In implementation, we exploit entropy~\cite{shannon1948mathematical} to estimate the sample uncertainty and weight adversarial loss of different target samples adaptively (\textit{i.e.}, small weight for well-aligned samples and large weight for poorly-aligned samples).

The second design is uncertainty-guided curriculum learning (UgCL) that aims to optimize the domain adaptation across the two-stage detection pipeline progressively. Specifically, UgCL first aligns target samples at an easier image level for region proposal generation. It exploits entropy to estimate the uncertainty of image-level predictions for uncertainty-aware alignment at the image level. After target samples are well-aligned at the image level (with low uncertainty), UgCL aligns at a harder instance level for final object detection. Here the uncertainty is exploited to guide the instance-level alignment in the similar manner. The two-stage progressive alignment strategy alleviates error accumulation effectively by reducing image-level misalignment and its effects over the later instance-level alignment.

The contributions of this work can be summarized in three aspects. \textit{First}, we propose an uncertainty-aware domain adaptation network that assesses the uncertainty of sample predictions and employs it for adaptive sample alignment effectively. \textit{Second}, we develop an uncertainty-aware adversarial learning method that alleviates misalignment of well-aligned samples by assigning high weights to high-uncertainty samples and low weights to low-uncertainty samples. 
\textit{Third}, we design an uncertainty-guided curriculum learning technique that aligns samples first at the easier image level and then at the harder instance level progressively, ultimately leading to robust cross-domain alignment.

The remainder of this paper is organized as follows. We first review related works in Section \uppercase\expandafter{\romannumeral2}. Our proposed uncertainty-aware domain adaptation method is then presented in Section \uppercase\expandafter{\romannumeral3}. Extensive experimental results are presented in Section \uppercase\expandafter{\romannumeral4}, and Section \uppercase\expandafter{\romannumeral5} finally concludes this paper.

\section{Related Works}

\subsection{Object detection}
Existing object detection methods can be broadly divided into two categories: proposal-based and proposal-free. Proposal-free methods consider detection as a bounding box regression problem. For example, YOLO~\cite{redmon2016you} directly predicts detection results by regressing the coordinates of predefined anchors and simultaneously classifying categories. SSD~\cite{liu2016ssd} integrates predictions computed from hierarchical networks with multiple respective fields to deal with instances of different scales. Several extensions~\cite{redmon2017yolo9000,lin2017focal,law2018cornernet,tian2019fcos} have been proposed for more effective proposal-free object detectors. 
Proposal-based methods first generate region proposals and then classify them for final detection results. For example, R-CNN~\cite{girshick2014rich} uses a hierarchical grouping algorithm to extract dense region proposals and then classify these proposals to obtain detection results. Fast R-CNN~\cite{girshick2015fast} speeds up the R-CNN by introducing a region-of-interest (ROI) pooling layer to share features across each proposal. Faster R-CNN~\cite{ren2015faster} proposes a more efficient and accurate region proposal network (RPN) to replace the hierarchical grouping applied in Fast R-CNN~\cite{girshick2015fast}. Several extensions~\cite{dai2016r,lin2017feature,hu2018relation,cai2019cascade,pang2019libra} present more powerful proposal-based detectors for better detection performance. 

Most existing object detection methods require a large amount of labelled training data which often takes time to annotate. This paper presents a domain adaptive detector built upon Faster R-CNN~\cite{ren2015faster} that aims to optimally exploit the training data that are collected and annotated in prior scenes.

\subsection{Domain adaptive detection}
Our work is closely related to the area of knowledge transfer~\cite{nie2017enhancing,jing2016predicting, azzam2020ktransgan} and domain adaptive object detection~\cite{chen2018domain,saito2019strong,zhang2021self,zhuang2020ifan,he2020domain,zhao2020collaborative,zheng2020cross,xu2020exploring}, which aims to learn a well-performed detector on unlabeled target domain without accessing any annotations in target domain. 
Quite a number of domain adaptive detectors have been reported. For example, DA~\cite{chen2018domain} presents a domain adaptive detector based on Faster R-CNN to minimize domain bias via adversarial learning at both image and instance levels. MTOR~\cite{cai2019exploring} minimizes domain discrepancy by integrating object relations into teacher-student consistency regularization. SWDA~\cite{saito2019strong} presents a powerful cross-domain detection method via strongly aligning local similar features and weakly aligning global dissimilar features. IFAN~\cite{zhuang2020ifan} aligns feature distributions at both image and instance levels in a coarse-to-fine manner. ATF~\cite{he2020domain} presents a tri-stream Faster R-CNN to alleviate the collapse risk caused by parameter sharing between source and target domains. CTR~\cite{zhao2020collaborative} trains the region proposal network (RPN) and the object detection classifier via collaborative self-training. GPA~\cite{xu2020cross} aligns graph-induced prototype representations in two stages to minimize domain discrepancy. CRDA~\cite{xu2020exploring} presents an effective categorical regularization module that improves DA~\cite{chen2018domain} and SWDA~\cite{saito2019strong} consistently. 

The aforementioned cross-domain detection methods mainly employ adversarial learning which merely considers alignment of global distributions and may pull well-aligned samples incorrectly. This work instead aims for protecting well-aligned samples by identifying well-aligned and poorly-aligned samples and aligning them separately in different manners.

\begin{figure*}[!ht]
\centering
\includegraphics[width=.98\linewidth]{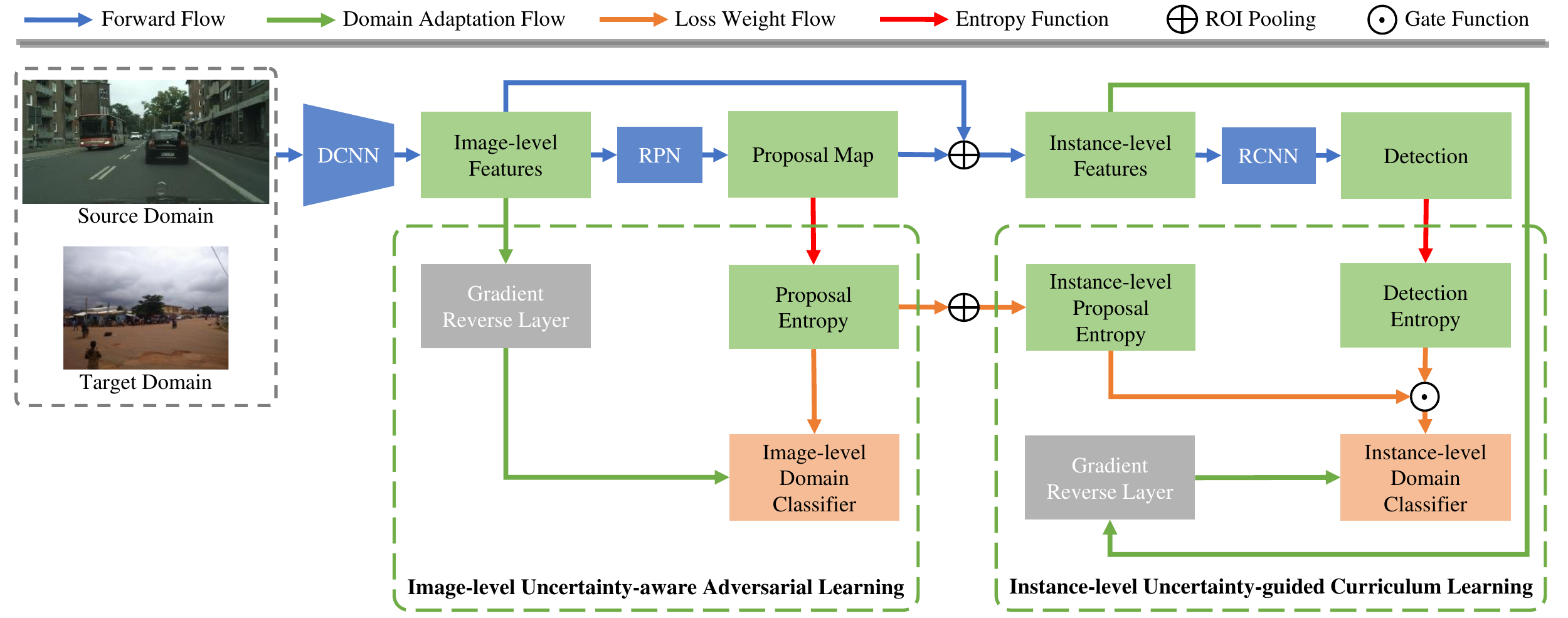}
\caption{The framework of our uncertainty-aware domain adaption network (UaDAN): UaDAN adopts Faster R-CNN architecture with a deep convolutional neural network (DCNN), an RPN and a region convolutional neural network (RCNN). RPN is fed with image-level features from DCNN and generates region proposals. RCNN takes instance-level features from a ROI pooling layer as input and produces the final detection. In image-level adaptation, image-level domain classifier is fed with image-level features from a gradient reversal layer (GRL) where uncertainty-aware adversarial loss is weighted by classification entropy of region proposals. In instance-level adaptation, instance-level domain classifier is fed with instance-level features from the GRL and optimized by our uncertainty-guided curriculum learning (UgCL). With UgCL, detection entropy is filtered by instance-level proposal entropy using a gate function and then fed as the loss weight of instance-level domain classifier. Instance-level proposal entropy is computed from proposal entropy map with an ROI pooling layer.} 
\label{fig:method}
\end{figure*}

\subsection{Curriculum learning}
Curriculum learning has been widely explored in the past decade~\cite{bengio2009curriculum,kumar2010self,khan2011humans}. Bengio \textit{et al.}~\cite{bengio2009curriculum} argues that the generalization of supervised networks could be gradually increased via training from easy examples to harder ones. Kumar \textit{et al.}~\cite{kumar2010self} determines the order of training samples based on the context of non-convex optimization.
Recently, curriculum learning has been widely applied in supervised learning~\cite{pentina2015curriculum,sarafianos2018curriculum,matiisen2019teacher, huang2020curricularface} and semi-/weakly-supervised learning~\cite{gong2016multi,guo2018curriculumnet,zhang2019leveraging,chang2020transductive,yu2020multi}. In the context of cross-domain adaptation, Zhang \textit{et al.}~\cite{zhang2017curriculum} minimizes the domain bias in semantic segmentation via a curriculum-style learning method, where easy tasks are solved first for inferring necessary target-domain properties. Zheng \textit{et al.}~\cite{zheng2020cross} coarsely aligns foreground regions in feature space and finely aligns the class prototype distance.

Our work builds on top of the curriculum learning idea by creating a curriculum over subtasks that a domain adaptive algorithm needs to solve. Different from existing works, it introduces uncertainty awareness into curriculum learning to alleviate the side effect of adversarial learning in the harder instance-level alignment subtask.

\section{Method}

This work focuses on domain adaptive object detection that consists of two sub-tasks in proposal generation and instance detection. We design an uncertainty-aware domain adaptation technique that introduces uncertainty-aware adversarial learning and uncertainty-guided curriculum learning for aligning image-level and instance-level features adaptively and progressively as illustrated in Fig.~\ref{fig:method}, more details to be described in the following subsections.

\subsection{Problem Definition}
We consider a set of source images $X_{s} \in \mathbb{R}^{H \times W \times 3}$ with object labels $\hat{Y}_{s} = \{\hat{C}_{s}, \hat{B}_{s}\}$, where $\hat{C}_{s}\in (1,{C})^{N}$ and $\hat{B}_{s}\in (1,{C})^{N \times 4}$ represent object categories ${C}$ and bounding box coordinates respectively, and a set of target images $X_{t} \in \mathbb{R}^{H \times W \times 3}$ without labels. Here, $H$, $W$, $N$ denote image height, image width, and the class number of objects, respectively. The goal of domain adaptive object detection is to learn a well-performed detector $G$ with access of $\{X_{s}, \hat{Y}_{s}, X_{t}\}$.

Motivated by domain adaptation theory in~\cite{ben2010theory}, recent domain adaptive detection methods~\cite{chen2018domain,saito2019strong,zhang2021self,zhuang2020ifan,he2020domain,zhao2020collaborative,zheng2020cross,xu2020exploring} adopt Faster R-CNN as backbone and employ traditional adversarial learning (TDA) to minimize the cross-domain discrepancy at image and instance levels. They usually exploit $G$ to distill knowledge from $\{X_{s},\hat{Y}_{s}\}$ by minimizing a supervised loss $\mathcal{L}_{det}$, and learn domain-invariant representations via a minimax game between $G$ and domain classifiers ($C_{img}$ and $C_{ins}$) under adversarial learning losses ($\mathcal{L}_{img}$ and $\mathcal{L}_{ins}$). The objective function of TDA in object detection is thus a combination of the three losses:
\begin{equation}
\begin{split}
\mathcal{L}_{TDA} = \mathcal{L}_{det}(F) + \mathcal{L}_{img}(C_{img}) + \mathcal{L}_{ins}(C_{ins}),
\end{split}
\label{eq1}
\end{equation}

Under the guidance of adversarial losses $\mathcal{L}_{img}$ and $\mathcal{L}_{ins}$ in Eq.~\ref{eq1}, TDA minimizes cross-domain discrepancy by aligning features at image and instance levels directly. However, such brute-force inter-domain alignment may introduce misalignment. Specifically, TDA may misalign well-aligned samples to incorrect categories as it assigns adversarial loss of the same weight to all samples. We define this issue as a global alignment problem, and design conditional adversarial losses to identify and align well-aligned and poorly-aligned samples separately in different manners. 

\subsection{Network Architecture}

As illustrated in Fig.~\ref{fig:method}, the proposed cross-domain object detection network $G$ adopts Faster R-CNN as the backbone. The network $G$ consists of a DCNN, an RPN and an RCNN that focus on deep feature extraction, region proposal generation and bounding box detection, respectively. RPN is fed with image-level features from DCNN to predict region category (\textit{i.e.}, 0 for background and 1 for foreground) and generate candidate bounding boxes via dense anchors. RCNN is fed with instance-level features from a ROI pooling layer to classify the candidate bounding boxes generated by RPN into pre-defined $C$ classes and refine their coordinates. 

Source images $X_{s}$ are fed to the detection network for optimizing $G$ via a supervised loss with $Y_{s}$, and simultaneously produce source features (\textit{i.e.}, $F_{img_{s}}\in\mathbb{R}^{U \times V \times D} $ at image level and $F_{ins_{s}}\in \mathbb{R}^{M \times M \times K}$ at instance level). Here, $\{U, V, D\}$ and $\{M, M, K\}$ denote the three dimensions (\textit{i.e.}, height, width and depth) of the image-level and instance-level features, respectively.
Target images $X_{t}$ are fed into the network $G$ to generate target features (\textit{i.e.}, $F_{img_{t}}\in\mathbb{R}^{U \times V \times D}$ at image level and $F_{ins_{t}} \in\mathbb{R}^{M \times M \times K}$ at instance level) where domain classifiers learn how target features map to source features via image-level and instance-level domain adaptation modules. We adopt the GRL~\cite{ganin2015unsupervised} for extracting domain-invariant features via adversarial learning. In implementation, GRL lying between the detection network $G$ and each domain classifier (\textit{i.e.,}, $C_{img}$ at the image level or $C_{ins}$ at the instance level) works by reversing gradients from each domain classifier to the $G$ (in back-propagation). $G$ thus receives the reversed gradients and updates its parameters in an opposite direction, and this guides $G$ to generate domain-invariant features for deceiving the domain classifiers. We followed ~\cite{chen2018domain} to construct the image-level and instance-level domain classifiers. Specifically, the image-level domain classifier has two convolutional layers whose dimensions are $1\times{}1\times{}1024\times{}512$ and $1\times{}1\times{}512\times{}1$, respectively. The instance-level domain classifier has three fully-connected layers, where the first two layers have 1024 channels each and the third layer contains 1 channel. To avoid overfitting in the instance-level domain classifier, the dropout rate is set to $0.5$ for the first two fully-connected layers.

In image-level domain adaptation, image-level domain classifier $C_{img}$ is fed with image-level features ($F_{img_{s}}$ and $F_{img_{t}}$) from GRL and optimized via an image-level uncertainty-aware adversarial loss $\mathcal{L}_{img}^{ua}$ (weighted by proposal entropy). The proposal entropy map (${E}_{P_{s}}\in\mathbb{R}^{U \times V}$ and ${E}_{P_{t}}\in\mathbb{R}^{U \times V}$) is computed from proposal maps ($P_{s}\in\mathbb{R}^{U \times V \times R}$ and $P_{t}\in\mathbb{R}^{U \times V \times R}$) in both source and target domains. Here, $R$ denote the anchors of different scales and ratios at each location $(u, v)$. With the uncertainty-aware adversarial loss, our network will focus on aligning under-aligned features with high entropy while keeping well-aligned features with low entropy less affected. 

In the instance-level domain adaptation, instance-level domain classifier $C_{ins}$ is fed with instance-level features ($F_{ins_{s}}$ and $F_{ins_{t}}$) from the GRL and optimized by instance-level uncertainty-guided curriculum loss $\mathcal{L}_{ins}^{ug}$ (weighted by the filtered detection entropy). The filtered detection entropy is computed by a gate function $\mathcal{G}$ whose input is detection entropy (${E}_{D_{s}}\in\mathbb{R}^{K}$ and ${E}_{D_{t}}\in\mathbb{R}^{K}$) and truncation parameter is instance-level proposal entropy (${E}_{ins_{s}}\in\mathbb{R}^{C,K}$ and ${E}_{ins_{t}}\in\mathbb{R}^{C,K}$). We apply an ROI pooling layer to generate instance-level proposal entropy (${E}_{ins_{s}}$ and ${E}_{ins_{t}}$) from proposal entropy (${E}_{P_{s}}$ and ${E}_{P_{t}}$). With the uncertainty-guided curriculum loss, instance-level alignment can only be activated when the instance-level proposal entropy is lower than a given truncation threshold. It can ensure more robust alignment by first aligning features at easier image level and then aligning features at harder instance level progressively. The symbols used in this paper are listed in Table~\ref{tab:symbol}.

\begin{table}[!h]
\renewcommand{\arraystretch}{1.3}
\centering
\caption{Glossary of symbols.}
\begin{tabular}{p{1cm}|p{6.5cm}}
 \hline
 Symbols &Description\\[0.05cm]
 \hline
  $X_{s}$ &The set of source images  \\
  $\hat{Y}_{s}$ &The set of source labels  \\
  $X_{t}$ &The set of target images  \\
  $G$ &The object detection network  \\
  $C_{img} $ &The image-level domain classifier  \\
  $C_{ins} $ &The instance-level domain classifier   \\
  $P_{s}$ &The region proposals in source domain  \\
  $P_{t}$ &The region proposals in target domain  \\
  $D_{s}$ &The object detection in source domain  \\
  $D_{t}$ &The object detection in target domain  \\
  $F_{img_{s}}$ &The image-level features in source domain  \\
  $F_{img_{t}}$ &The image-level features in target domain \\
  $F_{ins_{s}}$ &The instance-level features in source domain \\
  $F_{ins_{t}}$ &The instance-level features in target domain \\
  ${E}_{P_{s}}$ &The proposal entropy in source domain \\
  ${E}_{P_{t}}$ &The proposal entropy in target domain \\
  $E_{ins_{s}}$ &The instance-level proposal entropy in source domain \\
  $E_{ins_{t}}$ &The instance-level proposal entropy in target domain \\
  ${E}_{D_{s}}$ &The detection entropy in source domain \\
  ${E}_{D_{t}}$ &The detection entropy in target domain \\
  $L_{s}$ &The source domain labels \\
  $L_{t}$ &The target domain labels \\
  $O_{img_{s}}$ &The image-level source domain predictions  \\
  $O_{img_{t}}$ &The image-level target domain predictions \\
  $O_{ins_{s}}$ &The instance-level source domain predictions  \\
  $O_{ins_{t}}$ &The instance-level target domain predictions \\
  $\xi$ &The hyper-parameter to activate instance-level alignment \\
\hline
\end{tabular}
\label{tab:symbol}
\end{table}

\subsection{Training Objective}

The network $G$ is optimized with three loss terms, \textit{i.e.}, a supervised detection loss $\mathcal{L}_{det}$ for distilling knowledge in source domain, an image-level uncertainty-aware adversarial loss $\mathcal{L}_{img}^{ua}$ and an instance-level uncertainty-guided adversarial loss $\mathcal{L}_{ins}^{ug}$ for unsupervised domain adaptation. In the forward processing, we fed the network $G$ with a pair of source and target images ($x_{s}$ and $x_{t}$) and obtain region proposals ($p_{s}$ and $p_{t}$) from RPN, detection results ($d_{s}$ and $d_{t}$) from RCNN,  image-level features ($f_{img_{s}}$ and $f_{img_{t}}$) from DCNN, and instance-level features ($f_{ins_{s}}$ and $f_{ins_{t}}$) from an ROI pooling layer which extract local information from image-level features in line with the proposal. Our training objective is to maximize mixed likelihood of source and target features.

\subsubsection{\textbf{Source-domain supervised learning}}
For source-domain supervised learning, we aim to build a relationship between the source-domain outputs (\textit{i.e.}, region proposals $p_{s}$ and object defections $d_{s}$) and annotations $\hat{y}_{s}$. Given a source-domain image $x_{s}$ and the network $G$, the supervised detection loss is defined as:
\begin{equation}
\begin{split}
    \mathcal{L}_{det}(x_{s},\hat{y}_{s};F) = &\mathcal{L}_{rpn}(p_{s},\hat{y}_{s}) + \mathcal{L}_{rcnn}(d_{s},\hat{y}_{s}),
\end{split}
\label{eq2}
\end{equation}
where $p_{s}$ and $d_{s}$ denote the region proposals and object detection generated from $G$ on source domain, $\mathcal{L}_{rpn}$ and $\mathcal{L}_{rcnn}$ represent RPN loss term and RCNN loss term respectively, as defined in~\cite{ren2015faster}.

\subsubsection{\textbf{Image-level uncertainty-aware adversarial learning}}

For first-stage image-level domain adaptation, we propose an uncertainty-aware adversarial learning method to align the image-level features in source and target domains. Given image-level source and target features ($f_{img_{s}}$ and $f_{img_{t}}$), region proposals ($p_{s}$ and $p_{t}$), and the image-level domain classifier ($C_{img}$), the image-level uncertainty-aware adversarial loss is defined as:
\begin{equation}
\begin{split}
    \mathcal{L}^{ua}_{img}(C_{img}) = & \sum_{u,v} [ \mathcal{E}_{p}(p_{s}^{(r,u,v)}) \mathcal{L}^{ce}_{img}(o_{img_{s}}^{(u,v)},l_{s}) + \\ & \mathcal{E}_{p}(p_{t}^{(r,u,v)}) \mathcal{L}^{ce}_{img}(o_{img_{t}}^{(u,v)},l_{t}) ] ,
\end{split}
\label{eq5}
\end{equation}
where $\mathcal{L}^{ce}_{img}$ denotes an image-level cross-entropy loss, $\mathcal{E}_{p}$ represents a proposal entropy function, $p^{(r,u,v)}$ is the $r$-th classification output located at $(u,v)$ of region proposal prediction maps, $r$ denotes the index of proposals with different scales and ratios in the same location, $o_{img_{s}}^{(u,v)}$ and $o_{img_{t}}^{(u,v)}$ are the activation located at $(u,v)$ of the domain prediction maps generated from $C_{img}$ in each domain, $l_{s}^{(u,v)}$ and $l_{t}^{(u,v)}$ are the pixel-level domain label locate at $(u,v)$ of image-level domain labels (\textit{i.e.}, 0 for source domain and 1 for target domain).
The image-level cross-entropy loss $\mathcal{L}^{ce}_{img}$ in Eq.~\ref{eq5} is defined as:
\begin{equation}
\begin{split}
    \mathcal{L}^{ce}_{img}(o^{(u,v)},l^{(u,v)}) = & -l^{(u,v)}{\rm log}(o^{(u,v)}) - \\ & (1-l^{(u,v)}){\rm log}{(1-o^{(u,v)})} ,
\end{split}
\label{eq6}
\end{equation}

For estimating the prediction confidence, we choose the lowest entropy of proposals in each location. Following information theory~\cite{shannon1948mathematical}, the proposal entropy function $\mathcal{E}_{p}$ in Eq.~\ref{eq5} is defined as:
\begin{equation}
\begin{split}
    \mathcal{E}_{p}(p^{(r,u,v)}) = & \min_{r} [ - p^{(r,u,v)}\cdot{}{\rm log}(p^{(r,u,v)})  \\ & - (1-p^{(r,u,v)})(1-{\rm log}(p^{(r,u,v)}))].
\end{split}
\label{eq7}
\end{equation}

\subsubsection{\textbf{Instance-level uncertainty-guided curriculum learning}}

For second-stage instance-level domain adaptation, we introduce an uncertainty-guided curriculum learning approach for more robust domain alignment. Given instance-level source-domain and target-domain features ($f_{ins_{s}}$ and $f_{ins_{t}}$), detection results ($d_{s}$ and $d_{t}$), instance-level proposal entropy ($e_{ins_{s}}$ and $e_{ins_{t}}$) generated from proposal entropy using an ROI layer, and the instance-level domain classifier $C_{ins}$, the instance-level uncertainty-guided curriculum loss is defined as:

\begin{equation}
\begin{split}
    \mathcal{L}^{ug}_{ins}(C_{ins})
     =\sum_{k} & [ \mathcal{G}(\mathcal{E}_{d}(d_{s}^{(c,k)}),e_{ins_{s}}^{(k)})) \mathcal{L}^{ce}_{ins}(o_{ins_{s}}^{(k)},l_{s}^{(k)}) + \\ & \mathcal{G}(\mathcal{E}_{d}(d_{t}^{(c,k)}),e_{ins_{t}}^{(k)})) \mathcal{L}^{ce}_{ins}(o_{ins_{t}}^{(k)},l_{t}^{(k)}) ] ,
\end{split}
\label{eq8}
\end{equation}

where $\mathcal{L}^{ce}_{ins}$ denotes an instance-level cross-entropy loss, $\mathcal{G}$ represents a gate function, $\mathcal{E}_{d}$ represents a detection entropy function,  $d^{(c,k)}$ represents the predicted probability of $c$-th class in $k$-th detection, $e^{(k)}$ is instance-level proposal entropy corresponding to $k$-th detection, $o_{ins_{s}}^{(k)}$ and $o_{ins_{t}}^{(k)}$ are the activation located at $(k)$ of instance-level domain prediction vector generated from $C_{ins}$ in each domain, $l^{(k)}$ is the instance-level domain label (\textit{i.e.}, 0 for source domain and 1 for target domain) for $k$-th detection.

The instance-level cross-entropy loss $\mathcal{L}^{ce}_{ins}$ is defined by:
\begin{equation}
\begin{split}
    \mathcal{L}^{ce}_{ins}(o^{(k)},l^{(k)}) = & -l^{(k)}{\rm log}(o^{(k)}) - \\ 
    & (1-l^{(k)}){\rm log}(1-o^{(k)})).
\end{split}
\label{eq9}
\end{equation}

The gate function $\mathcal{G}$ in Eq.~\ref{eq8} is defined as:
\begin{equation}
\mathcal{G}(\mathcal{E}_{d}(d^{(c,k)}),{e}_{ins}^{k})=\left\{
\begin{aligned}
& \mathcal{E}_{d}(d^{(c,k)}) & {e}_{ins}^{k} < \xi \\
& 0 & others
\end{aligned}
\right.
\label{eq10}
\end{equation}
where $\xi$ is a hyper-parameter to estimate whether image-level features are well-aligned or not. The instance-level adversarial learning can only be activated after image-level features are well-aligned, \textit{i.e.}, the prediction entropy of proposals is low. 

Following information theory~\cite{shannon1948mathematical}, the detection entropy function $\mathcal{E}_{d}$ in Eq.~\ref{eq8} is defined as:
\begin{equation}
\begin{split}
    \mathcal{E}_{d}(d^{(c,k)}) = & - \sum_{c}d^{(c,k)}\cdot{}{\rm log}d^{(c,k)}.
\end{split}
\label{eq11}
\end{equation}

\subsubsection{\textbf{Optimization.}}
The overall loss of UaDAN is:
\begin{equation}
\begin{split}
\mathcal{L}_{UaDAN} = \mathcal{L}_{det}(F) + \mathcal{L}^{ua}_{img}(C_{img}) + \mathcal{L}^{ug}_{ins}(C_{ins}).
\end{split}
\label{eq12}
\end{equation}

The training objective of UaDAN is:
\begin{equation}
\begin{split}
F^{\star}, C^{\star}_{img}, C^{\star}_{ins} = \arg\min_{F}\max_{C_{img}}\max_{C_{ins}} \mathcal{L}_{UaDAN} 
\end{split}
\label{eq13}
\end{equation}

We solve Eq.~\ref{eq13} by simultaneously optimizing $G$, $C_{img}$ and $C_{ins}$ until $\mathcal{L}_{UaDAN}$ converges.

\subsection{Analysis}

The major differences between our uncertainty-aware domain adaptation and traditional adversarial learning lie with uncertainty-aware adversarial learning and uncertainty-guided curriculum learning. We focus on instance-level domain adaptation to discuss the differences in this subsection. 

Given instance-level source and target features ($f_{ins_{s}}$ and $f_{ins_{t}}$) and the instance-level domain classifier $C_{ins}$, the instance-level traditional adversarial loss is defined as:
\begin{equation}
\begin{split}
    \mathcal{L}_{ins}(C_{ins})
     =\sum_{k} & [\mathcal{L}^{ce}_{ins}(o_{ins_{s}}^{(k)},l_{s}^{(k)}) + \mathcal{L}^{ce}_{ins}(o_{ins_{t}}^{(k)},l_{t}^{(k)}) ] ,
\end{split}
\label{eq14}
\end{equation}
and the instance-level uncertainty-aware adversarial learning loss without curriculum learning is defined as:
\begin{equation}
\begin{split}
    \mathcal{L}^{ua}_{ins}(C_{ins}) =\sum_{k} & [\mathcal{E}_{d}(d_{s}) \mathcal{L}^{ce}_{ins}(o_{ins_{s}}^{(k)},l_{s}^{(k)}) + \\ & \mathcal{E}_{d}(d_{t}) \mathcal{L}^{ce}_{ins}(o_{ins_{t}}^{(k)},l_{t}^{(k)}) ] .
\end{split}
\label{eq15}
\end{equation}

The differences between $\mathcal{L}_{ins}$ in Eq.~\ref{eq14} and $\mathcal{L}^{ua}_{ins}$ in Eq.~\ref{eq15} is that $\mathcal{L}_{ins}$ has the same weight while $\mathcal{L}^{ua}_{ins}$'s weight is decided by the prediction entropy. As studied in~\cite{grandvalet2005semi,lee2007learning,vu2019advent}, well-aligned features usually produce confident predictions with low-entropy while under-aligned features often produce unconfident predictions with high-entropy. Utilizing entropy to adjust the loss weights can naturally protect well-aligned features from large re-alignment and focus more on aligning under-aligned features. Our uncertainty-aware adversarial learning is thus able to decrease the influence of domain discrepancy minimization on well-aligned features and lead to stable cross-domain alignment in each subtask. 

We provide certain theoretical analysis with a negative transfer scenario that often happens in unsupervised domain adaptation~\cite{pan2009survey,wang2019characterizing} when the source-domain knowledge hurts the target-domain performance. Unsupervised domain adaptation aims to improve a predictive function $G$ over unlabeled target domain $D_{t}$ by learning transferable knowledge in labeled source domain $D_{s}$. We use $P_{s}(X,Y)$ and $P_{t}(X,Y)$ to denote the joint distribution in $D_{s}$ and $D_{t}$, respectively, where $X$ denotes input data and $Y$ denotes labels. The objective of the traditional adversarial learning assumes that for any $x_{t}\in X_{t}$, there exists $x_{s}\in X_{s}$ such that $P_{t}(x_{t},y_{t})=P_{s}(x_{s},y_{s})$~\cite{ganin2016domain}. However, $P_{t}(X,Y)$ and $P_{s}(X,Y)$ naturally have discrepancy between them. Considering the case with a well-classified (semantically well-aligned) target sample $x'_{t}\in X_{t}$ with domain specific features such that $P_{t}(x'_{t},y'_{t})\neq P_{s}(x_{s},y_{t})$ for any $x_{s}\in X_{s}$. If $x'_{t}$ is trained with the objective of traditional adversarial learning that assumes $P_{t}(x_{t},y_{t})=P_{s}(x_{s},y_{s})$, it will be misaligned to incorrect classes due to negative transfer~\cite{wang2019characterizing}, leading to $P_{t}(x'_{t},y'_{t}) \not\in P_{t}(X,Y)$. Our uncertainty-aware adversarial learning can instead mitigate such negative transfer by adjusting the loss weight of the well-aligned sample $x'_{t}$ (far from classifier boundary with entropy close to zero) using its entropy so that the well-aligned $x'_{t}$ will not be further aligned. This largely helps to ensure $P_{t}(x'_{t},y'_{t}) \in P_{t}(X,Y)$ and mitigate negative transfer effectively.

Compared with $\mathcal{L}_{ins}$ and $\mathcal{L}^{ua}_{ins}$, uncertainty-guided curriculum loss $\mathcal{L}^{ug}_{ins}$ in Eq.~\ref{eq8} further introduces a gate function $\mathcal{G}$ to activate instance-level alignment when the corresponding image-level representations are well-aligned (\textit{i.e.}, low entropy). At early phase of training, source and target predictions are inaccurate and source knowledge is steadily transferred to target domain with the supervision of source ground-truth. Intuitively, aligning nonsensical instances could easily impair semantic structures and make negative impact. Harder subtask can make positive impact only when the easier subtask becomes relatively stable. Our developed uncertainty-guided curriculum learning solves the image-level alignment (easier subtask) and instance-level alignment (harder subtask) progressively to ensure more stable cross-domain alignment.

\section{Experiments}

\begin{table*}[ht]
\renewcommand{\arraystretch}{1.3}
\caption{Quantitative comparison of UaDAN with state-of-the-art domain adaptive object detection methods over the cross camera adaptation task Cityscapes $\rightarrow$ Mapillary Vistas: AP (\%) of each category and mAP (\%) of all classes are evaluated over the Mapillary Vistas validation set.
}
\centering
\begin{tabular}{p{3cm}|*{8}{p{1cm}}|p{1cm}}
 \hline
 Methods &{person} &{rider} &{car} &{truck} &{bus} &{train} &{motorbike} &{bicycle} &mAP\\[0.05cm]
 \hline
 Source only &31.3  &30.0  &48.9  &20.3  &23.1  &7.1  &21.2  &24.3  &25.8  \\
 DA~\cite{chen2018domain}  &34.2  &28.8  &55.3  &20.0  &19.5  &17.8  &23.9  &28.2  &28.4   \\
 SWDA~\cite{saito2019strong} &33.4  &29.3  &50.9  &23.4  &26.6  &23.8  &28.2  &25.0  &30.1   \\
 CRDA~\cite{xu2020exploring} &34.0  &\textbf{32.6}  &51.4  &23.3  &24.0  &22.4  &28.2  &27.0  &30.4   \\
 CFA~\cite{zheng2020cross} &34.4 &30.0 &54.6 &22.7 &24.9 &21.9 &26.7 &26.7 &30.6 \\
 GPA~\cite{xu2020cross} &35.9  &31.1  &\textbf{55.9}  &20.1  &23.1  &25.6  &28.2  &28.6  &31.0   \\
 \textbf{UaDAN (Ours)} &\textbf{36.1}  &31.1  &55.7 &\textbf{24.3}  &\textbf{27.0}  &\textbf{28.3 } &\textbf{30.1}  &\textbf{28.9}  &\textbf{32.7}   \\
\hline
\end{tabular}
\label{tab:bench1}
\end{table*}

\begin{table*}[ht]
\renewcommand{\arraystretch}{1.3}
\caption{Quantitative comparison of UaDAN with state-of-the-art domain adaptive object detection methods over the weather adaptation task Cityscapes $\rightarrow$ Foggy Cityscapes: AP (\%) of each category and mAP (\%) of all classes are evaluated over the Foggy Cityscapes validation set.
}
\centering
\begin{tabular}{p{3cm}|*{8}{p{1cm}}|p{1cm}}
 \hline
 Methods &{person} &{rider} &{car} &{truck} &{bus} &{train} &{motorbike} &{bicycle} &mAP\\[0.05cm]
 \hline
 Source only &26.9 &26.9 &38.2 &18.3 &32.4 &9.6 &25.8 &28.6 &26.9  \\
 DA~\cite{chen2018domain}  &29.2 &40.4 &43.4 &19.7 &38.3 &28.5 &23.7 &32.7 &32.0  \\
 SWDA~\cite{saito2019strong} &31.8 &44.3 &48.9 &21.0 &43.8 &28.0 &28.9 &35.8 &35.3  \\
 CRDA~\cite{xu2020exploring} &32.2 &45.2 &50.0 &\textbf{30.3} &48.1 &36.3 &28.4 &36.8 &38.4  \\
 CFA~\cite{zheng2020cross} &\textbf{37.4} &45.3 &53.5 &25.8 &\textbf{50.5} &31.3 &30.2 &\textbf{39.2} &39.1 \\
 GPA~\cite{xu2020cross} &32.9 &\textbf{46.7} &\textbf{54.1} &24.7 &45.7 &41.1 &\textbf{32.4} &38.7 &39.5  \\
 \textbf{UaDAN (Ours)} &{36.5} &46.1 &53.6 &28.9 &{49.4} &\textbf{42.7} &32.3 &{38.9} &\textbf{41.1} \\
\hline
\end{tabular}
\label{tab:bench2}
\end{table*}

\subsection{Experimental Setup}
We follow the widely adopted protocol in domain adaptive object detection. Each task involves two datasets including a source dataset and a target dataset for training and evaluation. The training data consist of the labelled source training set and the unlabelled target training set. The validation set of the target datasets is used for evaluations in all methods. For the source and target datasets, only data with shared object categories are used in training and evaluations.

\subsubsection{\textbf{Dataset}}
Our experiments involves six public datasets including Mapillary Vistas~\cite{neuhold2017mapillary}, Cityscapes~\cite{cordts2016cityscapes}, Foggy Cityscapes~\cite{sakaridis2018semantic}, PASCAL VOC~\cite{everingham2010pascal}, Clipart~\cite{inoue2018cross}, SIM10k~\cite{johnson2017driving}. More details of the six datasets are listed below.

$\bullet$ {Mapillary Vistas}~\cite{neuhold2017mapillary} is a large-scale autonomous driving dataset with images recorded by different acquisition sensors. It contains 18,000 training images and 2,000 validation images, and the image resolution varies from $768\times1024$ to $4000 \times 6000$. Mapillary Vistas consists of 37 object categories.

$\bullet$ {Cityscapes}~\cite{cordts2016cityscapes} is a widely used autonomous driving dataset with images captured with a unique vehicle mounted image acquisition system. It consists of 2,975 training images and 500 validation images with dense detection annotations, which are transformed from instance segmentation labels with 8 categories. All the images have the same resolution of $1024 \times 2048$ with common weather conditions.

$\bullet$ Foggy Cityscapes~\cite{sakaridis2018semantic} dataset is established by applying fog simulation on the Cityscapes images\cite{cordts2016cityscapes}. The synthetic foggy images are rendered with visible images with common weather conditions and its depth counterpart in Cityscapes. The annotations are inherited from labels in Cityscapes.

$\bullet$ PASCAL VOC~\cite{everingham2010pascal} is a large-scale real world dataset with two sub-datasets, \textit{i.e.}, PASCAL VOC 2007~\cite{everingham2007pascal} and PASCAL VOC 2012~\cite{everingham2015pascal}. PASCAL VOC 2007 contains 2,501 for training and 2,510 for validation, while PASCAL VOC 2012 consists of 5,717 for training and 5,823 for validation. Bounding box annotations with 20 classes are provided.

$\bullet$ Clipart1k~\cite{inoue2018cross} dataset contains 1,000 comical images, in which 800 for training and 200 for validation. The manually created comical images have much dissimilarity as compared with real world images. It provides bounding box annotations with the same 20 categories as PASCAL VOC~\cite{everingham2010pascal}. 

$\bullet$  SIM10k~\cite{johnson2017driving} dataset contains 10,000 synthetic images with automatically generated labels from computer games. All the images have the same resolution of $1052 \times 1914$. It provides bounding box annotations for cars.

\subsubsection{\textbf{Implementation details}}
Following~\cite{chen2018domain,saito2019strong,xu2020exploring}, we adopt Faster R-CNN~\cite{ren2015faster} as our object detection network.
The backbone is initialized with deep convolutional layers of ResNet-50~\cite{he2016deep}, which is pre-trained on ImageNet~\cite{deng2009imagenet}. The detection modules of Faster R-CNN (\textit{i.e.}, RPN and RCNN) and the domain classifiers (\textit{i.e.}, image-level and instance-level) are randomly initialized from a zero-mean Gaussian distribution with a standard deviation 0.01. During training, we use back-propagation and stochastic gradient descent (SGD) to optimize all the networks with a momentum of $0.9$ and a weight decay of $5e-4$. The initial learning rate is set at $0.001$ for $50k$ iterations and then reduced to $0.0001$ for another $20k$ iterations.UaDAN uses one source image and one target image in each iteration as in~\cite{chen2018domain}. For all experiments, the hyper-parameter $\xi$ is set as $0.5$. All experiments are implemented on one GPU by employing PyTorch toolbox, where the maximum memory usage is less than 9 GB. For evaluation, average precision (AP) of each category and mean average precision (mAP) of all categories are computed with an intersection over union (IoU) threshold 0.5.

\begin{table*}[!ht]
\renewcommand{\arraystretch}{1.3}
\caption{Quantitative comparison of UaDAN with state-of-the-art domain adaptive object detection methods over the dissimilar adaptation task PASCAL VOC $\rightarrow$ Clipart1k: AP (\%) of each category and mAP (\%) of all classes are evaluated over the Clipart1k validation set. Note that aero, bicy, bott, motor, pers and tv are abbreviations of aeroplane, bicycle, bottle, motorbike, person and television, respectively.
}
\centering
\begin{tabular}{p{2cm}|*{20}{p{0.3cm}}|p{0.4cm}}
 \hline
 Methods &aero &bicy &bird &boat &bott &bus &car &cat &chair &cow &table &dog &horse &motor &pers &plant &sheep &sofa &train &tv &mAP\\[0.05cm]
 \hline
 Source only  &4.3 &75.0 &31.3 &21.5 &4.8 &64.5 &23.2 &\textbf{6.7} &33.1 &7.1 &32.5 &4.9 &56.5 &66.7 &49.6 &51.9 &9.4 &20.5 &34.5 &21.6 &31.0 \\
 DA~\cite{chen2018domain} &17.9 &\textbf{75.4} &32.9 &23.7 &10.9 &35.8 &31.0 &3.8 &42.2 &70.8 &13.6 &16.9 &31.3 &86.7 &61.7 &53.9 &10.0 &25.2 &29.9 &18.7 &34.6 \\
 SWDA~\cite{saito2019strong} &6.8 &50.2 &33.0 &19.2 &13.3 &61.4 &35.4 &5.9 &\textbf{40.8} &54.5 &\textbf{40.8} &24.6 &53.9 &76.7 &63.6 &61.0 &10.0 &\textbf{34.2} &32.1 &19.9 &36.8  \\
 CFA~\cite{zheng2020cross} &30.5 &54.5 &37.8 &\textbf{27.7} &11.5 &65.5 &44.7 &0.9 &35.4 &51.0 &32.6 &24.8 &35.1 &63.6 &64.4 &57.9 &12.1 &18.4 &46.2 &26.9 &37.1 \\
 CRDA~\cite{xu2020exploring}  &7.7 &50.5 &33.2 &19.3 &13.0 &68.2 &40.0 &3.0 &40.4 &{68.9} &29.9 &20.7 &\textbf{57.2} &\textbf{86.9} &\textbf{68.6} &55.5 &\textbf{15.1} &27.6 &14.8 &\textbf{30.5} &37.5  \\
 GPA~\cite{xu2020cross} &12.2 &54.5 &40.3 &25.6 &16.8 &\textbf{71.3} &39.9 &4.2 &38.9 &\textbf{73.1} &21.6 &25.7 &54.5 &63.6 &63.1 &58.6 &13.6 &11.1 &44.7 &31.6 &38.3 \\
 \textbf{UaDAN (Ours)} &\textbf{35.0} &72.7 &\textbf{41.0} &{24.4} &\textbf{21.3} &{69.8} &\textbf{53.5} &2.3 &34.2 &61.2 &31.0 &\textbf{29.5} &47.9 &63.6 &62.2 &\textbf{61.3} &13.9 &7.6 &\textbf{48.6} &23.9 &\textbf{40.2} \\
\hline
\end{tabular}
\label{tab:bench3}
\end{table*}

\begin{table}[ht]
\renewcommand{\arraystretch}{1.3}
\caption{Quantitative comparison of UaDAN with state-of-the-art domain adaptive object detection methods over the synthetic-to-realistic adaptation task Sim10k $\rightarrow$ Cityscapes: AP (\%) of car class is evaluated on the Cityscapes validation set.
}
\centering
\begin{tabular}{p{3cm}|p{3cm}}
 \hline
 Methods &car AP\\[0.05cm]
 \hline
 Source only &34.6  \\
 DA~\cite{chen2018domain}  &41.9  \\
 SWDA~\cite{saito2019strong} &44.6  \\
 CFA~\cite{zheng2020cross} &46.0 \\
 CRDA~\cite{xu2020exploring} &46.6  \\
 GPA~\cite{xu2020cross} &47.6  \\
 \textbf{UaDAN (Ours)} &\textbf{48.6}  \\
\hline
\end{tabular}
\label{tab:bench4}
\end{table}

\begin{figure*}[!ht]
\centering
\begin{minipage}[h]{0.245\linewidth}
\centering\footnotesize {Source only}
\end{minipage}
\vspace{2pt}
\begin{minipage}[h]{0.245\linewidth}
\centering\footnotesize {GPA~\cite{xu2020cross}} 
\end{minipage}
\begin{minipage}[h]{0.245\linewidth}
\centering\footnotesize {\textbf{UaDAN (Ours)}}
\end{minipage}
\begin{minipage}[h]{0.245\linewidth}
\centering\footnotesize {Ground Truth} 
\end{minipage}
\vspace{2pt}
\centering
\begin{minipage}[h]{0.245\linewidth}
\centering\includegraphics[width=.99\linewidth]{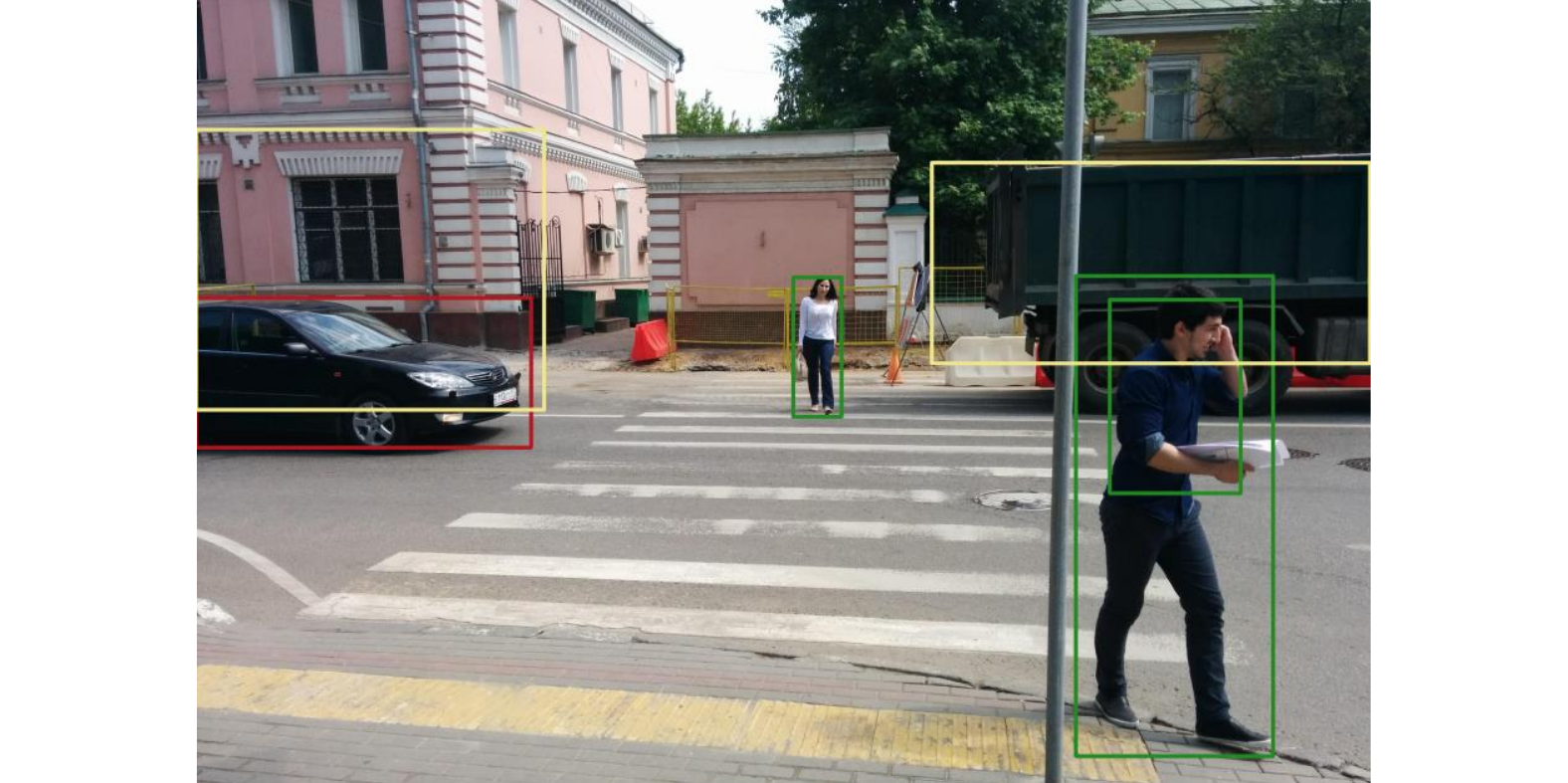}
\end{minipage}
\begin{minipage}[h]{0.245\linewidth}
\centering\includegraphics[width=.99\linewidth]{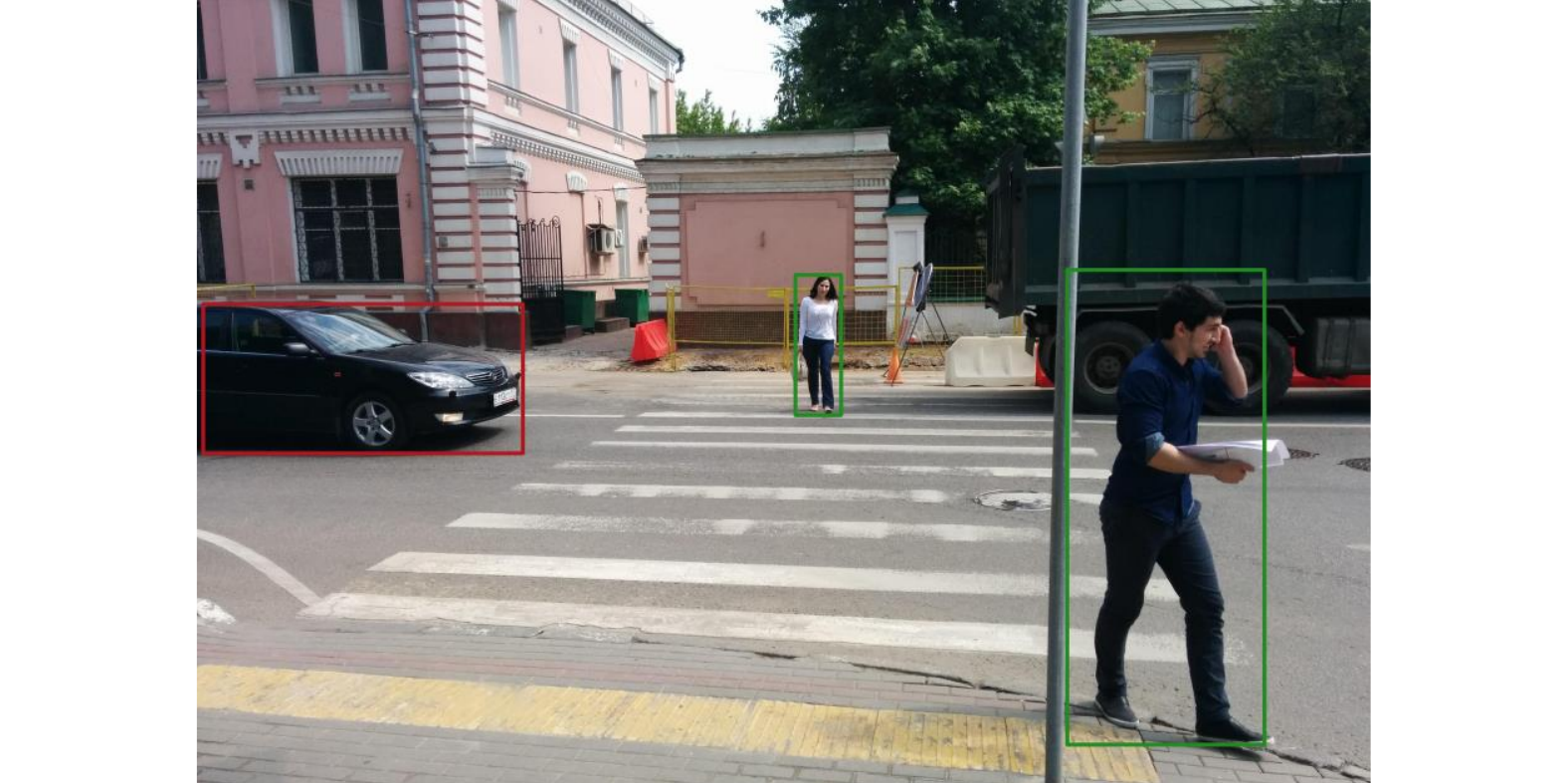}
\end{minipage}
\begin{minipage}[h]{0.245\linewidth}
\centering\includegraphics[width=.99\linewidth]{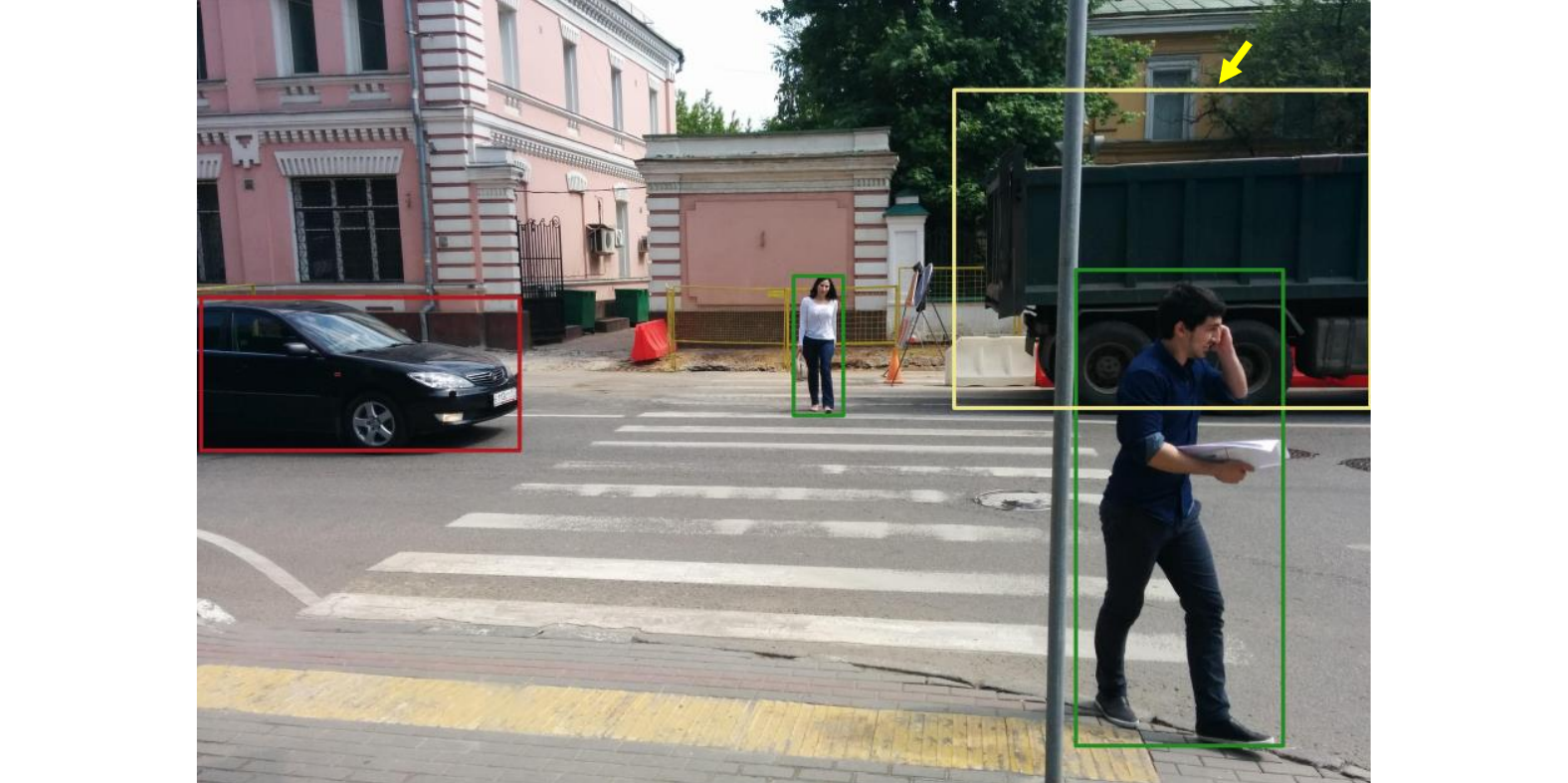}
\end{minipage}
\begin{minipage}[h]{0.245\linewidth}
\centering\includegraphics[width=.99\linewidth]{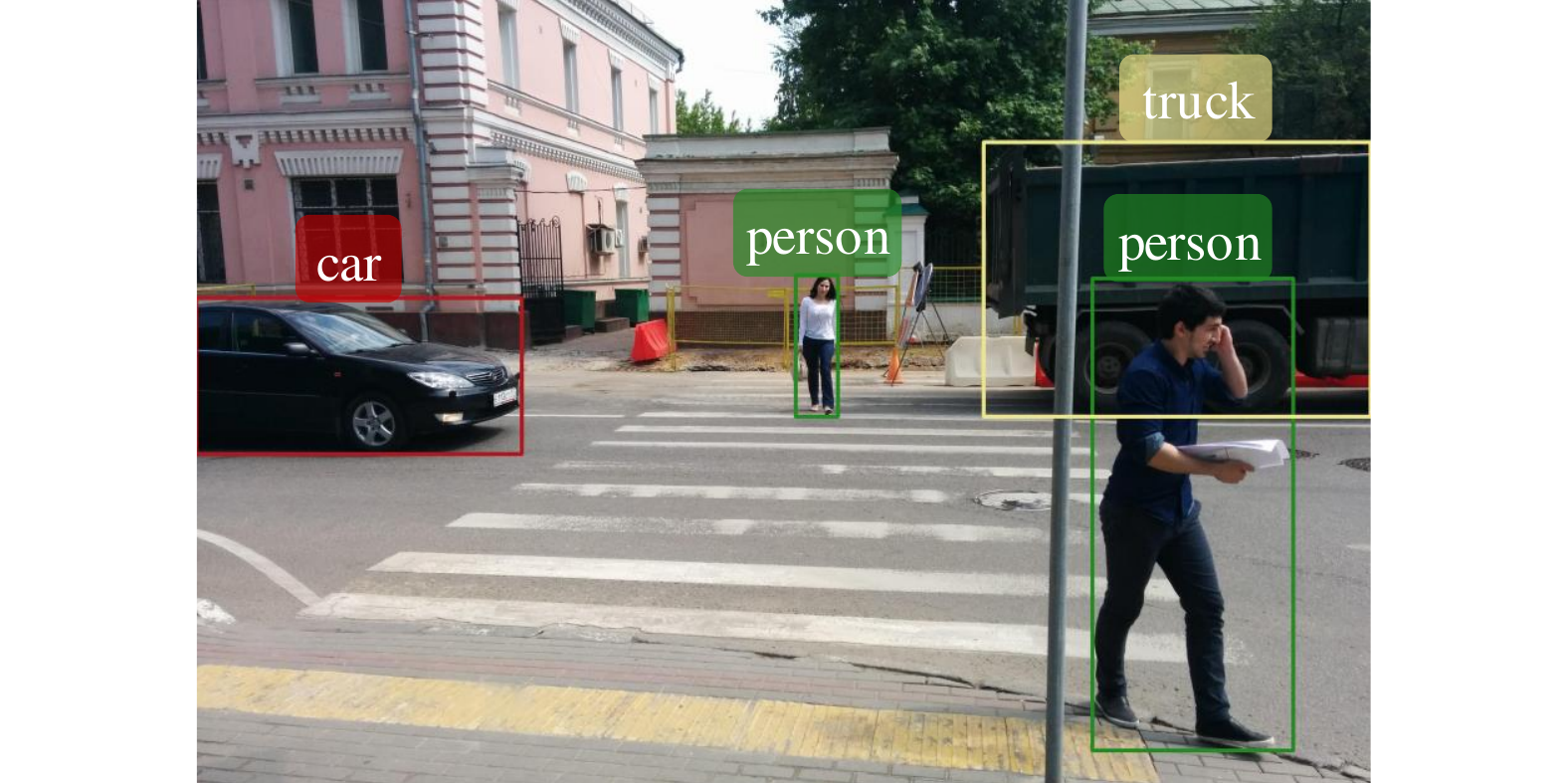}
\end{minipage}
\centering
\vspace{2pt}
\begin{minipage}[h]{0.245\linewidth}
\centering\includegraphics[width=.99\linewidth]{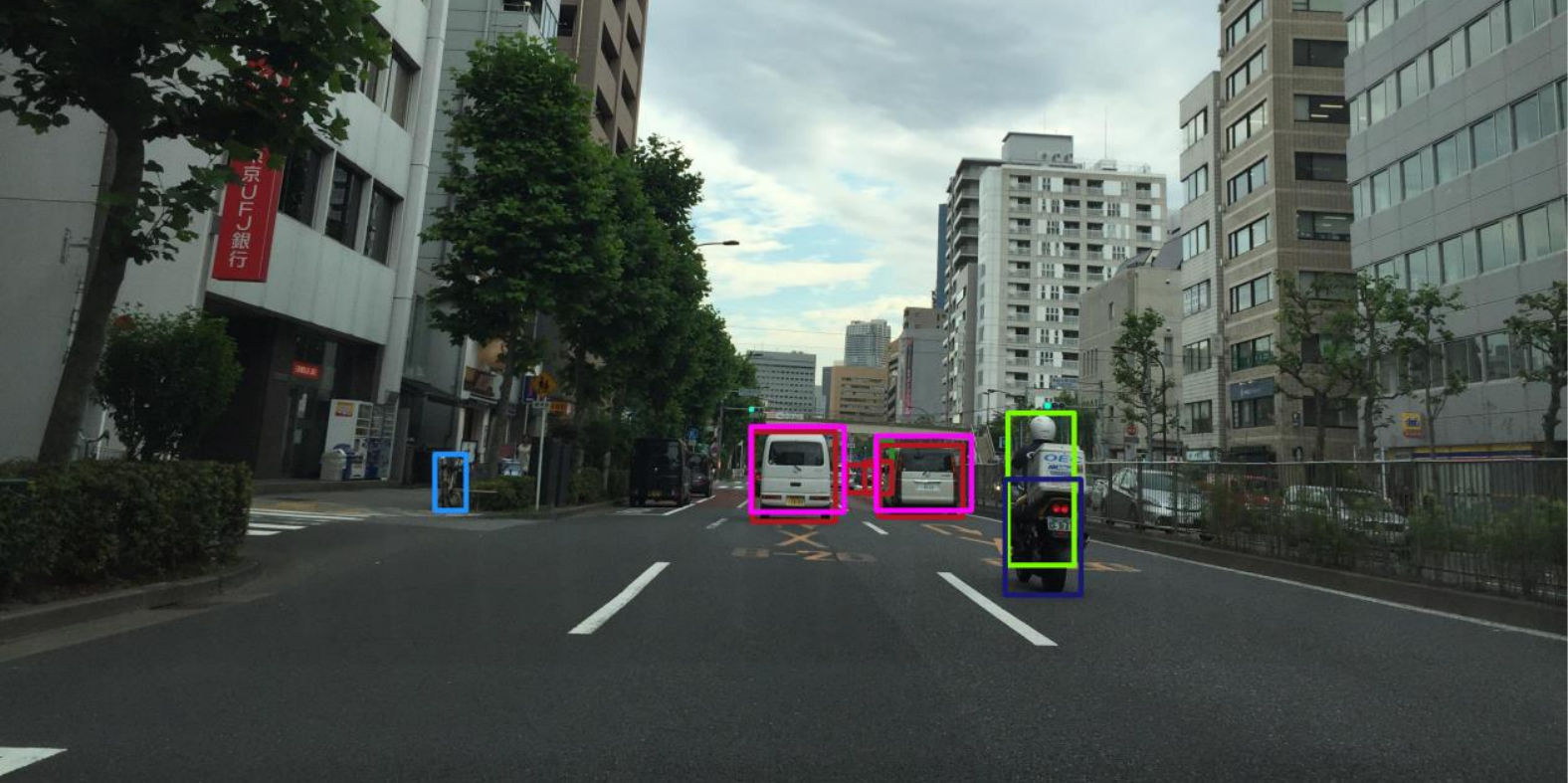}
\end{minipage}
\begin{minipage}[h]{0.245\linewidth}
\centering\includegraphics[width=.99\linewidth]{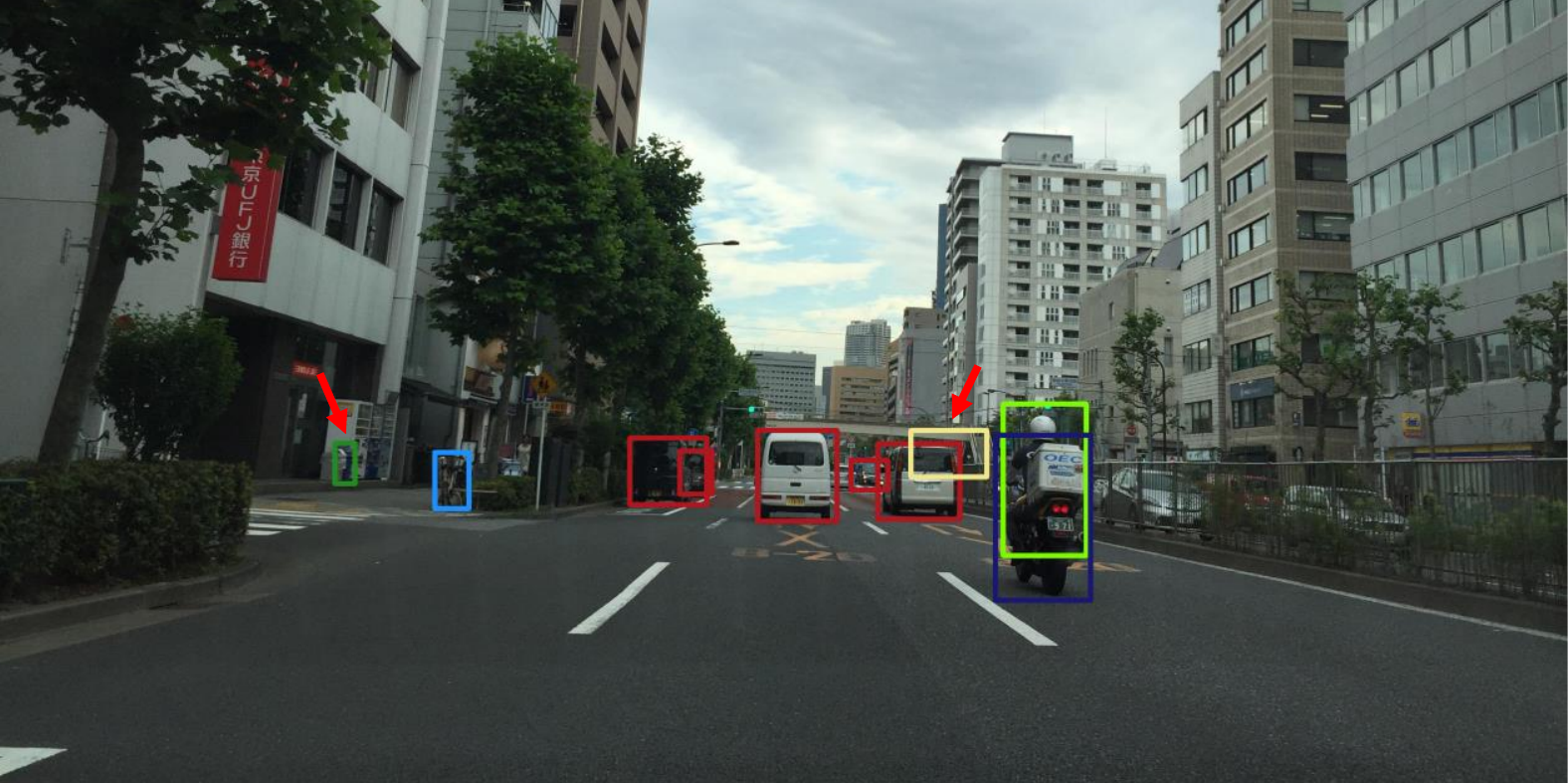}
\end{minipage}
\begin{minipage}[h]{0.245\linewidth}
\centering\includegraphics[width=.99\linewidth]{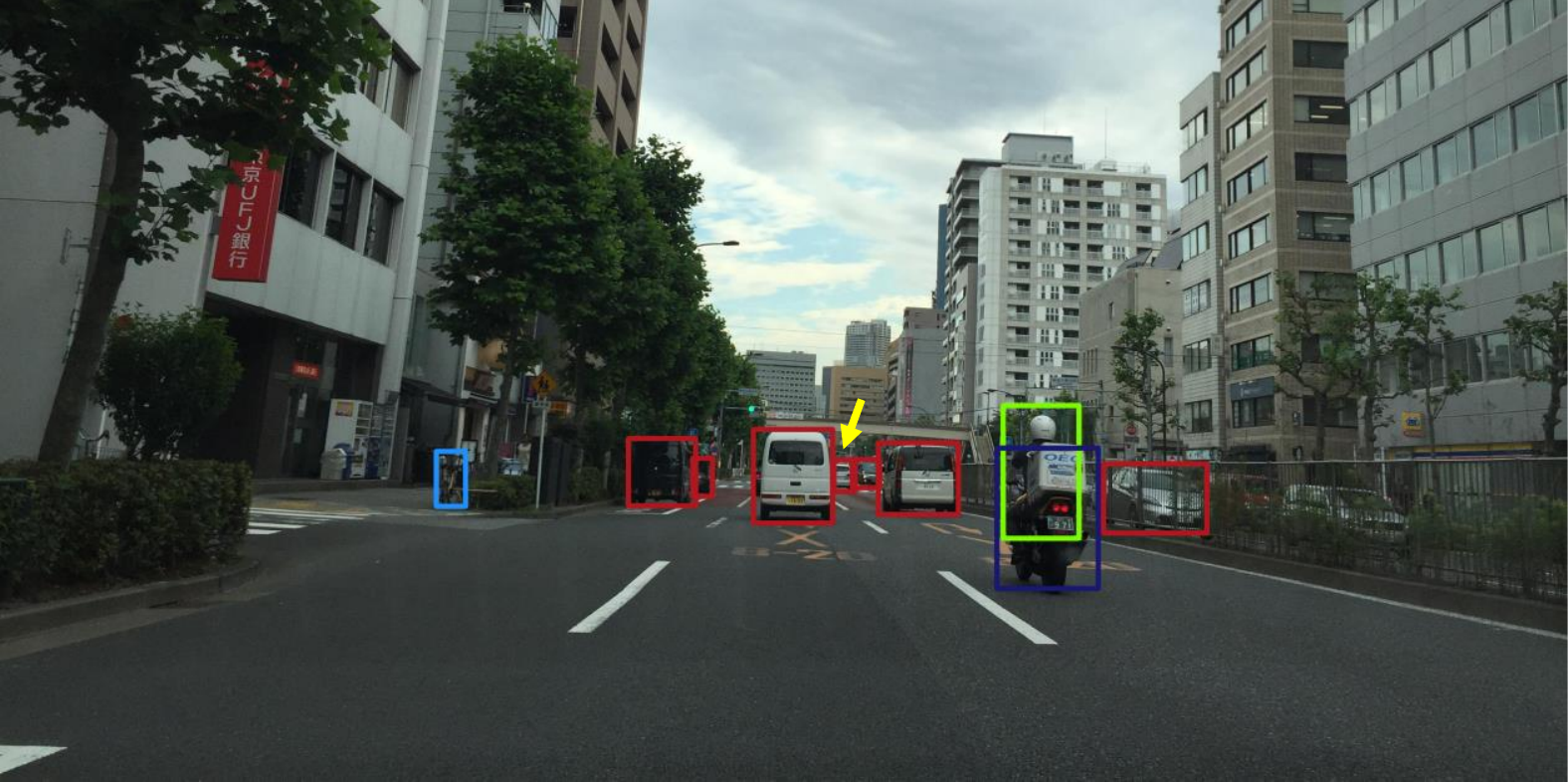}
\end{minipage}
\begin{minipage}[h]{0.245\linewidth}
\centering\includegraphics[width=.99\linewidth]{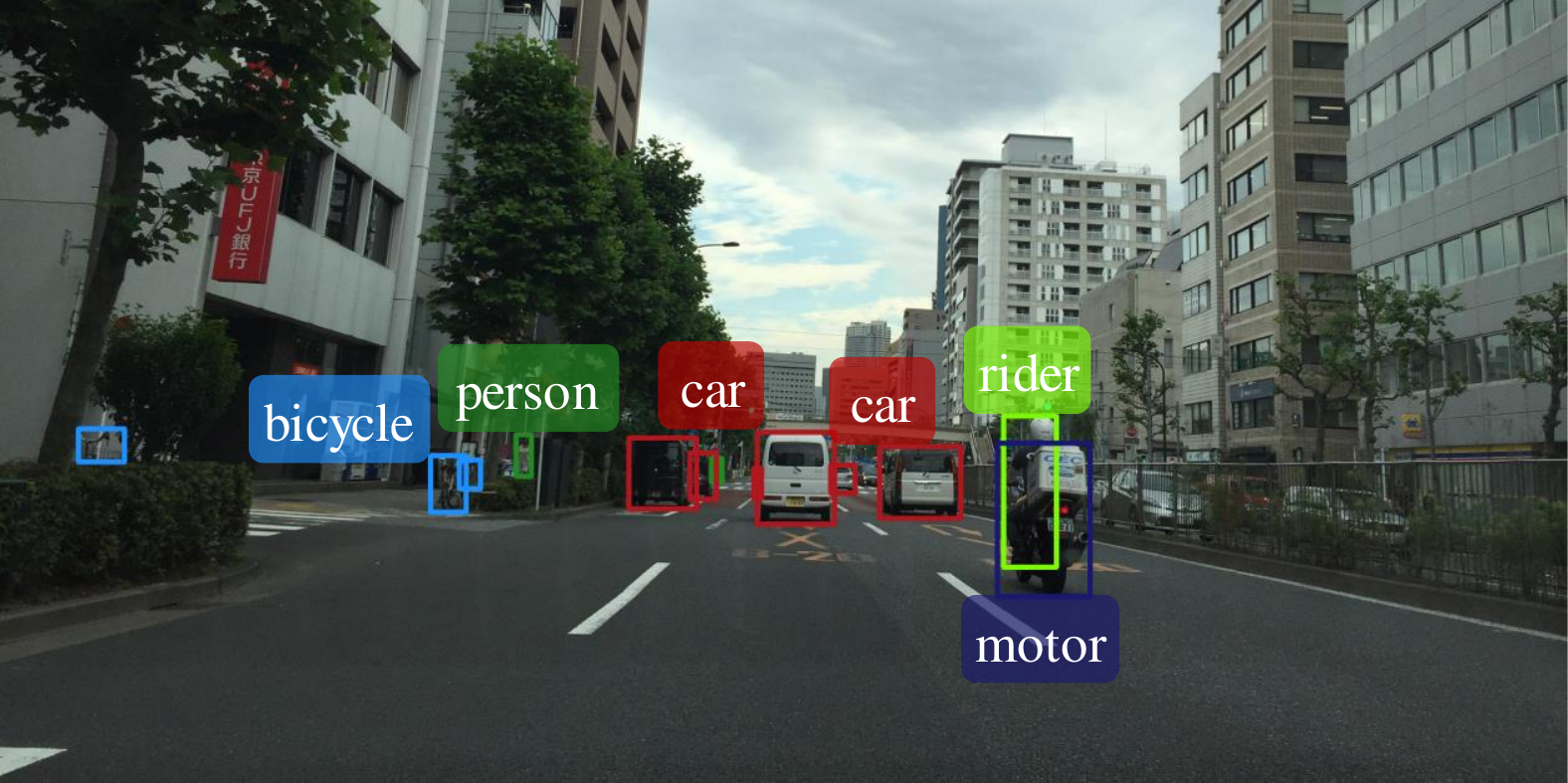}
\end{minipage}
\vspace{5pt}
\begin{minipage}[h]{0.99\linewidth}
\centering\footnotesize {(a) Cityscapes $\rightarrow$ Mapillary Vistas} 
\end{minipage}
\vspace{2pt}
\begin{minipage}[h]{0.245\linewidth}
\centering\includegraphics[width=.99\linewidth]{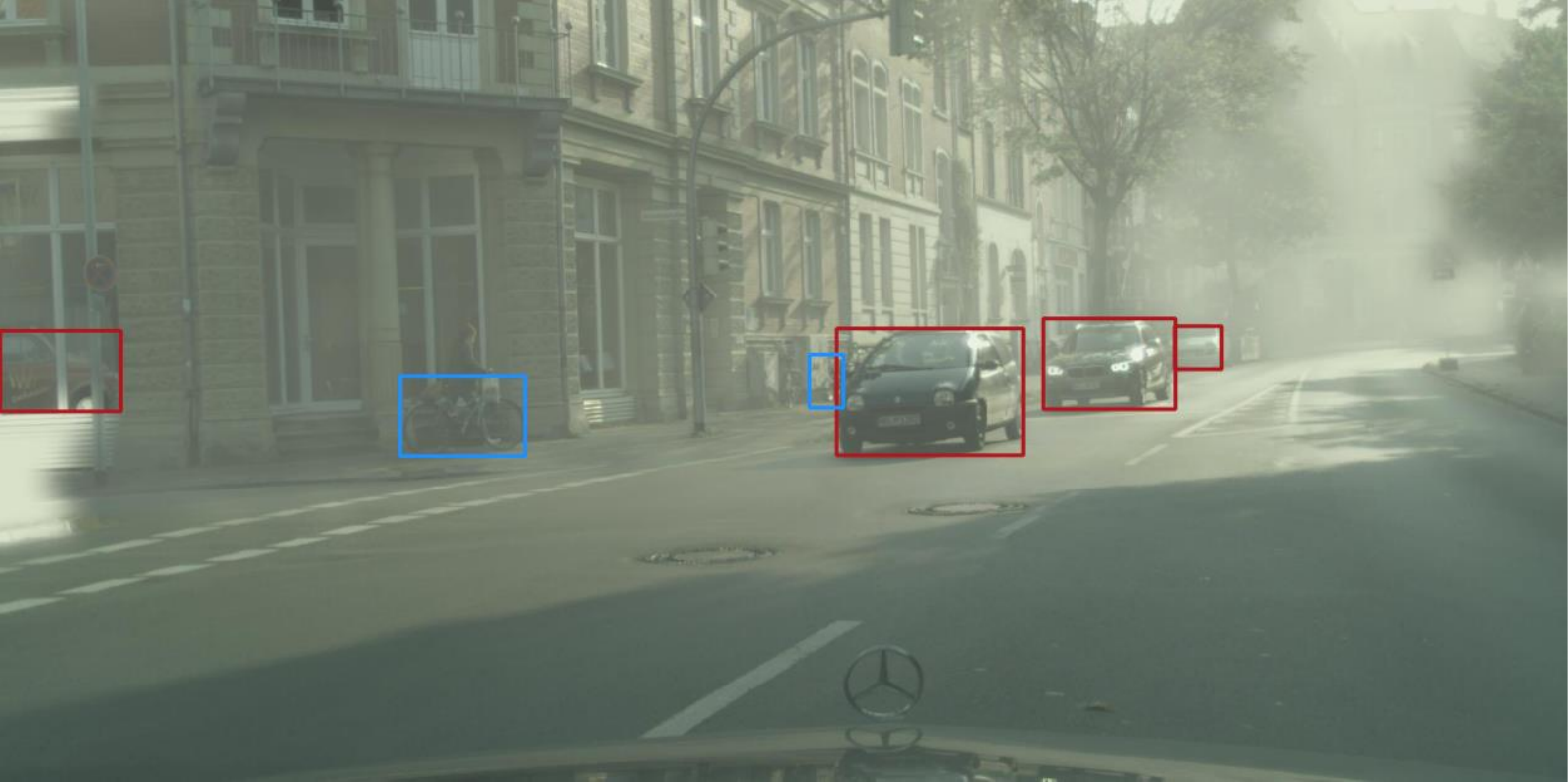}
\end{minipage}
\begin{minipage}[h]{0.245\linewidth}
\centering\includegraphics[width=.99\linewidth]{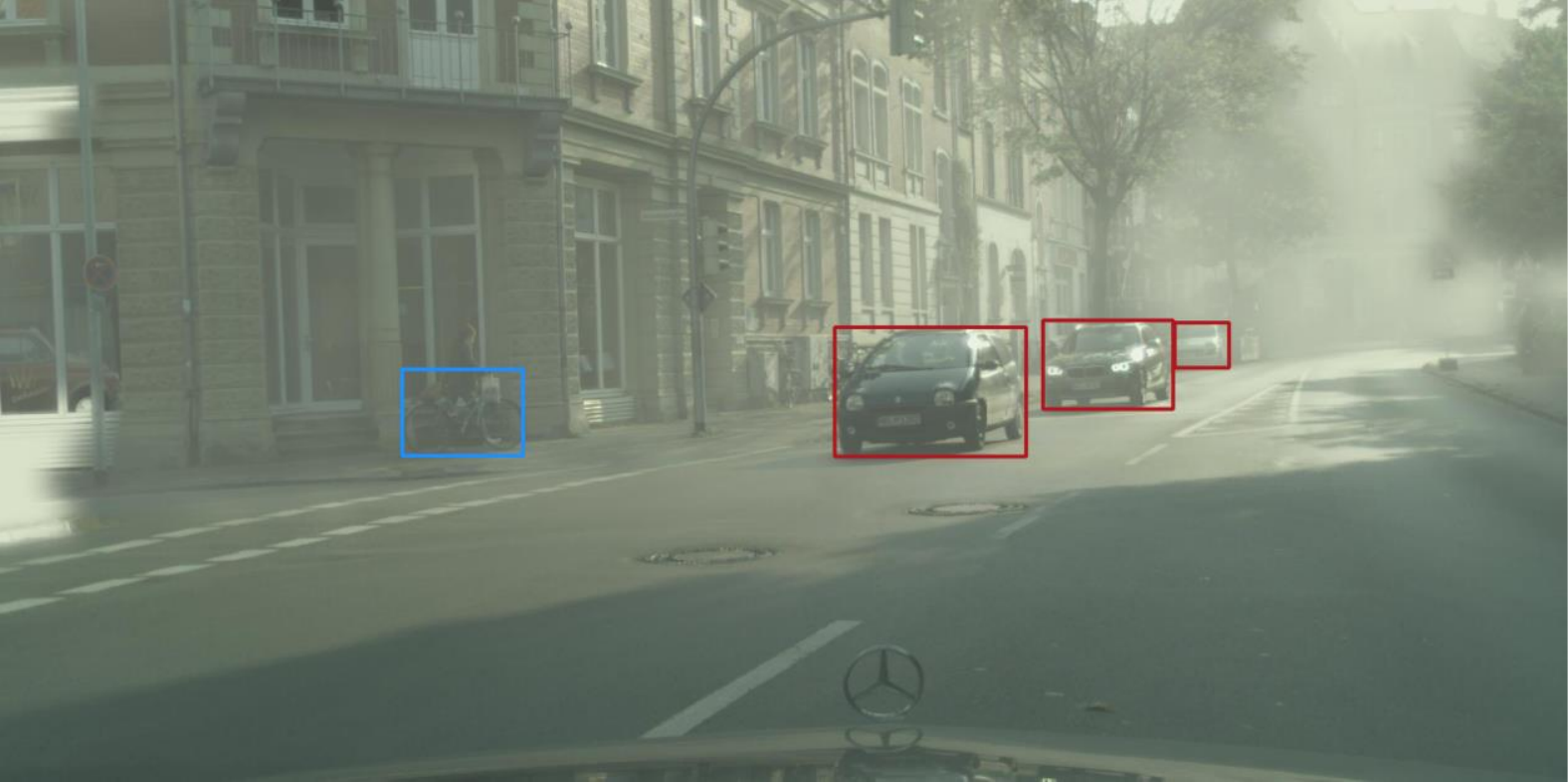}
\end{minipage}
\begin{minipage}[h]{0.245\linewidth}
\centering\includegraphics[width=.99\linewidth]{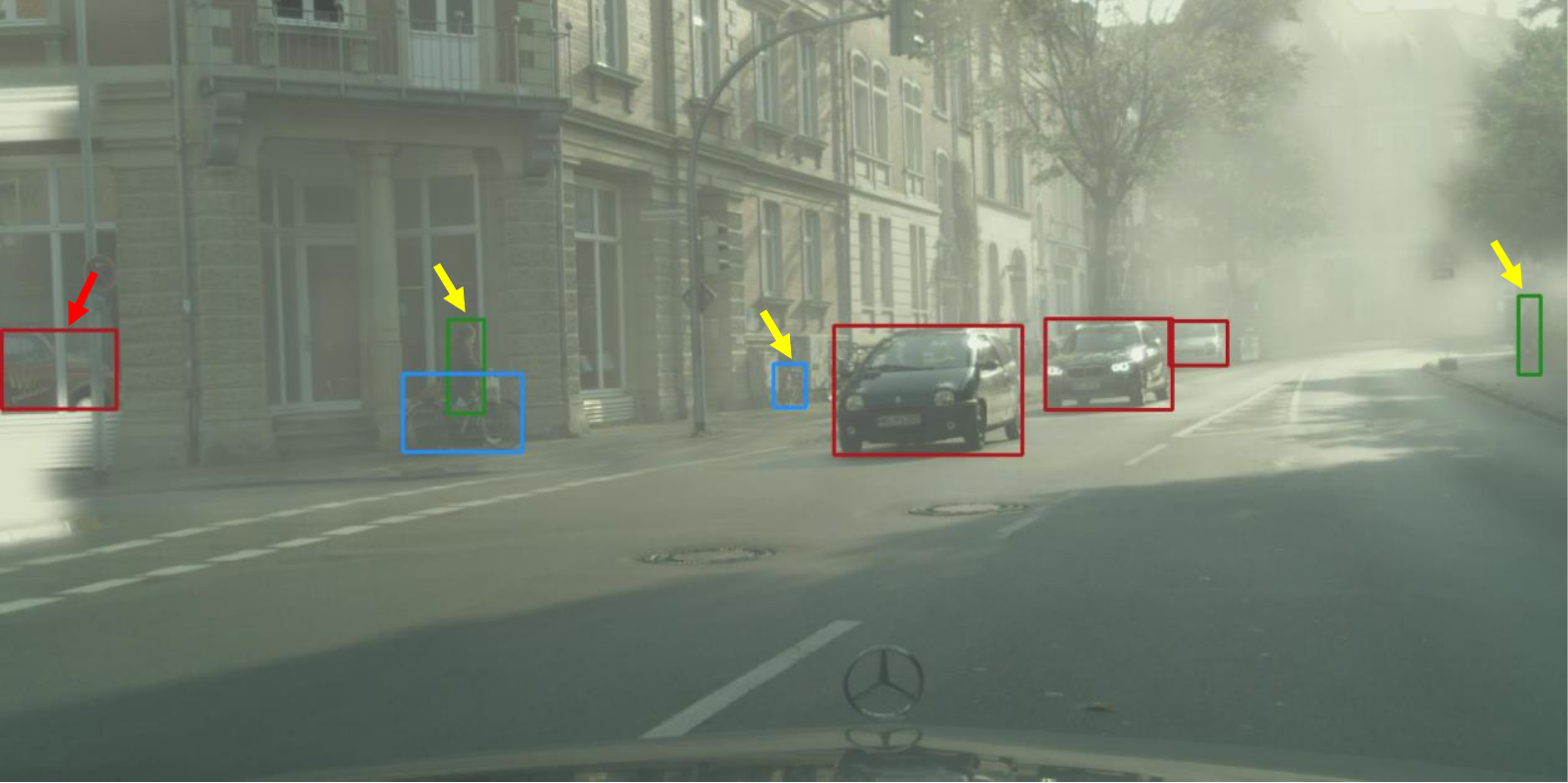}
\end{minipage}
\begin{minipage}[h]{0.245\linewidth}
\centering\includegraphics[width=.99\linewidth]{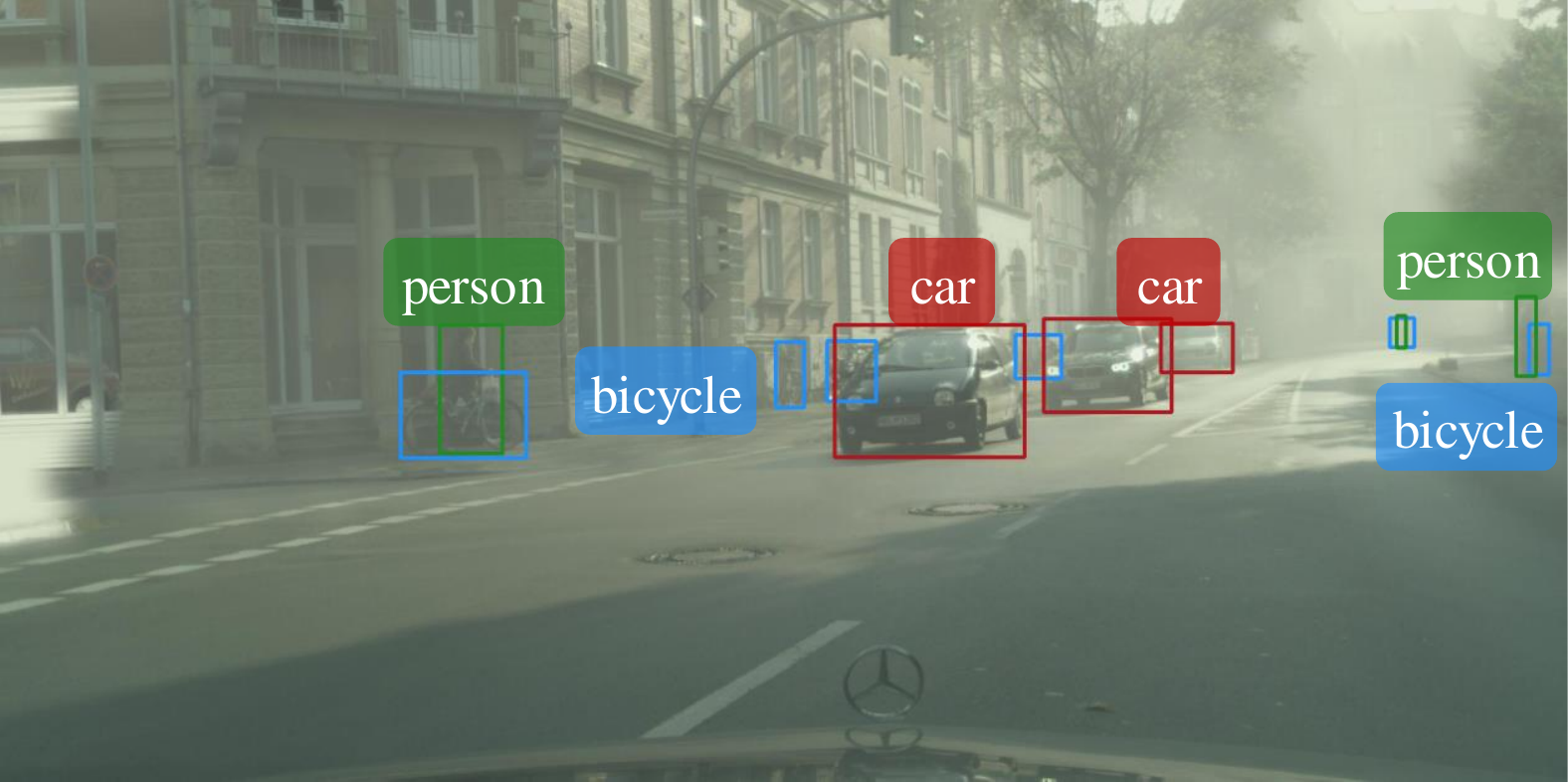}
\end{minipage}
\vspace{2pt}
\begin{minipage}[h]{0.245\linewidth}
\centering\includegraphics[width=.99\linewidth]{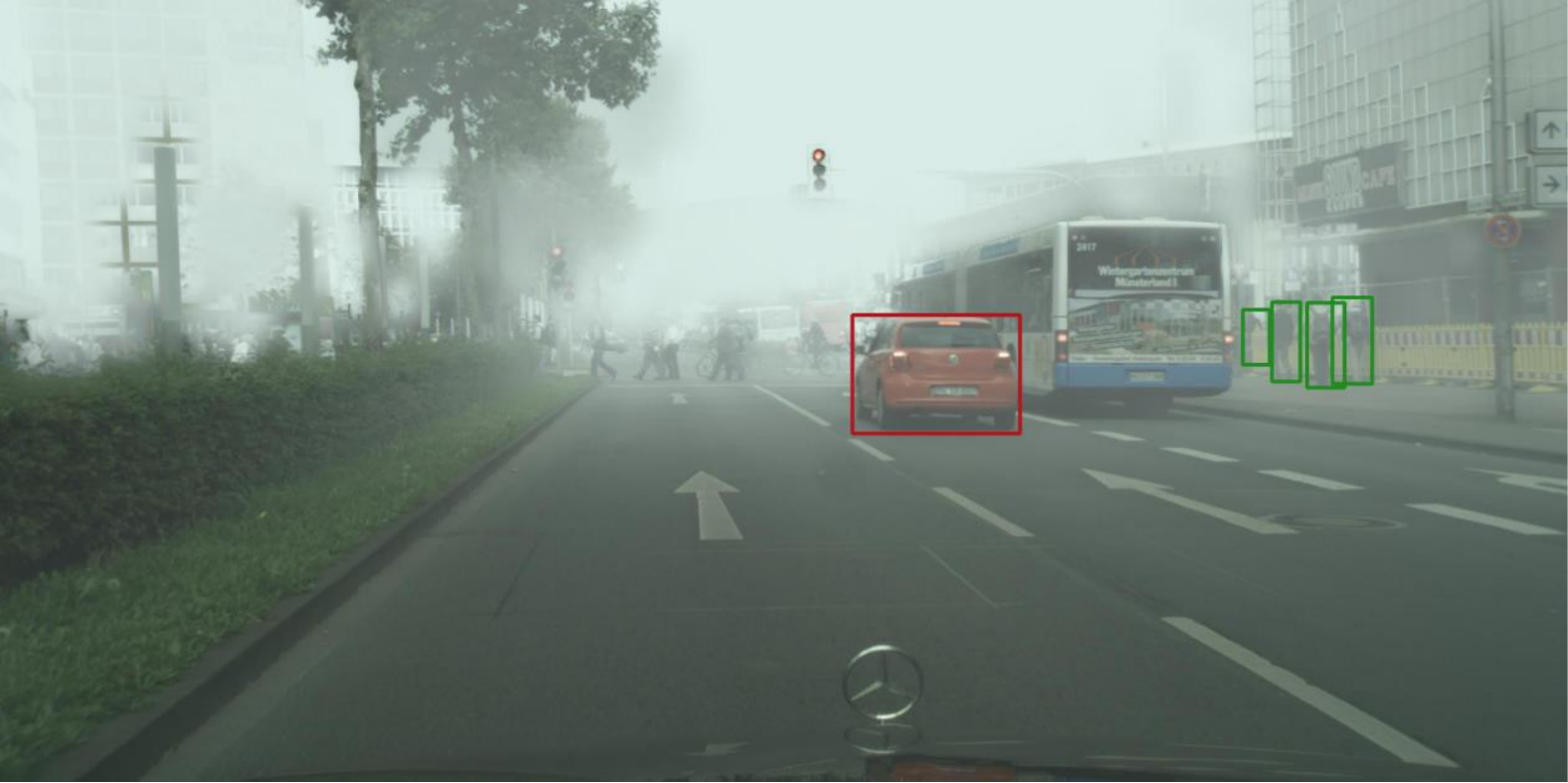}
\end{minipage}
\begin{minipage}[h]{0.245\linewidth}
\centering\includegraphics[width=.99\linewidth]{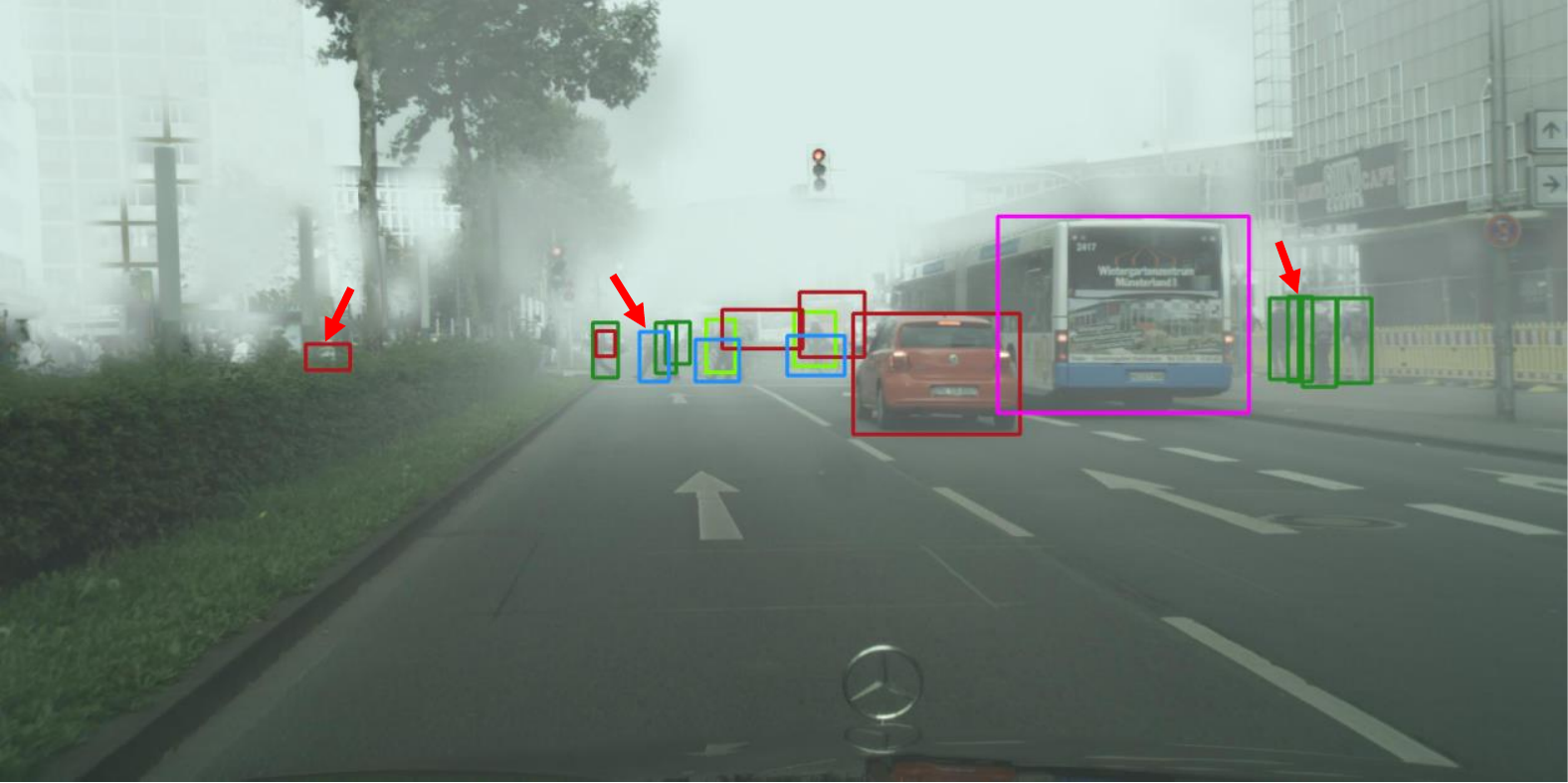}
\end{minipage}
\begin{minipage}[h]{0.245\linewidth}
\centering\includegraphics[width=.99\linewidth]{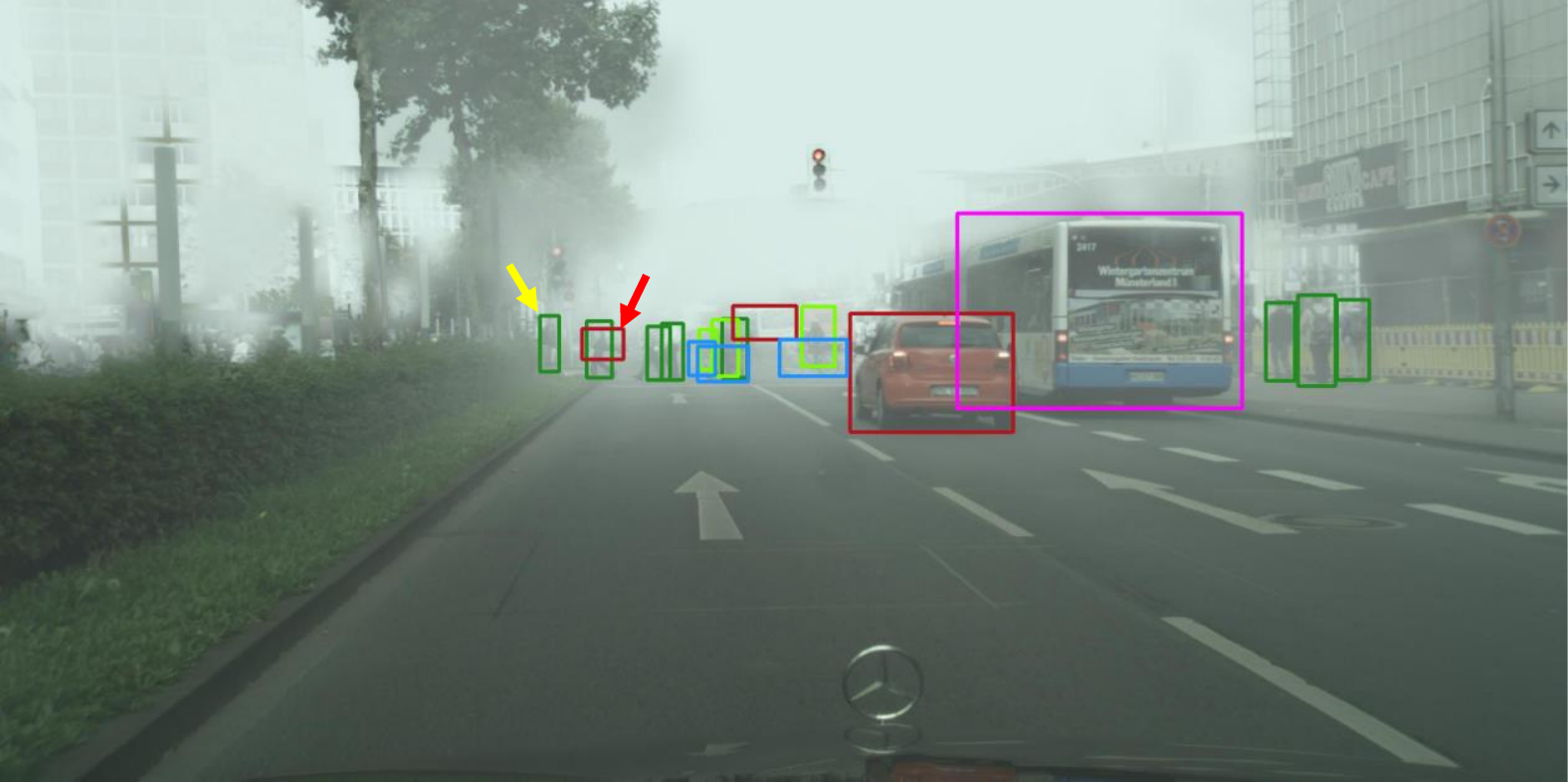}
\end{minipage}
\begin{minipage}[h]{0.245\linewidth}
\centering\includegraphics[width=.99\linewidth]{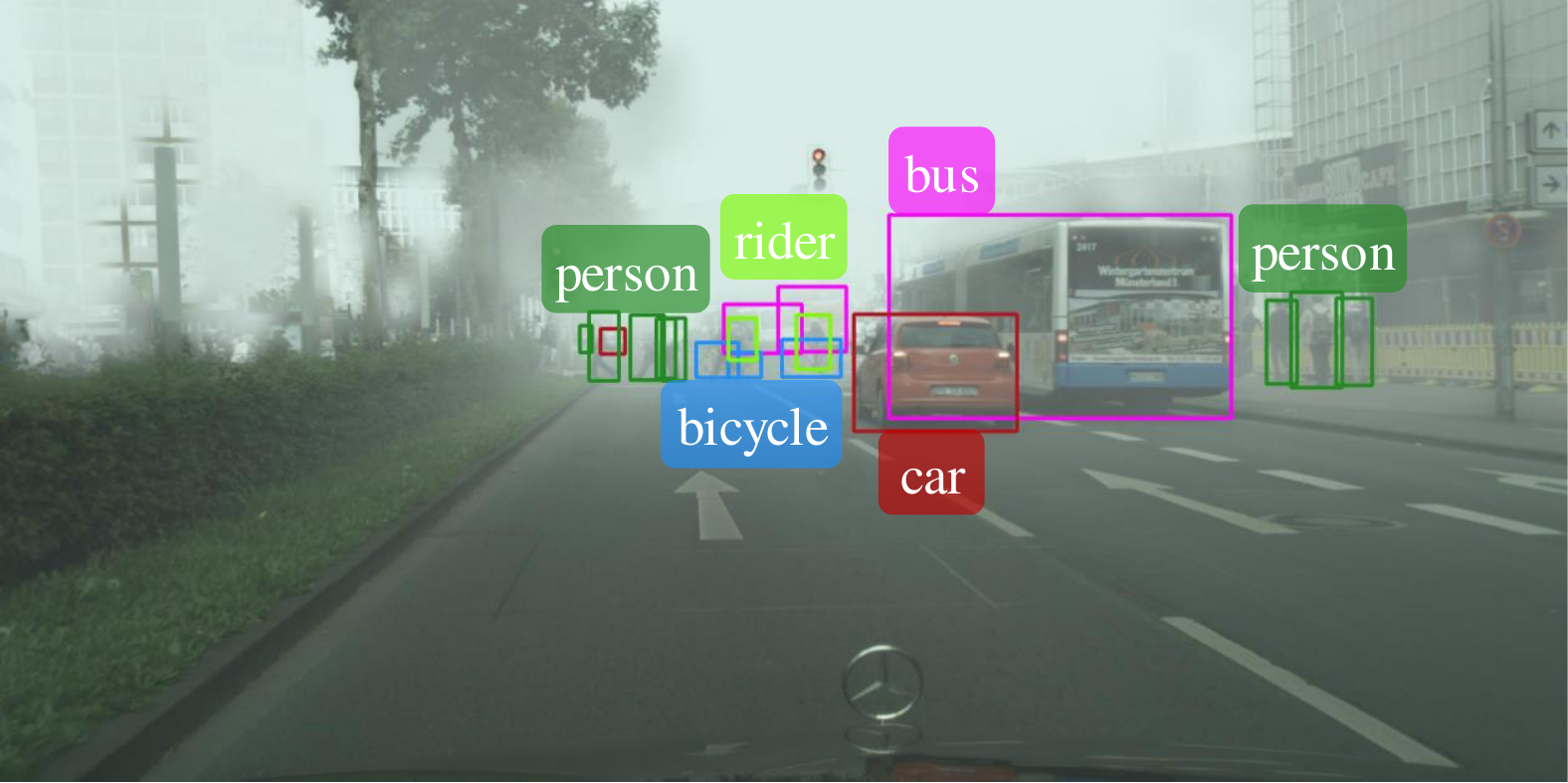}
\end{minipage}
\vspace{5pt}
\begin{minipage}[h]{0.99\linewidth}
\centering\footnotesize {(b) Cityscapes $\rightarrow$ Foggy Cityscapes} 
\end{minipage}
\vspace{2pt}
\begin{minipage}[h]{0.245\linewidth}
\centering\includegraphics[width=.99\linewidth]{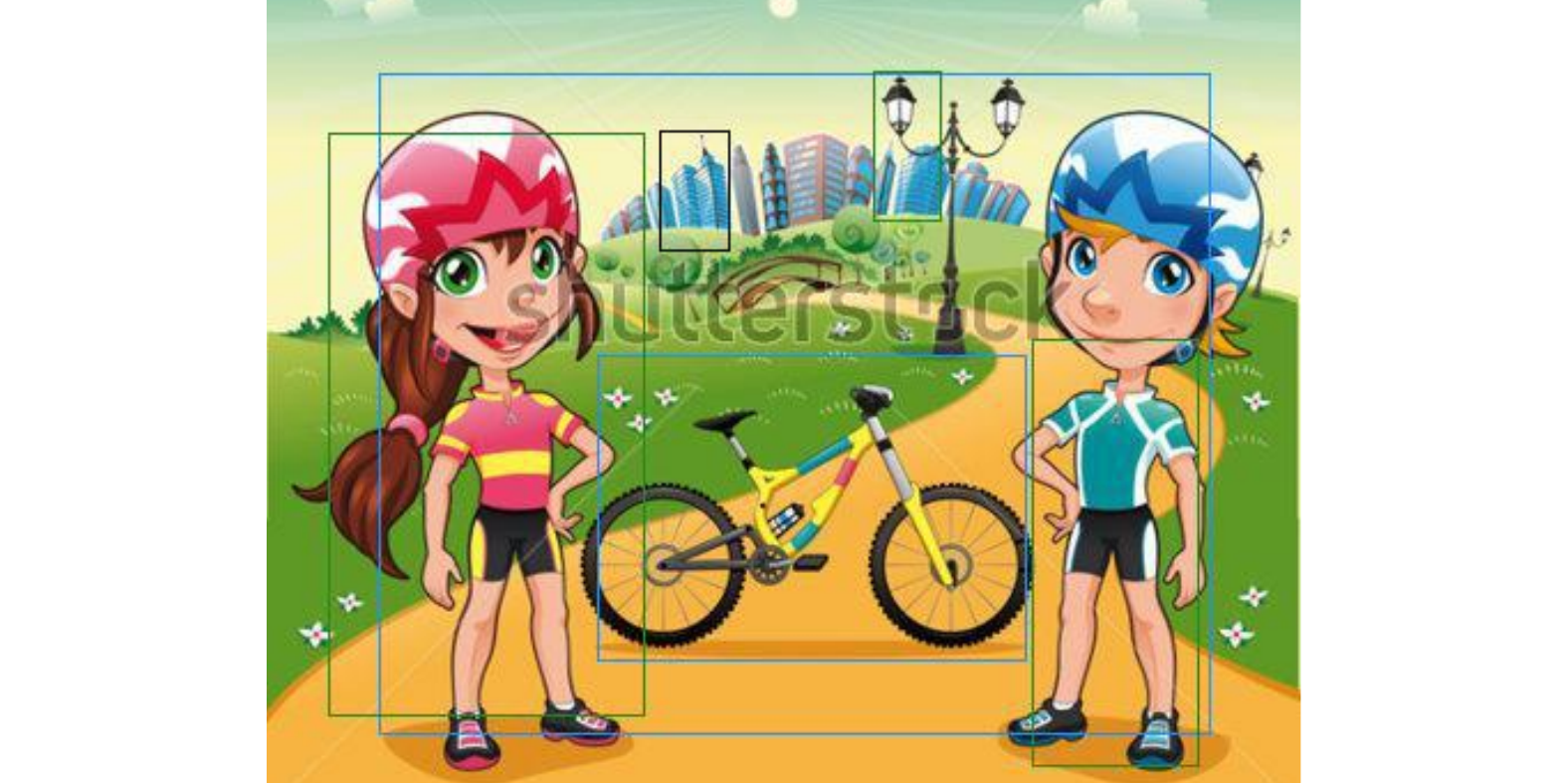}
\end{minipage}
\begin{minipage}[h]{0.245\linewidth}
\centering\includegraphics[width=.99\linewidth]{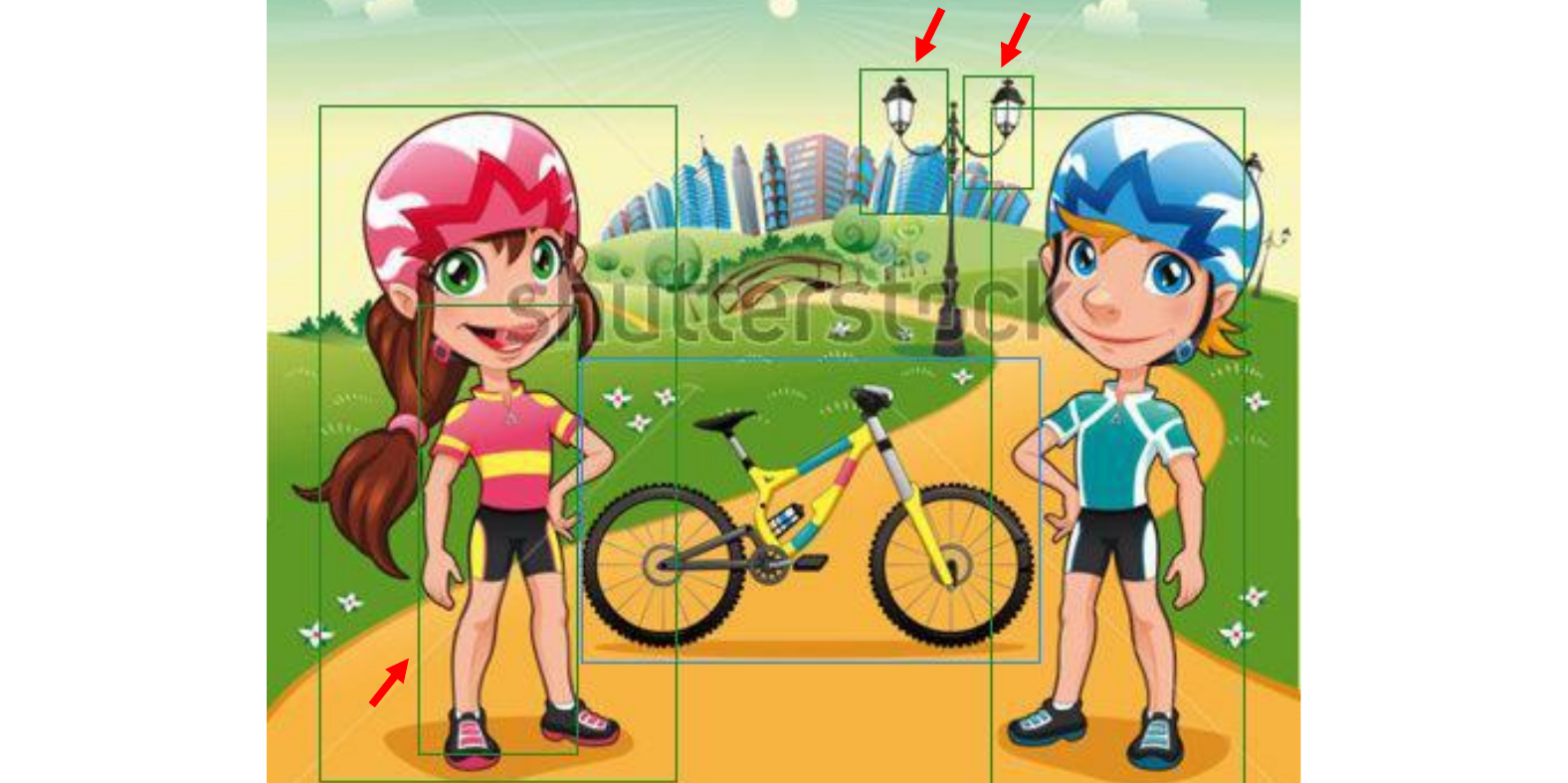}
\end{minipage}
\begin{minipage}[h]{0.245\linewidth}
\centering\includegraphics[width=.99\linewidth]{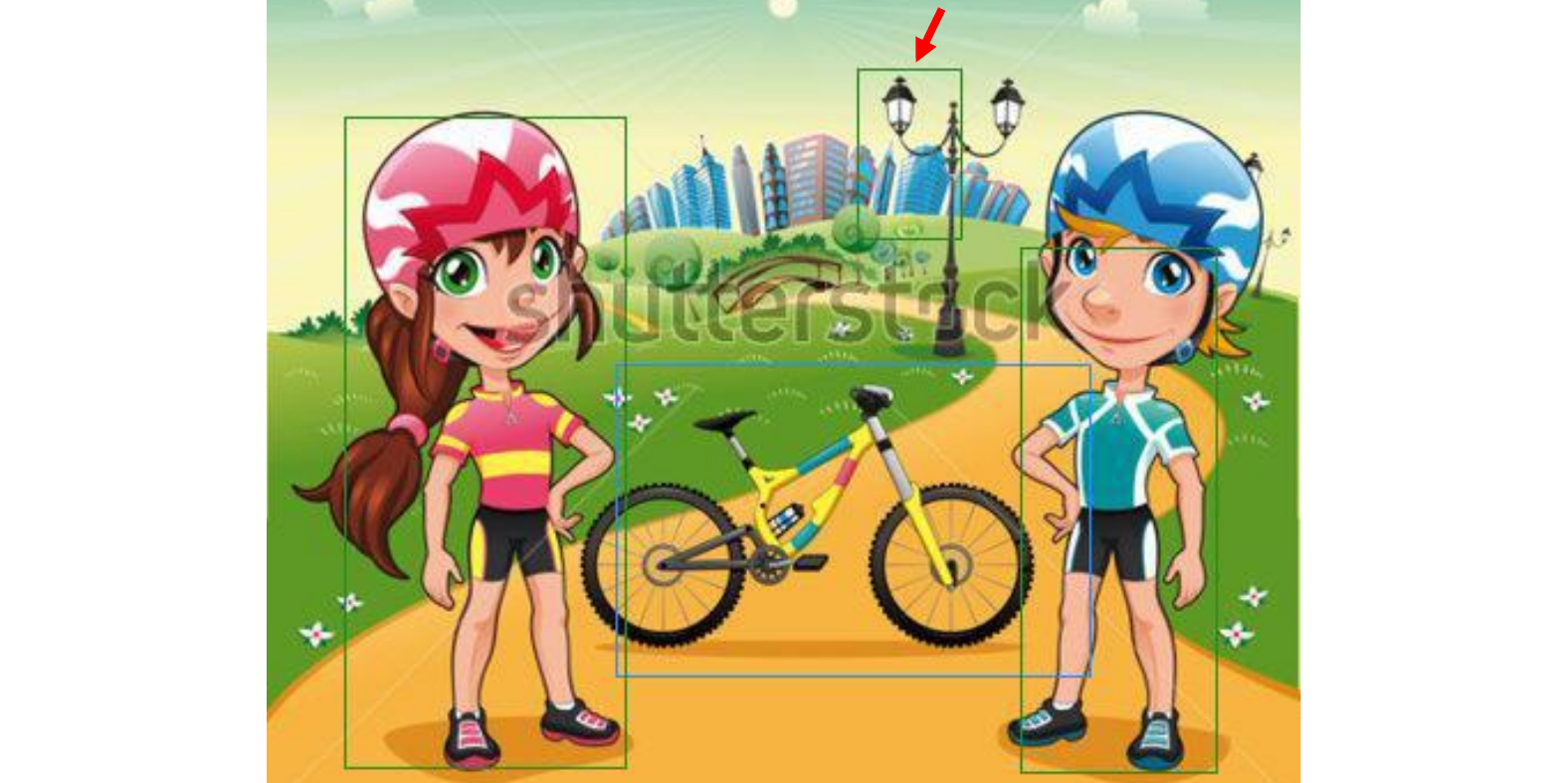}
\end{minipage}
\begin{minipage}[h]{0.245\linewidth}
\centering\includegraphics[width=.99\linewidth]{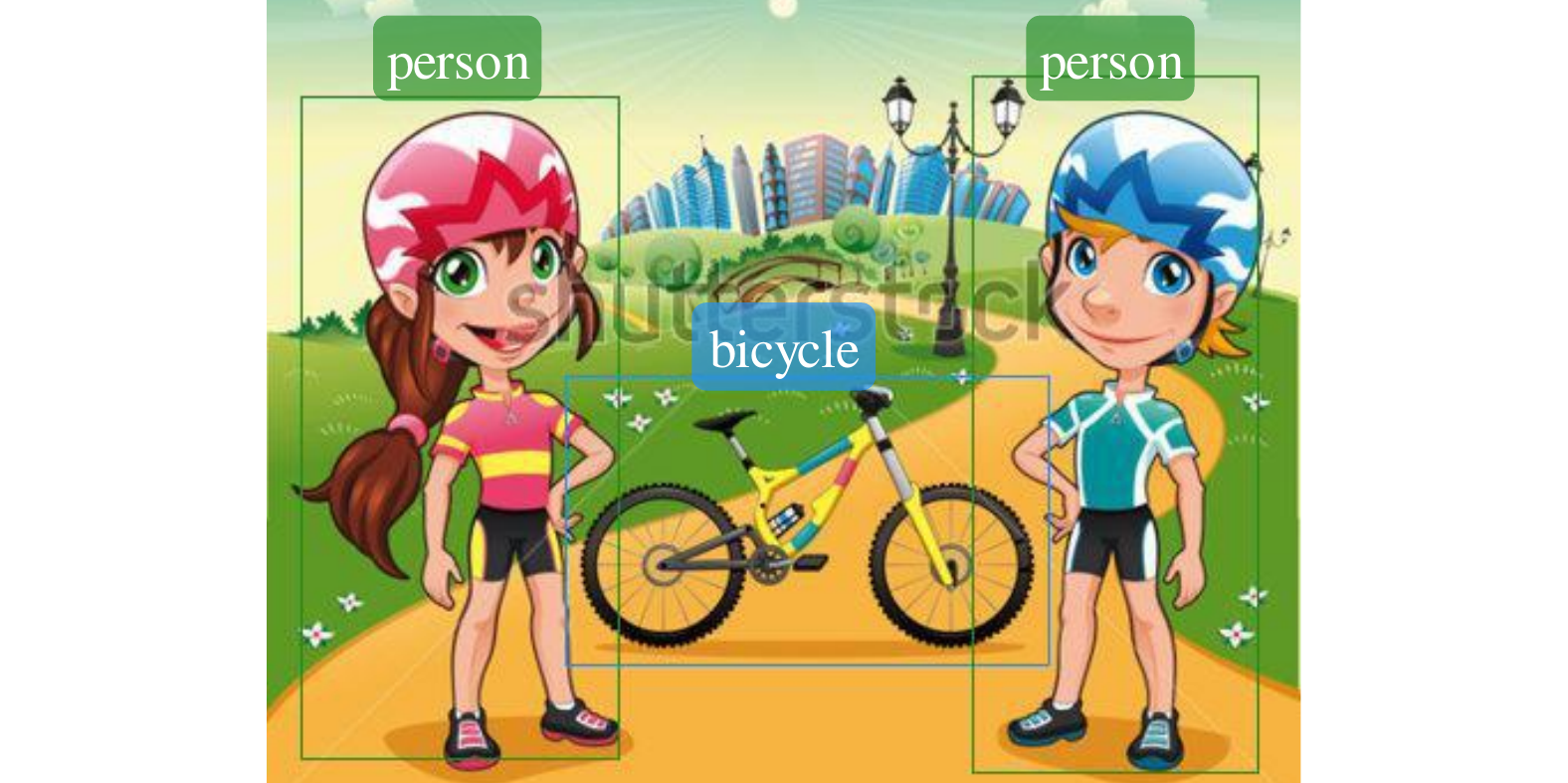}
\end{minipage}
\vspace{2pt}
\begin{minipage}[h]{0.245\linewidth}
\centering\includegraphics[width=.99\linewidth]{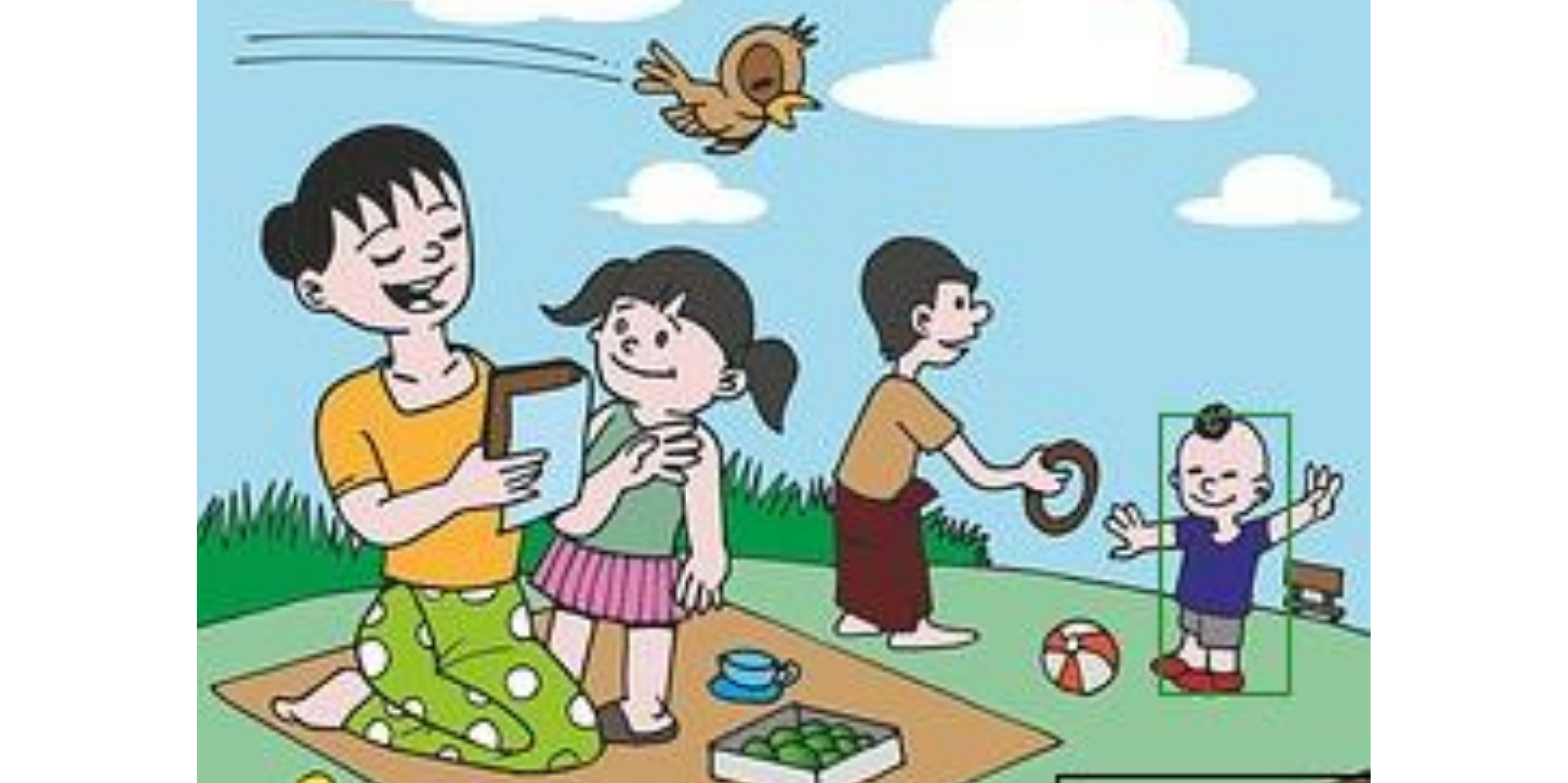}
\end{minipage}
\begin{minipage}[h]{0.245\linewidth}
\centering\includegraphics[width=.99\linewidth]{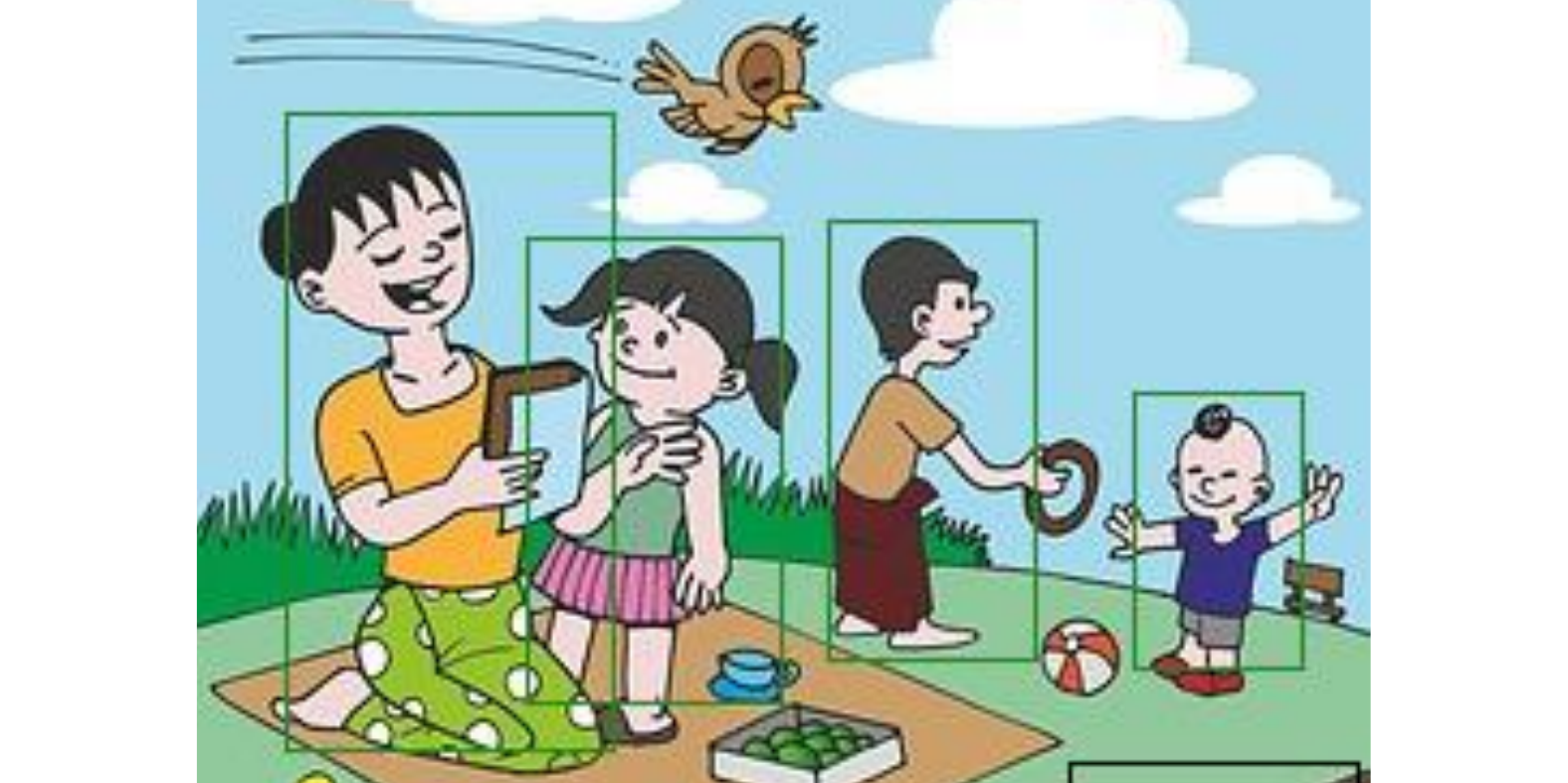}
\end{minipage}
\begin{minipage}[h]{0.245\linewidth}
\centering\includegraphics[width=.99\linewidth]{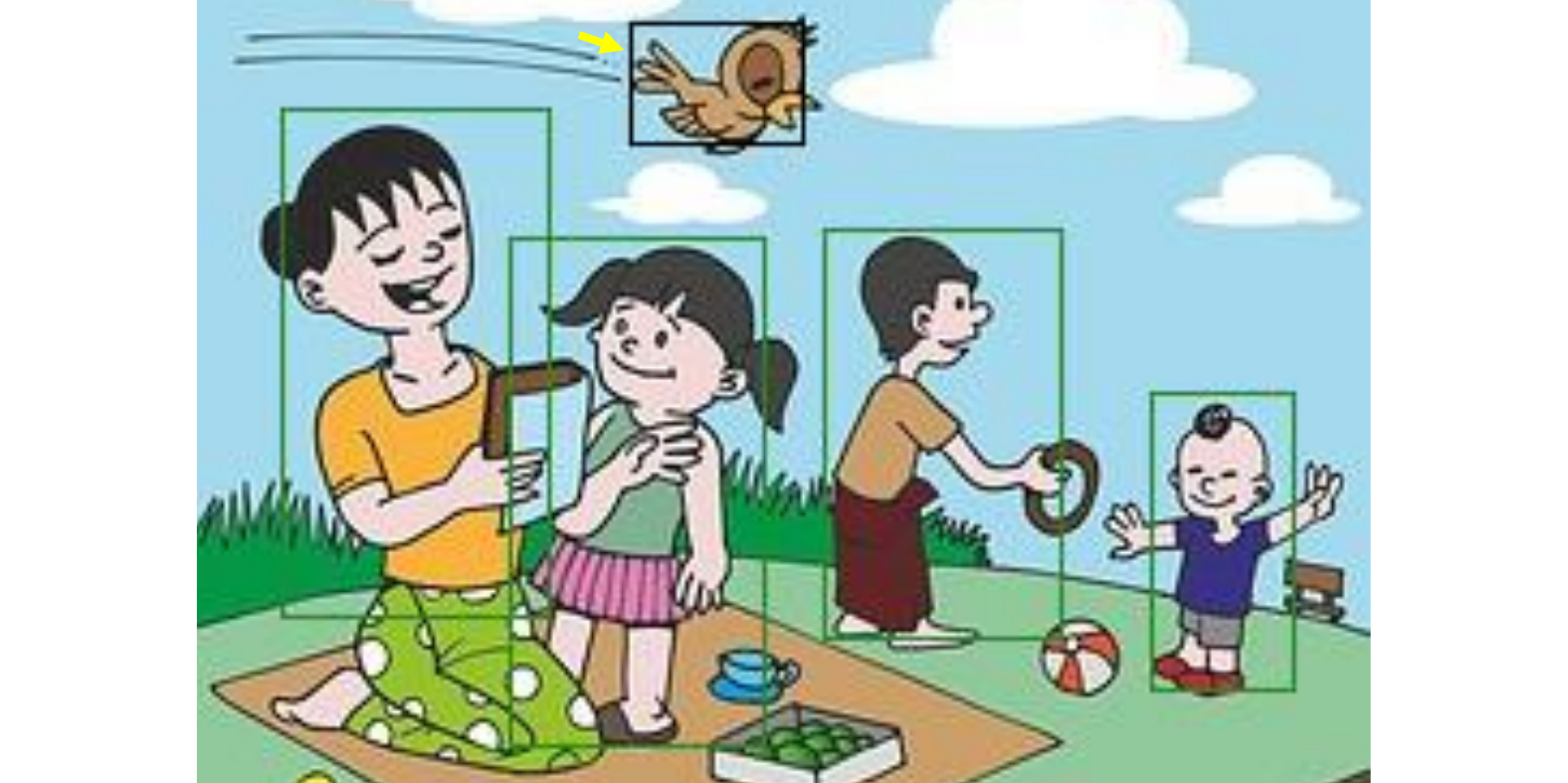}
\end{minipage}
\begin{minipage}[h]{0.245\linewidth}
\centering\includegraphics[width=.99\linewidth]{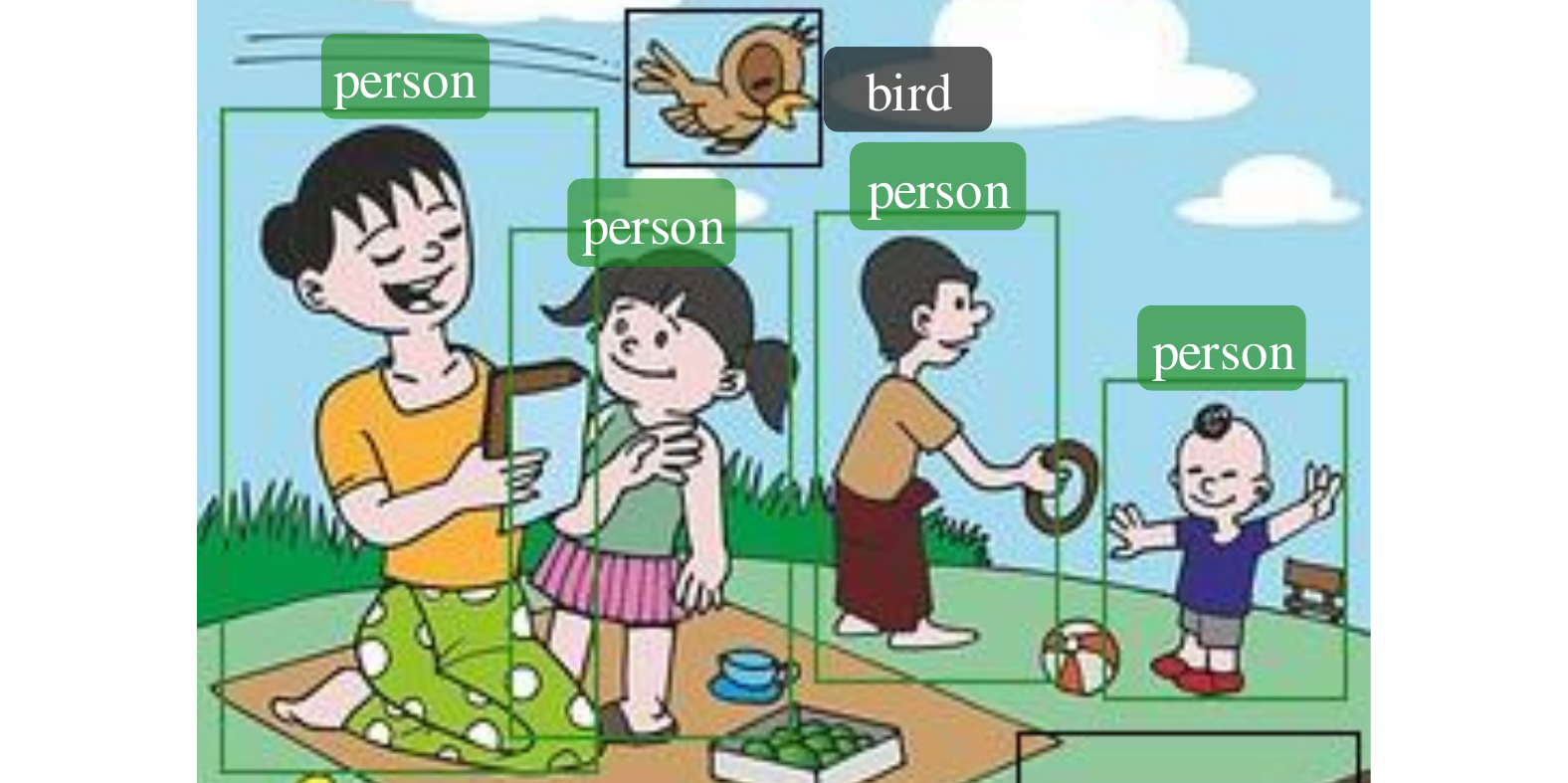}
\end{minipage}
\vspace{5pt}
\begin{minipage}[h]{0.99\linewidth}
\centering\footnotesize {(c) PASCAL VOC $\rightarrow$ Clipart1k} 
\end{minipage}
\vspace{2pt}
\begin{minipage}[h]{0.245\linewidth}
\centering\includegraphics[width=.99\linewidth]{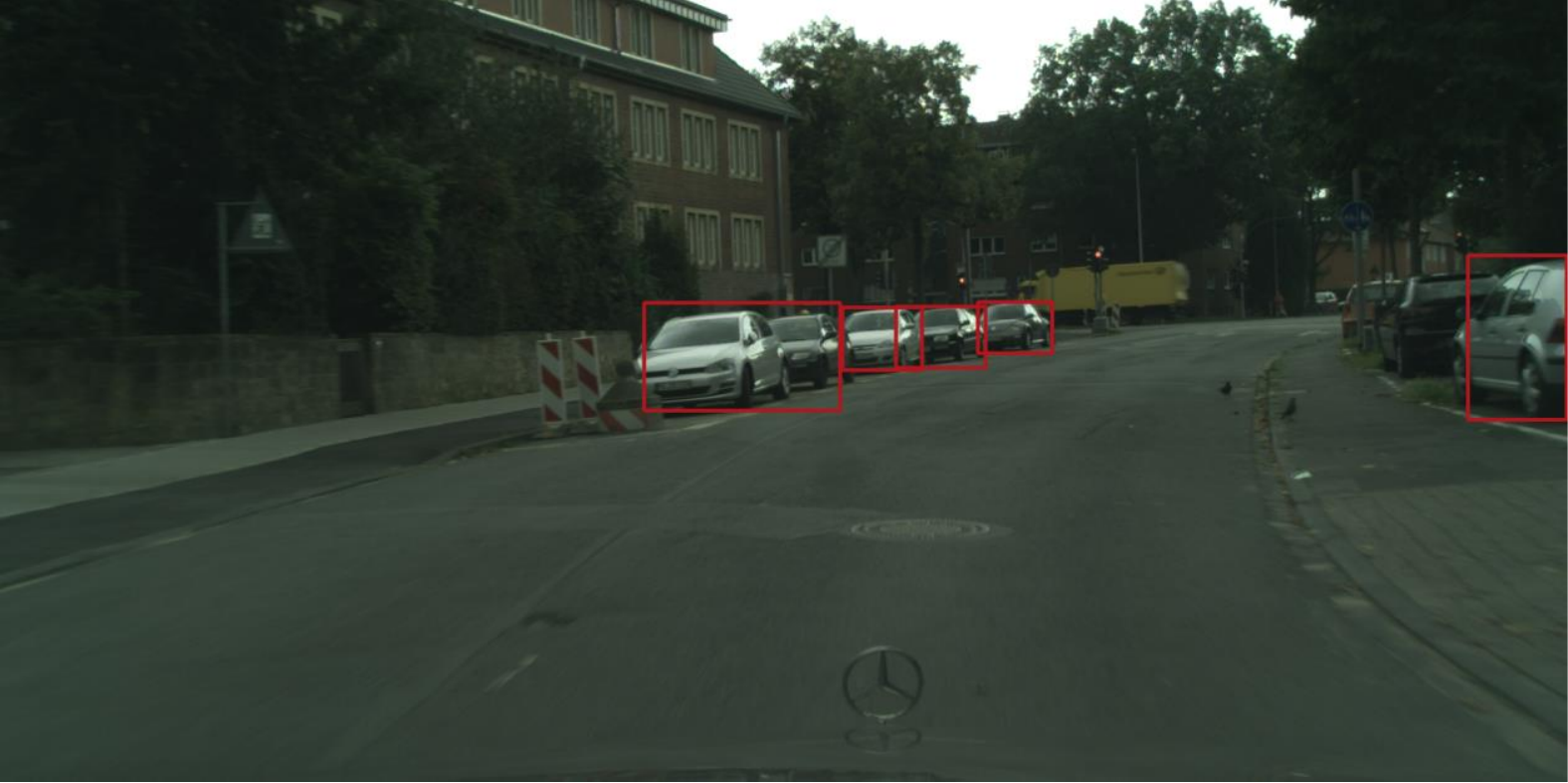}
\end{minipage}
\begin{minipage}[h]{0.245\linewidth}
\centering\includegraphics[width=.99\linewidth]{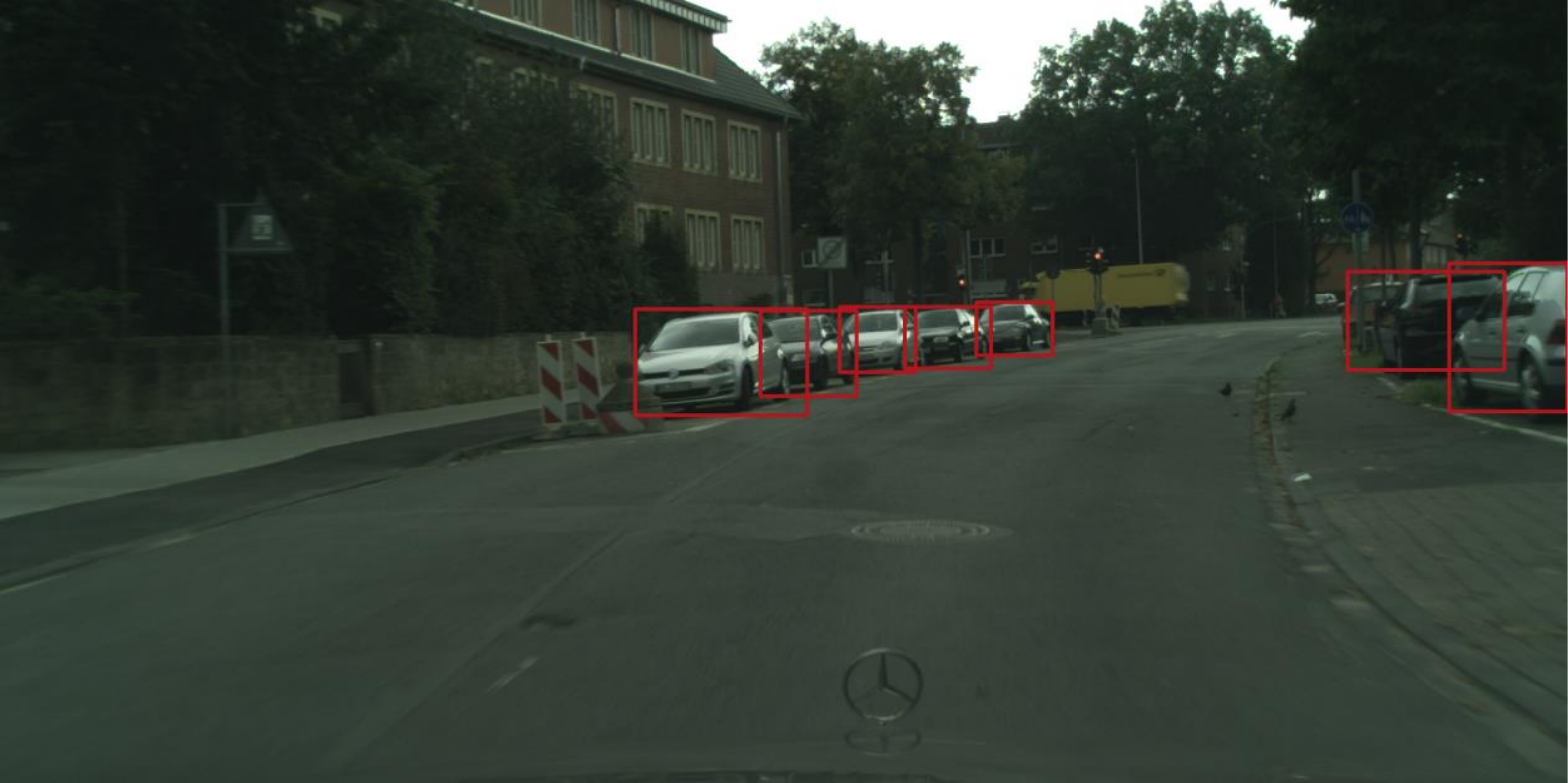}
\end{minipage}
\begin{minipage}[h]{0.245\linewidth}
\centering\includegraphics[width=.99\linewidth]{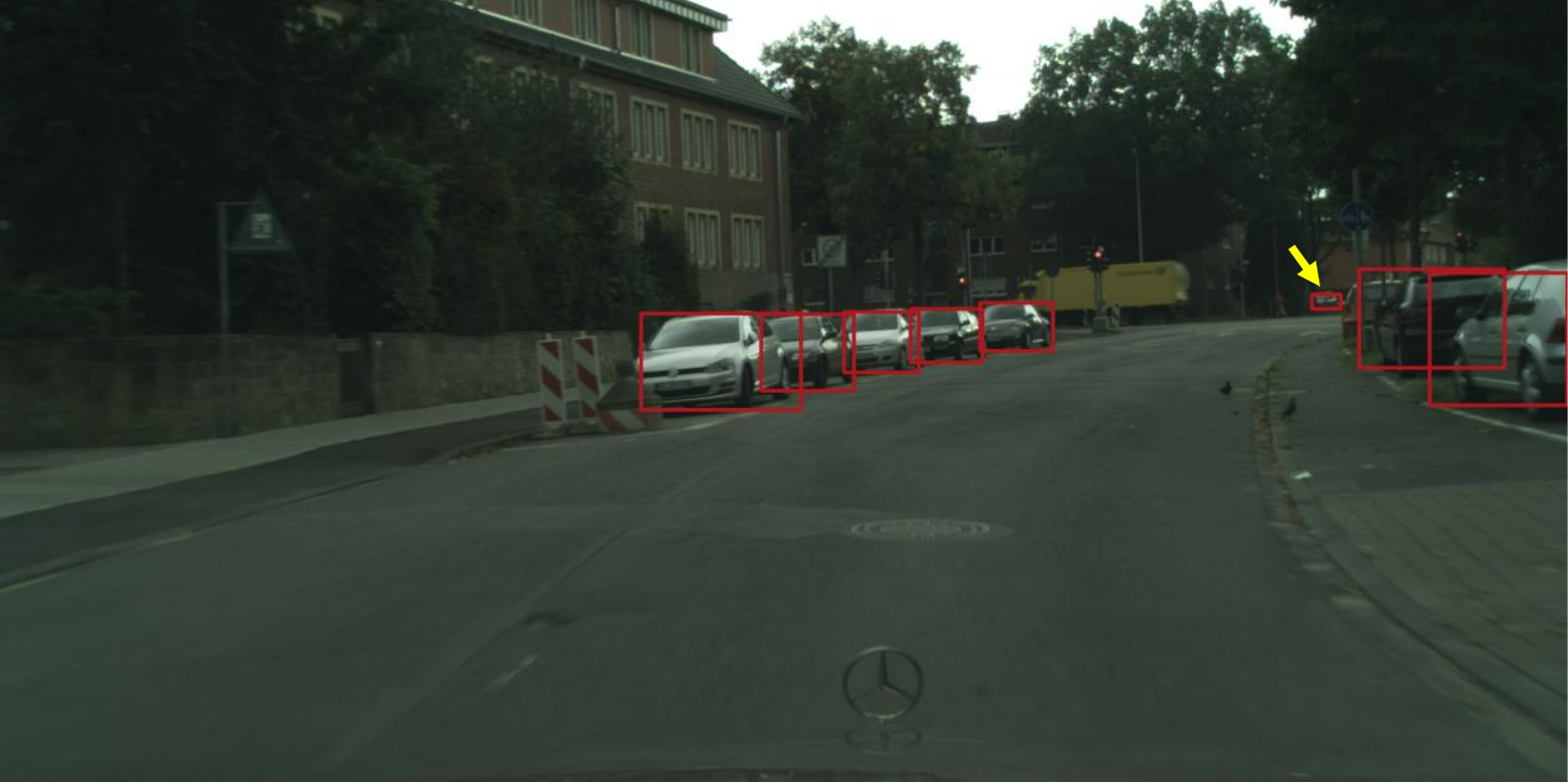}
\end{minipage}
\begin{minipage}[h]{0.245\linewidth}
\centering\includegraphics[width=.99\linewidth]{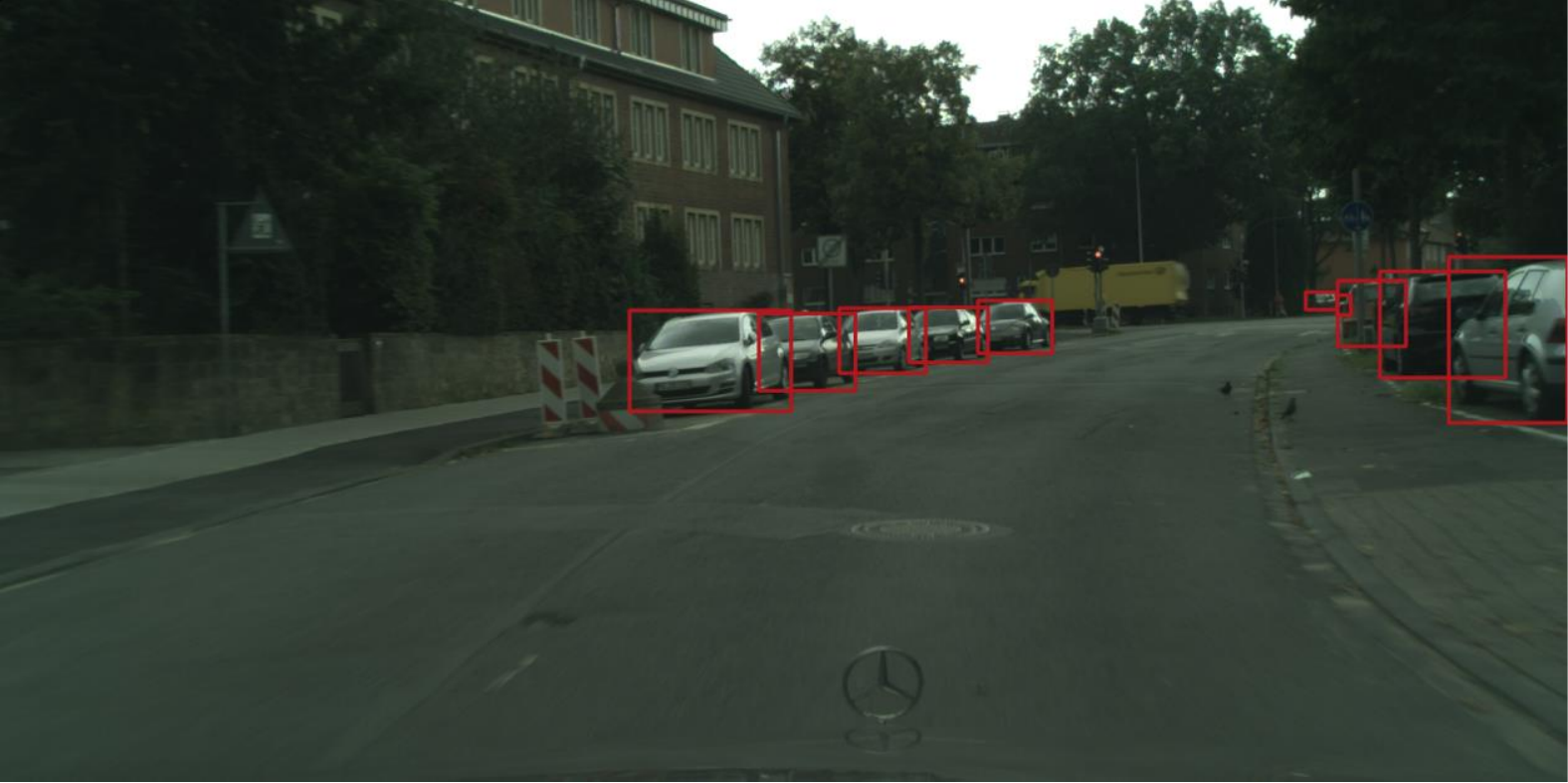}
\end{minipage}
\vspace{2pt}
\begin{minipage}[h]{0.245\linewidth}
\centering\includegraphics[width=.99\linewidth]{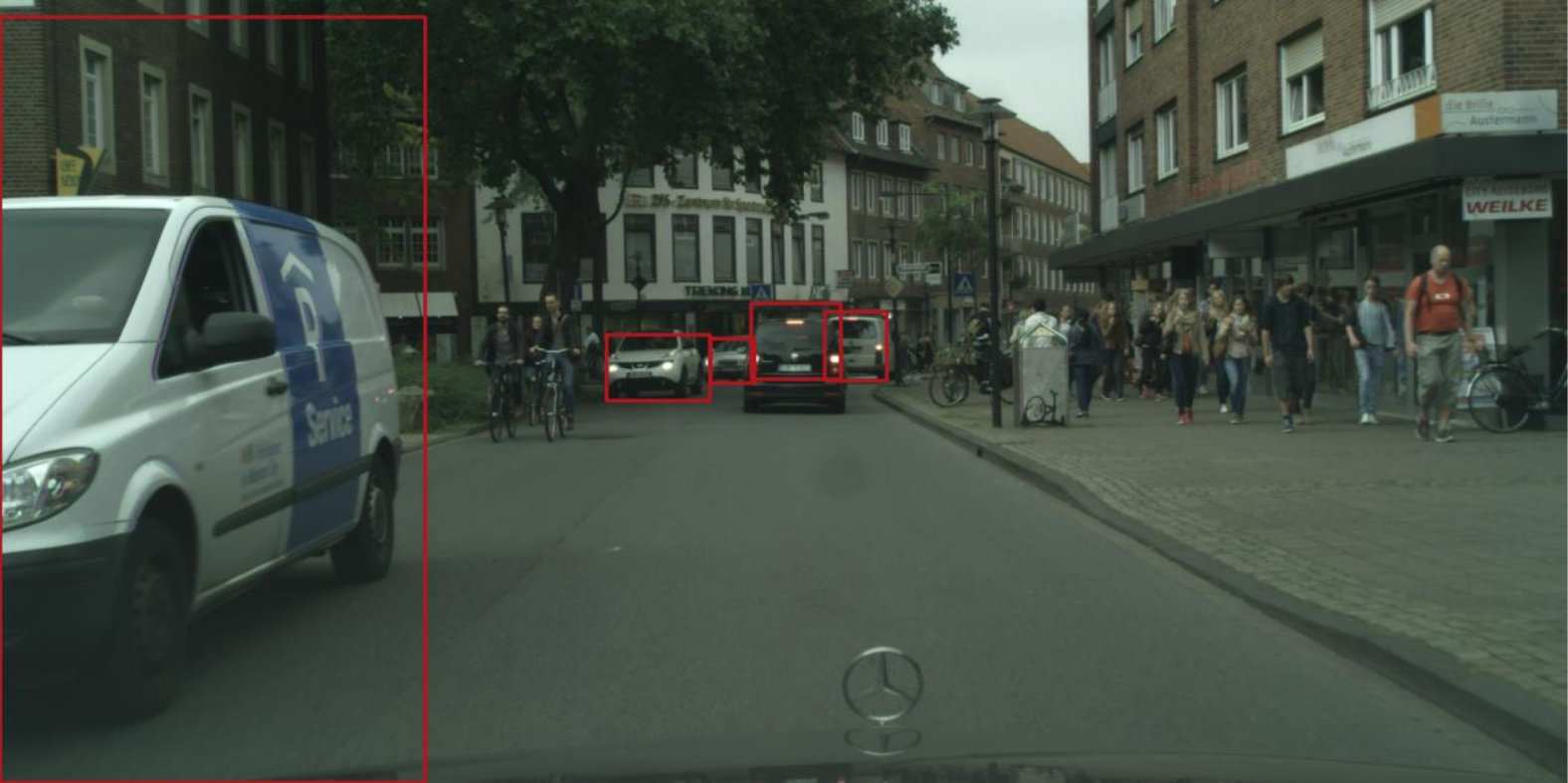}
\end{minipage}
\begin{minipage}[h]{0.245\linewidth}
\centering\includegraphics[width=.99\linewidth]{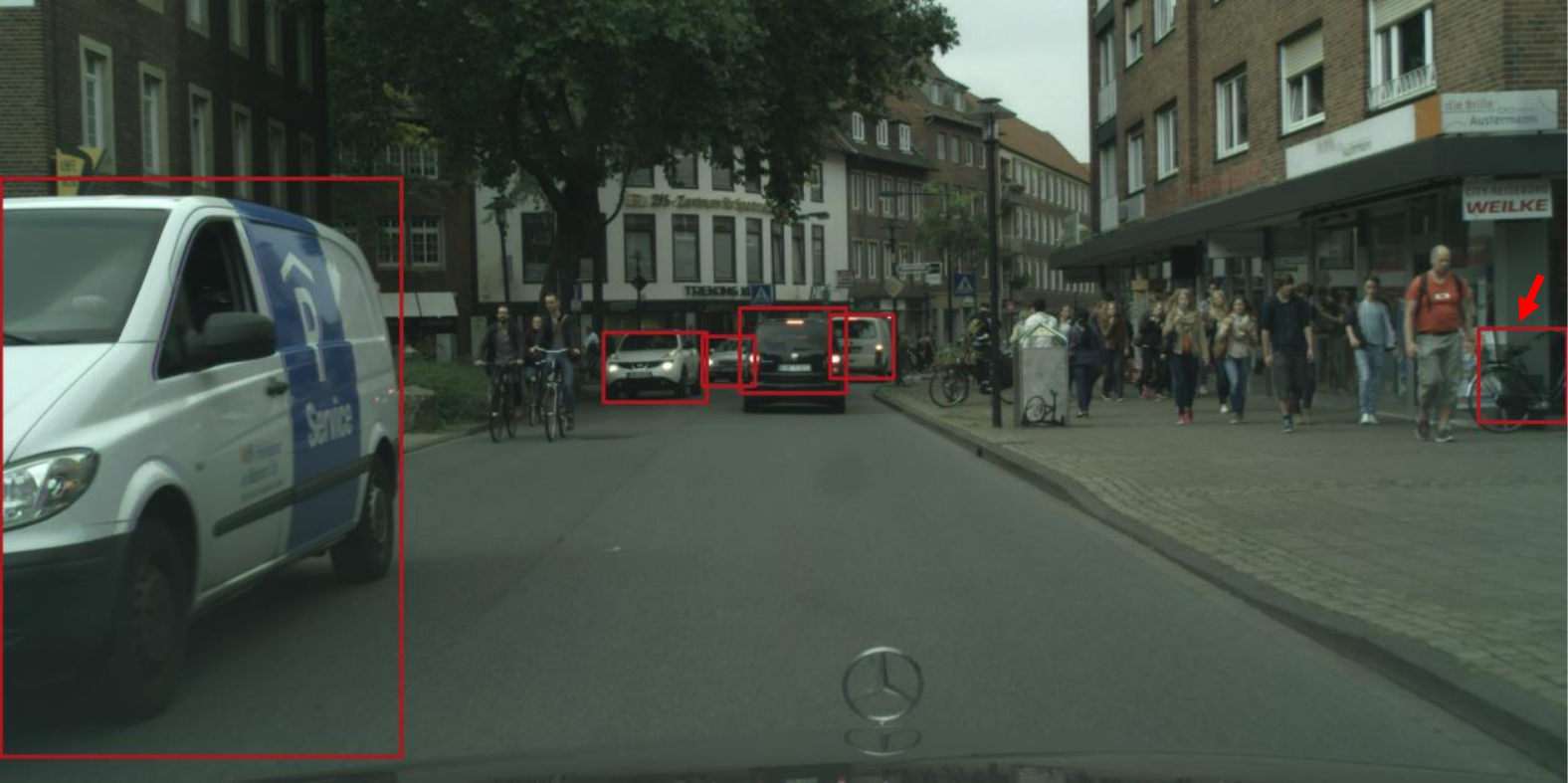}
\end{minipage}
\begin{minipage}[h]{0.245\linewidth}
\centering\includegraphics[width=.99\linewidth]{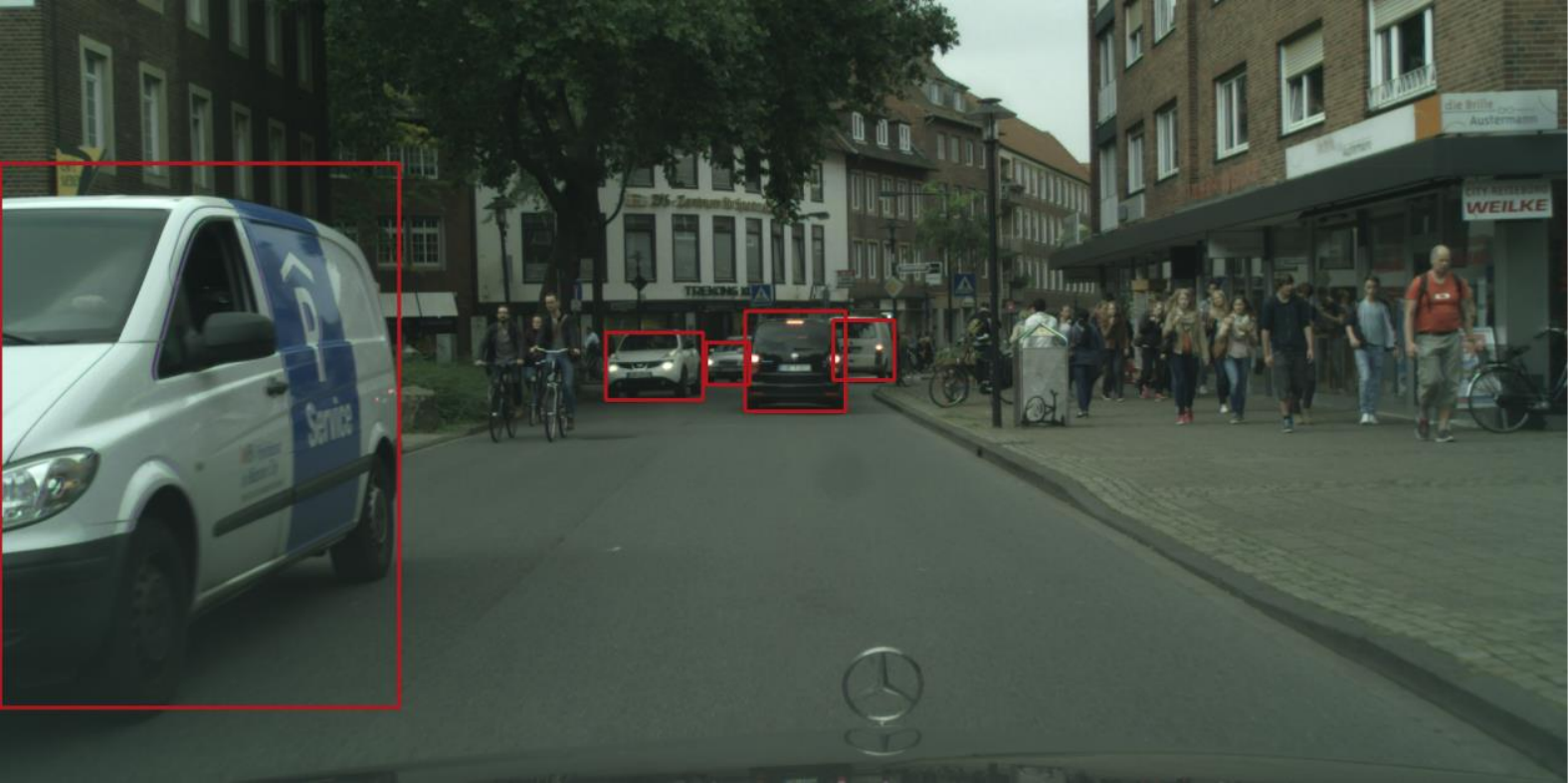}
\end{minipage}
\begin{minipage}[h]{0.245\linewidth}
\centering\includegraphics[width=.99\linewidth]{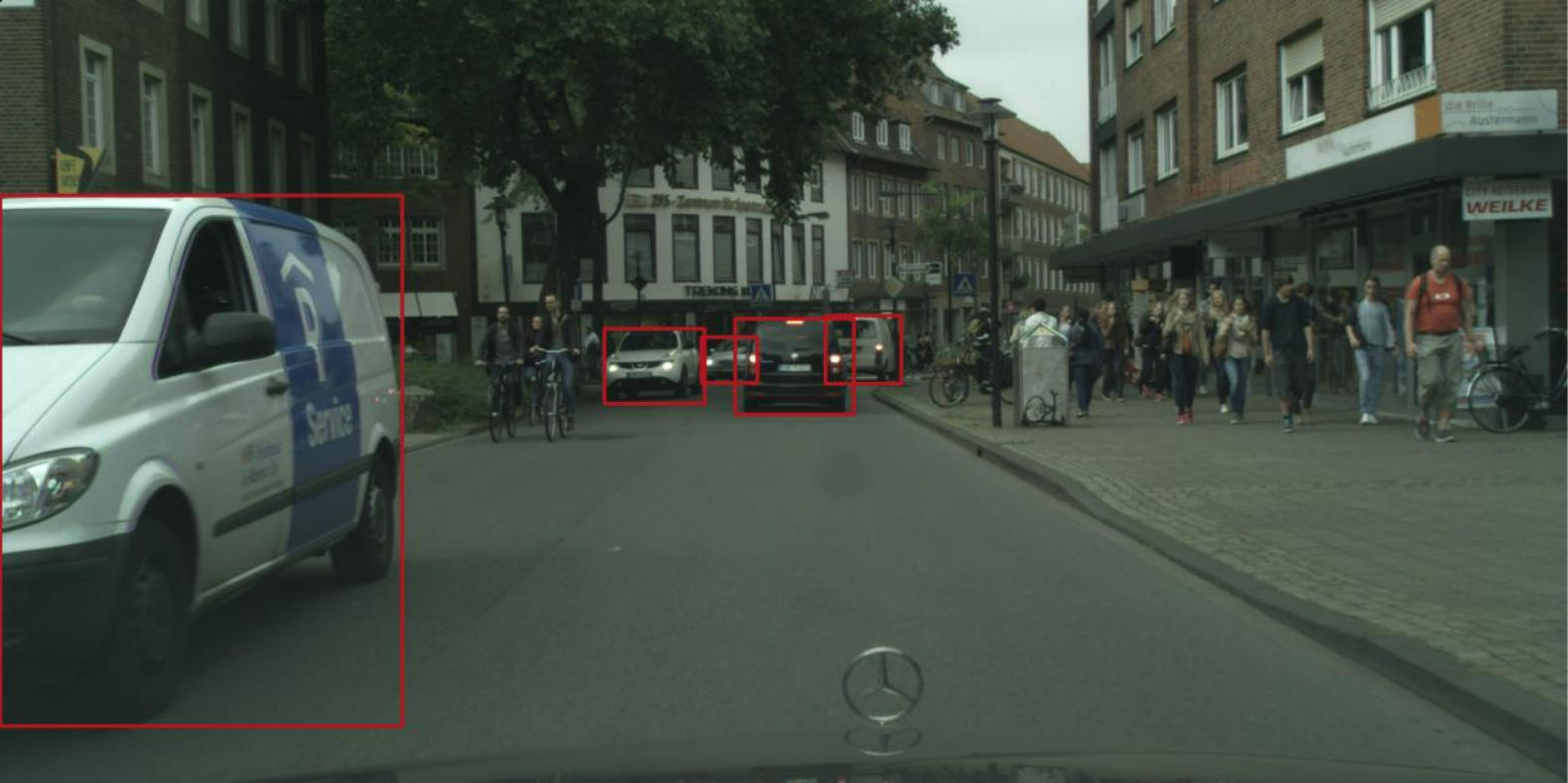}
\end{minipage}
\begin{minipage}[h]{0.99\linewidth}
\centering\footnotesize {(d) Sim10k $\rightarrow$ Cityscapes} 
\end{minipage}
\caption{Qualitative comparison of UaDAN with `Source only’ (no adaptation) and GPA~\cite{xu2020cross} over four domain adaptive detection tasks (a) Cityscapes $\rightarrow$ Mapillary Vistas, (b) Cityscapes $\rightarrow$ Foggy Cityscapes, (c) PASCAL VOC $\rightarrow$ Clipart1k and (d) Sim10k $\rightarrow$ Cityscapes: UaDAN outperforms GPA consistently by detecting more true positives (highlighted by yellow arrows) and less false positives (highlighted by red arrows) across all sample images. Bounding boxes of different color represent detection of different categories and a score threshold of 0.5 is used in visualization. Best viewed in color.}
\label{fig:result}
\end{figure*}

\begin{figure*}[ht]
\centering
\begin{minipage}[h]{0.32\linewidth}
\centering\footnotesize {Source only}
\end{minipage}
\begin{minipage}[h]{0.32\linewidth}
\centering\footnotesize {GPA~\cite{xu2020cross}} 
\end{minipage}
\begin{minipage}[h]{0.32\linewidth}
\centering\footnotesize {\textbf{UaDAN (Ours)}}
\end{minipage}
\begin{minipage}[h]{0.32\linewidth}
\centering\includegraphics[width=.99\linewidth]{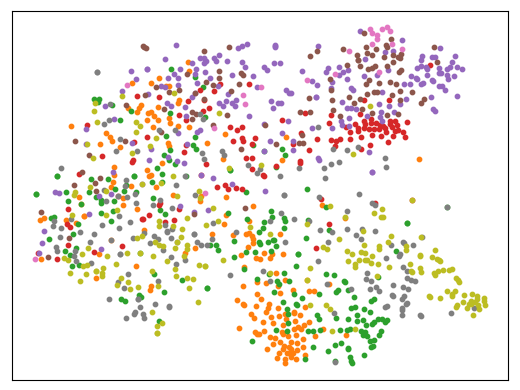}
\end{minipage}
\begin{minipage}[h]{0.32\linewidth}
\centering\includegraphics[width=.99\linewidth]{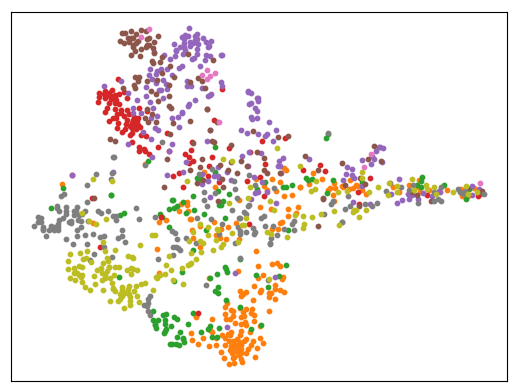}
\end{minipage}
\begin{minipage}[h]{0.32\linewidth}
\centering\includegraphics[width=.99\linewidth]{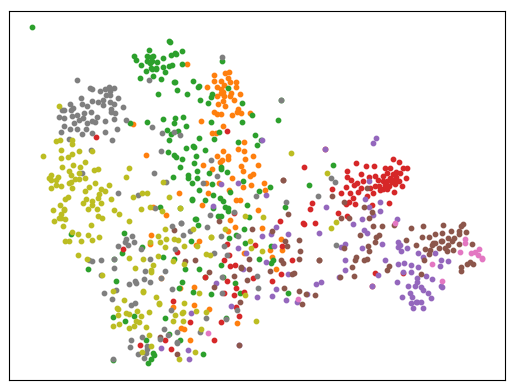}
\end{minipage}
\centering
\begin{minipage}[h]{0.32\linewidth}
\centering\footnotesize {$\sigma_{w}^{2}=416.8$, $\sigma_{b}^{2}=273.9$}
\end{minipage}
\vspace{2pt}
\begin{minipage}[h]{0.32\linewidth}
\centering\footnotesize {$\sigma_{w}^{2}=385.4$, $\sigma_{b}^{2}=302.9$} 
\end{minipage}
\begin{minipage}[h]{0.32\linewidth}
\centering\footnotesize {$\sigma_{w}^{2}=235.6$, $\sigma_{b}^{2}=507.2$}
\end{minipage}
\begin{minipage}[h]{0.99\linewidth}
\centering\footnotesize {(a) Cityscapes $\rightarrow$ Mapillary Vistas} 
\end{minipage}
\begin{minipage}[h]{0.32\linewidth}
\centering\includegraphics[width=.99\linewidth]{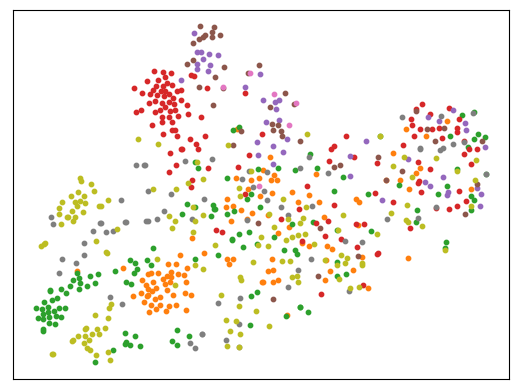}
\end{minipage}
\begin{minipage}[h]{0.32\linewidth}
\centering\includegraphics[width=.99\linewidth]{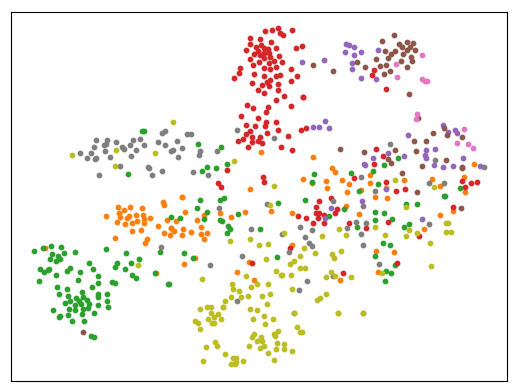}
\end{minipage}
\begin{minipage}[h]{0.32\linewidth}
\centering\includegraphics[width=.99\linewidth]{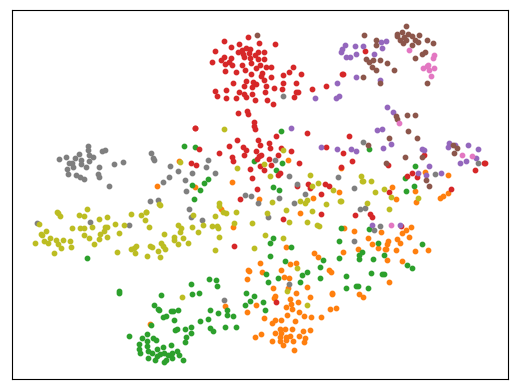}
\end{minipage}
\centering
\begin{minipage}[h]{0.32\linewidth}
\centering\footnotesize {$\sigma_{w}^{2}=370.8$, $\sigma_{b}^{2}=128.6$}
\end{minipage}
\vspace{2pt}
\begin{minipage}[h]{0.32\linewidth}
\centering\footnotesize {$\sigma_{w}^{2}=156.8$, $\sigma_{b}^{2}=223.3$} 
\end{minipage}
\begin{minipage}[h]{0.32\linewidth}
\centering\footnotesize {$\sigma_{w}^{2}=121.9$, $\sigma_{b}^{2}=276.3$}
\end{minipage}
\begin{minipage}[h]{0.99\linewidth}
\centering\footnotesize {(b) Cityscapes $\rightarrow$ Foggy Cityscapes} 
\end{minipage}
\caption{Visualization of target-domain feature distributions with t-SNE~\cite{maaten2008visualizing}: We calculate within-class variance $\sigma_{w}^{2}$ and between-class variance $\sigma_{b}^{2}$~\cite{otsu1979threshold} for two domain adaptive object detection tasks Cityscapes $\rightarrow$ Mapillary Vistas in (a) and Cityscapes $\rightarrow$ Foggy Cityscapes in (b). It can be seen that our method outperforms `Source only’ (no adaptation) and state-of-the-art GPA~\cite{xu2020cross} clearly. Note different colors represent different classes and best viewed in color.}
\label{fig:tsne}
\end{figure*}

\begin{figure*}[!ht]
\centering
\begin{minipage}[t]{0.49\linewidth}
\centering\includegraphics[width=.99\linewidth]{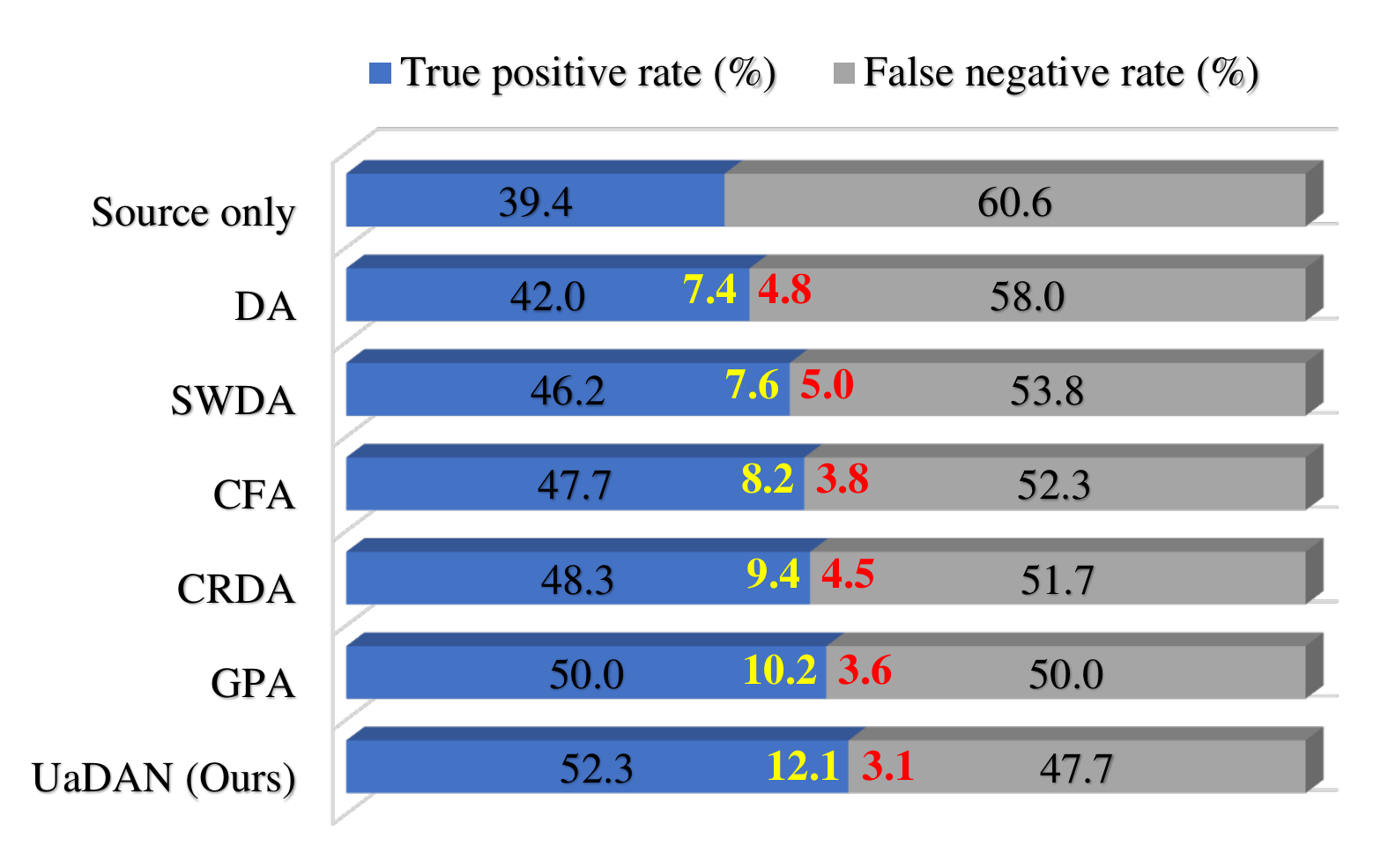}
\end{minipage}
\vspace{-5pt}
\begin{minipage}[t]{0.49\linewidth}
\centering\includegraphics[width=.99\linewidth]{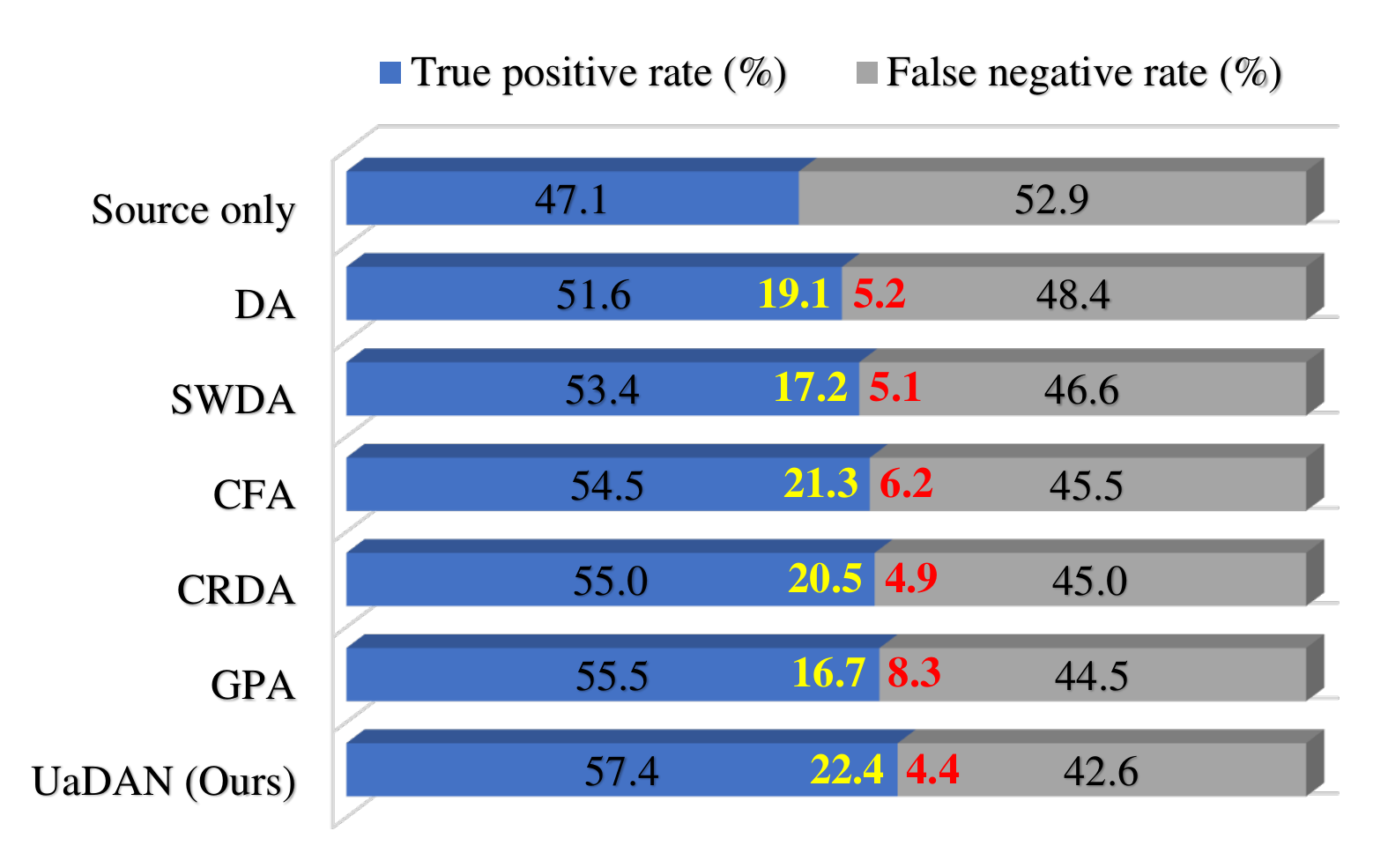}
\end{minipage}
\begin{minipage}[h]{0.49\linewidth}
\centering\footnotesize{(a) Cityscapes $\rightarrow$ Mapillary Vistas}
\end{minipage}
\begin{minipage}[h]{0.49\linewidth}
\centering\footnotesize{(b) PASCAL VOC $\rightarrow$ Clipart1k} 
\end{minipage}
\caption{Error analysis of UaDAN and state-of-the-art domain adaptive detection methods over two domain adaptive detection tasks: (a) Cityscapes $\rightarrow$ Mapillary Vistas and (b) PASCAL VOC $\rightarrow$ Clipart1k: UaDAN can generate most true positives which are falsely predicted by `Source only' (highlighted in yellow color), and fewest false negatives which are accurately predicted by `Source only' (highlighted in red color). Best viewed in color.}
\label{fig:discussion}
\end{figure*}

\subsection{Experimental Results}

We evaluate our uncertainty-aware domain adaptation method in four different domain shift scenarios: 1) \textit{Cross camera adaptation} with source and target images captured with different acquisition systems; 2) \textit{Weather adaptation} with source images captured in good weather conditions and target images in foggy weather; 3) \textit{Dissimilar adaptation} with source domain consisting of real-world images while target domain consisting of manually created images; and 4) \textit{Synthetic-to-realistic adaptation} with source and target domains consisting of synthetic and real-world images. In each domain shift scenario, we compare UaDAN with a number of state-of-the-art unsupervised domain adaptive methods DA~\cite{chen2018domain}, SWDA~\cite{saito2019strong}, CRDA~\cite{xu2020exploring}, CFA~\cite{zheng2020cross} and GPA~\cite{xu2020cross}.

\subsubsection{\textbf{Cross camera adaptation}}

Domain variance exists widely among images of different resolutions and qualities that are captured by using different acquisition sensors. For the task of object detection, we're facing more dramatic variations of object appearance in scale, viewpoints, etc. In this experiment, we study the effectiveness of our uncertainty-aware domain adaptation method while handling domain shifts among different real-image datasets. Specifically, we use 8 common object classes between the source dataset Cityscapes~\cite{cordts2016cityscapes} and the target dataset Mapillary Vistas~\cite{neuhold2017mapillary}. The validation set of the Mapillary Vista is used in evaluations.

Table~\ref{tab:bench1} shows comparisons of our UaDAN with state-of-the-art domain adaptive detection methods. We can see that UaDAN achieves the best detection with a mAP of 32.7\%. For the classes (\textit{e.g.}, train) which can hardly be detected by `Source only' (with only 7.1\% AP), UaDAN outperforms other methods by large margins (over 2.7\% AP). A possible reason is that the 'train' features in target domain tend to be under-aligned due to large domain bias, and UaDAN can focus on aligning under-aligned features while keeping the well-aligned features less affected.

\subsubsection{\textbf{Weather adaptation}}

Object detectors trained with normal-weather images usually do not perform well for images captured under adverse weather (\textit{e.g.}, foggy, rainy and nighttime). In practice, it's desired that object detectors can work well under different weather instead. In this study, we examine this issue by studying domain adaptation from normal weather to foggy weather. We use Cityscapes~\cite{cordts2016cityscapes} as the source dataset, Foggy Cityscapes~\cite{sakaridis2018semantic} as the target dataset, and the validation set of Foggy Cityscapes in evaluations.

Table~\ref{tab:bench2} shows the comparison of UaDAN with state-of-the-art domain adaptive methods over Cityscapes $\rightarrow$ Foggy Cityscapes. We can see that UaDAN achieves the best detection accuracy with 41.1\% mAP across all classes. Similarly for object classes (\textit{e.g.}, train) which cannot be accurately detected using `Source only' model (with only 9.6\% AP), UaDAN outperforms other methods by over 1.6\% in AP which is consistent with the experimental results on cross camera task. This experiment further verifies that our UaDAN can focus on aligning under-aligned samples. 

\subsubsection{\textbf{Dissimilar adaptation}}
Both cross camera adaptation and weather adaptation work in certain similar domains. We perform one more experiment to study how UaDAN adapts across dissimilar domains from real to artistic images. We use
PASCAL VOC~\cite{everingham2010pascal} and Clipart~\cite{inoue2018cross} as source and target datasets which share 20 common object classes. The validation set of Clipart is used in evaluations.

Table~\ref{tab:bench3} shows the comparison of UaDAN with state-of-the-art methods on this new task. We can see that UaDAN achieves the best accuracy with 40.2\% mAP across all classes. For object classes that cannot be detected well by `Source only' (\textit{e.g.}, aero, bottle, dog), UaDAN outperforms other methods by large margins (over 3.8\% in AP for each category). This further shows that our uncertainty-aware adversarial learning can generalize to different domain adaption tasks.

\subsubsection{\textbf{Synthetic-to-realistic adaptation}}
Using synthetic images in deep network training has been attracting increasing interest in recent years. However, models trained using synthetic images usually experiences clear performance drops while applied to real images. In this experiment, we study how UaDAN performs for adaptation from synthetic to real images. We use
SIM10k~\cite{johnson2017driving} and Cityscapes~\cite{cordts2016cityscapes} as the source and target datasets that share only one object category `car'. The validation set of Cityscapes is used in evaluations.

Table~\ref{tab:bench4} shows the comparison of UaDAN with the state-of-the-art over the synthetic-to-real task. We can see that UaDAN achieves the best accuracy with a mAP 48.6\%, showing that UaDAN is rather powerful in cross-domain detection task when the object category is small. We conjecture that the well-aligned car features could be over-aligned by the brute-force alignment while UaDAN keeps them less affected.

\begin{table*}[ht]
\renewcommand{\arraystretch}{1.3}
\caption{Ablation study of our method over domain adaptive detection task Cityscapes $\rightarrow$ Mapillary Vistas: uncertainty-aware adversarial learning (UaAL) outperforms the traditional adversarial learning (AL) consistently at both image level and instance level. uncertainty-guided curriculum learning (UgCL) further boosts the performance of the adversarial model. Note that mAP (\%) is evaluated over the Mapillary Vistas validation set. }
\centering
\begin{tabular}{p{3cm}|*{6}{p{0.8cm}}|p{1cm}} \hline
Method &$\mathcal{L}_{det}$ &$\mathcal{L}_{img}$ &$\mathcal{L}_{img}^{ua}$ & $\mathcal{L}_{ins}$ &$\mathcal{L}_{ins}^{ua}$ &$\mathcal{L}_{ins}^{ug}$ &{mAP} \\\hline
Baseline  &\checkmark & & & & & &25.8 \\ 
Image-level AL  &\checkmark &\checkmark & & & & &29.7 \\ 
Image-level UaAL  &\checkmark & &\checkmark & & & &30.8 \\ 
Instance-level AL &\checkmark & & &\checkmark & & &26.7 \\ 
Instance-level UaAL &\checkmark & & & &\checkmark & &29.8 \\ 
UaDAN w/o UgCL &\checkmark & &\checkmark & &\checkmark & &31.5 \\ 
\textbf{UaDAN}  &\checkmark & &\checkmark & &\checkmark &\checkmark &\textbf{32.7} \\ \hline
\end{tabular}
\label{tab:ablation}
\end{table*}

\subsubsection{\textbf{Qualitative comparison}}
The qualitative experimental results are well aligned with the quantitative results as shown in Fig.~\ref{fig:result}. We can see that UaDAN identifies more correct objects with less false positives than GPA~\cite{xu2020cross} across all four target datasets. This also shows that UaDAN can keep well-aligned features less affected while brute-force alignment in GPA~\cite{xu2020cross} maps them to incorrect categories. 
Detection may fail if `Source only' predicts false positives with high confidence (low entropy) as shown in the first rows of Figs.~\ref{fig:result} (b) and (c).
Though UaDAN may generate false positives by keeping falsely identified well-aligned features from alignment, it can globally generate less false positives and more true positives by focusing on aligning poorly-aligned features in these scenarios.

\subsection{Discussion}

\subsubsection{\textbf{Feature Visualization}}

In the previous sections, we have shown that UaDAN achieves superior object detection performance as compared with state-of-the-art domain adaptive methods. To further analyse the behaviour of detection models, we employ t-SNE~\cite{maaten2008visualizing} to visualize the target-domain feature distributions as learnt by different cross-domain detection methods. In the visualization experiment, we randomly select 200 instances for each category (all instances are selected for categories with less than 200 instances). Fig.~\ref{fig:tsne} shows the visualization, where the within-class variance and between-class variances are calculated for quantitative analysis. As Fig.~\ref{fig:tsne} shows, the within-class and between-class variances are highly consistent with the object detection in Fig.~\ref{fig:result}.

\subsubsection{\textbf{Error Analysis}}

To further validate the effectiveness of the uncertainty awareness in protecting the well-aligned features from misalignment, we analyse the errors induced by domain adaptive methods as compared with `Source only' (no adaptation). As Fig.~\ref{fig:discussion} shows, we calculate the rate of true positives that are falsely predicted by `Source only' (highlighted in yellow), and the rate of false negatives that are accurately predicted by `Source only' (highlighted in red). 

We can see that certain samples are accurately predicted by `Source only' but falsely predicted by domain adaptive methods due to misalignment. UaDAN can alleviate this problem and produce less false negatives than other domain adaptive methods by keeping well-aligned features less affected. For samples that are falsely predicted by `Source only', UaDAN can also produce more true positives than other methods by focusing on aligning under-aligned features.

\subsubsection{\textbf{Computational Overhead}}
We would clarify that UaDAN introduces very limited extra computation overhead in training (less than $0.0003$ second per iteration on one GPU which translates to less than 0.1\% in percentages) as compared to the traditional adversarial method~\cite{chen2018domain}, as entropy computation is simple and efficient. Meanwhile, UaDAN introduces no computation overhead during inference, as entropy computation is included in training stage only.
However, it outperforms traditional adversarial methods by large margins, e.g. it outperform~\cite{chen2018domain} by over $4.3\%$ in AP over all domain adaptive detection tasks as shown in Tables~\uppercase\expandafter{\romannumeral2}$\sim$\uppercase\expandafter{\romannumeral5}.

Since the framework of UaDAN is built upon Faster R-CNN, we further discuss the gap between them in in terms of accuracy, detection speed and model complexity. Faster R-CNN~\cite{ren2015faster} with no adaptation trains a `Source only' model by using labelled source data. The trained model does not perform well for target data due to domain gaps. The proposed UaDAN instead introduces uncertainty-aware domain classifiers that align source and target domains for better performance over target samples. Specifically, UaDAN achieves much better AP (improved by over $6.9\%$) over all domain adaptive detection tasks as shown in Tables~\uppercase\expandafter{\romannumeral2}$\sim$\uppercase\expandafter{\romannumeral5}. In addition, it has the same detection speed and model complexity as Faster R-CNN since domain adaptation modules are included in training stage only.

\begin{figure}[!h]
\centering
\includegraphics[width=.99\linewidth]{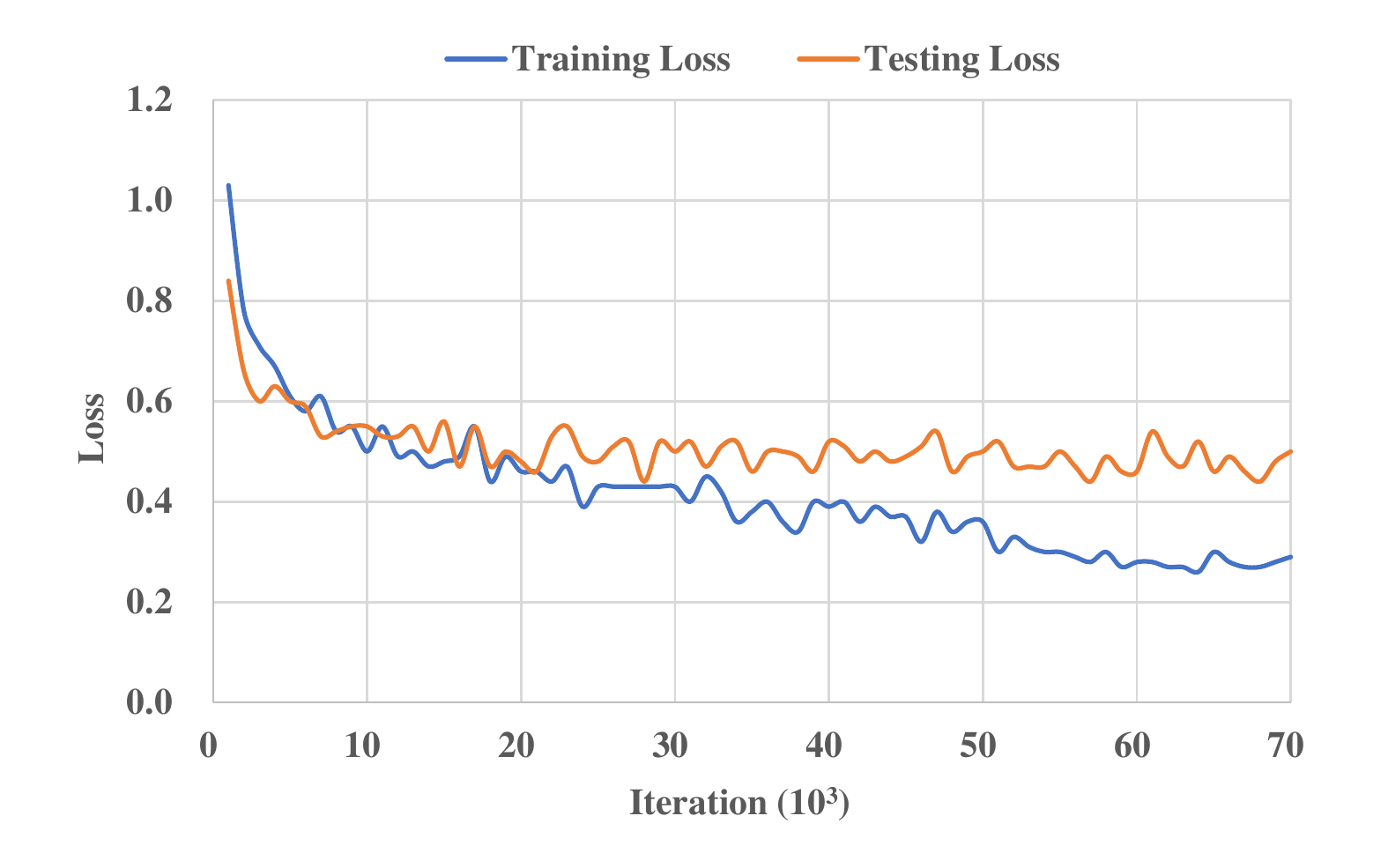}
\centering
\caption {The training and testing losses of our method over domain adaptive detection task Cityscapes $\rightarrow$ Mapillary Vistas.}
\label{fig:loss}
\end{figure}

\subsubsection{\textbf{Training and Testing Processing}}
We also compute the training and testing losses of the UaDAN in every $10^3$ iterations. As shown in Fig.~\ref{fig:loss}, both training and testing losses decrease quickly during the first 10k iterations and then fluctuate in a small range after 50k iterations. 

\subsection{Ablation Studies}
We perform a series of ablation experiments to study how UaDAN components contribute to overall performance. Seven models are trained as listed in Table~\ref{tab:ablation}: 1) \textit{Baseline} which is `Source only’ as trained by source samples without adaptation; 2) \textit{Image-level AL} which is image-level adversarial model as trained by $\mathcal{L}_{det}$ and image-level adversarial loss $\mathcal{L}_{img}$; 3) \textit{Image-level UaAL} which is image-level uncertainty-aware adversarial model as trained by $\mathcal{L}_{det}$ and entropy-weighed image-level adversarial loss $\mathcal{L}_{img}^{ua}$; 4) \textit{Instance-level AL} which is instance-level adversarial model as trained by $\mathcal{L}_{det}$ and instance-level adversarial loss $\mathcal{L}_{ins}$; 5) \textit{Instance-level UaAL} which is instance-level entropy-weighed adversarial model as trained by $\mathcal{L}_{det}$ and instance-level entropy-weighed adversarial loss $\mathcal{L}_{ins}^{ua}$; 6) \textit{UaDAN w/o UgCL} which is UaDAN model without uncertainty-guided curriculum learning. (trained by $\mathcal{L}_{det}$, $\mathcal{L}_{img}^{ua}$ and $\mathcal{L}_{ins}^{ua}$); and 7) \textit{UaDAN} which is the complete UaDAN model as trained by $\mathcal{L}_{det}$, $\mathcal{L}_{img}^{ua}$, $\mathcal{L}_{ins}^{ua}$, and $\mathcal{L}_{ins}^{ug}$.

As Table~\ref{tab:ablation} shows, \textit{Image-level AL}  and \textit{Instance-level AL} both outperform the \textit{Baseline} consistently, which demonstrates the importance of feature representation alignment at both image level and instance level in cross-domain detection tasks. Specifically, the gain of \textit{Image-level AL} is much higher than \textit{Instance-level AL}, largely because harder instance-level alignment can work only when the easier image-level alignment works. In addition, we can observe that \textit{Image-level UaAL} and  \textit{Instance-level UaAL} outperform \textit{Image-level AL} and \textit{Instance-level AL} consistently in both image-level and instance-level detection tasks, demonstrating the importance of uncertainty-aware alignment in keeping well-aligned features less affected. Further, \textit{UaDAN w/o UgCL} outperforms both \textit{Instance-level UaAL} and \textit{Instance-level UaAL}, which shows that Image-level UaAL and Instance-level UaAL are  complementary.
Finally, \textit{UaDAN} outperforms \textit{UaDAN w/o UgCL} with a large margin, which verifies the effectiveness of our proposed uncertainty-guided curriculum learning.

\begin{table}[!h]
\renewcommand{\arraystretch}{1.3}
\centering
\caption{The sensitivity of parameter $\xi$: UaDAN obtains the best performance consistently when $\xi=0.5$. Note that mAP (\%) is evaluated over the validation set of each domain adaptive detection task.}
\begin{tabular}{p{1.5cm}|*{5}{p{1.5cm}}} 
\hline 
\multicolumn{1}{c|}{$\xi$} & \multicolumn{1}{c}{0} & \multicolumn{1}{c}{0.25} & \multicolumn{1}{c}{0.5} & \multicolumn{1}{c}{0.75} & \multicolumn{1}{c}{1}
\\\hline
\multicolumn{1}{c|}{Cityscapes $\rightarrow$ Mapillary Vistas} &\multicolumn{1}{c}{30.8} &\multicolumn{1}{c}{31.5} &\multicolumn{1}{c}{\textbf{32.7}}  &\multicolumn{1}{c}{32.1}  &\multicolumn{1}{c}{31.5}\\
\multicolumn{1}{c|}{Cityscapes $\rightarrow$ Foggy Cityscapes} &\multicolumn{1}{c}{39.5} &\multicolumn{1}{c}{40.5} &\multicolumn{1}{c}{\textbf{41.1}}  &\multicolumn{1}{c}{40.7}  &\multicolumn{1}{c}{40.0}\\
\multicolumn{1}{c|}{PASCAL VOC $\rightarrow$ Clipart1k} &\multicolumn{1}{c}{37.6} &\multicolumn{1}{c}{38.7} &\multicolumn{1}{c}{\textbf{40.2}}  &\multicolumn{1}{c}{39.6}  &\multicolumn{1}{c}{38.8}\\
\hline
\end{tabular}
\label{tab:xi}
\end{table}

we also study the sensitivity of parameter $\xi$ over three domain adaptive detection tasks as shown in Table~\ref{tab:xi}. Specifically, \textit{UaDAN} degrades to \textit{Image-level UaAL} when $\xi=0$, where the instance-level alignment is not activated as entropy cannot be negative. It degrades to \textit{UaDAN w/o UgCL} (described in the Section of Ablation Studies) when $\xi=1$, where the instance-level alignment is always activated as the entropy is smaller than 1. In addition, the best performance is obtained when $\xi=0.5$ consistently over all three tasks.

\section{Conclusion}
This paper presents an uncertainty-aware domain adaption technique for unsupervised domain adaptation in object detection. We design an uncertainty-aware adversarial learning algorithm that can keep well-aligned features less affected in both proposal generation and object detection tasks. In addition, we design a uncertainty-guided curriculum learning algorithm that can alleviate the side effect of the traditional adversarial learning in handling harder detection tasks. Extensive experiments over four challenging cross-domain detection tasks demonstrate the effectiveness of the proposed method.  
Moving forwards, we will explore how to adapt the proposed technique to other domain adaptive tasks such as image classification and semantic segmentation.

\section*{Acknowledgment}
This research was conducted at Singtel Cognitive and Artificial Intelligence Lab for Enterprises (SCALE@NTU), which is a collaboration between Singapore Telecommunications Limited (Singtel) and Nanyang Technological University (NTU) that is funded by the Singapore Government through the Industry Alignment Fund ‐ Industry Collaboration Projects Grant. 

\ifCLASSOPTIONcaptionsoff
  \newpage
\fi



%
\bibliographystyle{IEEEtran}
\bibliography{bibliography}

\end{document}